\pdfoutput=1

\documentclass{article} 
\usepackage{iclr2025_conference,times}

\usepackage{amsmath,amsfonts,bm}

\def\eqref#1{equation~\ref{#1}}

\def\1{\bm{1}}

\DeclareMathAlphabet{\mathsfit}{\encodingdefault}{\sfdefault}{m}{sl}
\SetMathAlphabet{\mathsfit}{bold}{\encodingdefault}{\sfdefault}{bx}{n}

\usepackage[utf8]{inputenc} 
\usepackage[T1]{fontenc}    
\usepackage{hyperref}       
\usepackage{hyperref}
\usepackage{url}
\usepackage{booktabs}       
\usepackage{amsfonts}       
\usepackage{nicefrac}       
\usepackage{microtype}      
\usepackage{xcolor}         
\usepackage{graphicx} 
\usepackage{longtable}
\usepackage{afterpage}
\usepackage{makecell}
\usepackage{booktabs}
\usepackage{amsmath}
\usepackage{tabularx}
\usepackage{authblk}
\usepackage{subcaption}
\usepackage{wrapfig}
\usepackage{multirow}
\usepackage{multicol}
\usepackage{utfsym}
\usepackage{arydshln}
\usepackage{listings}

\newcommand{\benchmark}{UGMathBench}
\newcommand{\revise}[1]{\textcolor{black}{#1}}
\NewDocumentCommand{\xx}
{ mO{} }{\textcolor{blue}{\textsuperscript{\textit{todo}}\textsf{\textbf{\small[#1]}}}}

\title{\benchmark: A Diverse and Dynamic Benchmark for Undergraduate-Level Mathematical Reasoning with Large Language Models}

\author{Xin Xu$^*$, Jiaxin Zhang$^*$, Tianhao Chen$^*$, Zitong Chao, Jishan Hu, Can Yang \\
Department of Mathematics\\
The Hong Kong University of Science and Technology\\
\texttt{\{xxuca, jzhangat, tchenbb, zchaoaa\}@connect.ust.hk} \\
\texttt{\{majhu, macyang\}@ust.hk}
}

\iclrfinalcopy 
\begin{document}
\maketitle
\renewcommand*{\thefootnote}{\fnsymbol{footnote}}
\footnotetext{* Equal contribution.}
\renewcommand*{\thefootnote}{\arabic{footnote}}
\begin{abstract}
Large Language Models (LLMs) have made significant strides in mathematical reasoning, underscoring the need for a comprehensive and fair evaluation of their capabilities. 
However, existing benchmarks often fall short, either lacking extensive coverage of undergraduate-level mathematical problems or probably suffering from test-set contamination. 
To address these issues, we introduce {\benchmark}, a diverse and dynamic benchmark specifically designed for evaluating undergraduate-level mathematical reasoning with LLMs. {\benchmark} comprises 5,062 problems across 16 subjects and 111 topics, featuring 10 distinct answer types. 
Each problem includes three randomized versions, with additional versions planned for release as leading open-source LLMs become saturated in {\benchmark}. 
Furthermore, we propose two key metrics: effective accuracy (EAcc), which measures the percentage of 
correctly solved problems across all three versions, and reasoning gap ($\Delta$), which assesses reasoning robustness by calculating the difference between the average accuracy across all versions and EAcc. 
Our extensive evaluation of 23 leading LLMs reveals that the highest EAcc achieved is 56.3\% by OpenAI-o1-mini, with large $\Delta$ values observed across different models.
This highlights the need for future research aimed at developing "large reasoning models" with high EAcc and $\Delta = 0$.
We anticipate that the release of {\benchmark}, along with its detailed evaluation codes, will serve as a valuable resource to advance the development of LLMs in solving mathematical problems.
Codes and data are available at \href{https://github.com/YangLabHKUST/UGMathBench}{https://github.com/YangLabHKUST/UGMathBench}.
\end{abstract}

\section{Introduction}\label{sec: intro}

Mathematical reasoning and problem-solving are critical components of human intelligence, and the ability of machines to understand and address mathematical challenges is crucial for their deployment \citep{llm4math2024Ahn, MathBench2024Liu, OlympiadBench2024He}. 
Solving mathematical problems with machines has been a significant research topic in natural language processing since the 1960s \citep{MWP1964bobrow}, initially focusing on elementary math word problems \citep{SVAMP2021patel, mwpmethod2017wang, mwpmethod12017ling, mwpmethod32017welbl, gsm8k2021cobbe}.
With the advent of Large Language Models (LLMs) \citep{GPT32020Brown, InstructGPT2022Ouyang, GPT42023openai, gemini2023team, claude3}, interest in using these advanced technologies to solve math problems has continued to grow. 
Researchers are exploring various approaches to improve the mathematical reasoning capabilities of LLMs, including prompting \citep{CoT2022Wei, CoT-SC2022Wang, zero-shot-CoT2022Kojima, CR2023Zhang, PHP2023Zheng, xue2024decompose}, supervised fine-tuning \citep{mammoth2023yue, metamath2023yu, tora2023Gou, xwinmath2024li, E-GSM2024Xu, dartmath2024tong, yan2024s,xu2025tata, yan2025mathfimer}, and continued pretraining \citep{minerva202lewkowycz, deepseekmath2024shao, llemma2023azerbayev}. 
Consequently, LLMs have become increasingly capable of solving complex mathematical problems \citep{MATH2021hendrycks, llm4math2024Ahn}.

With the rapid advancements in LLMs, evaluating their reasoning capabilities has become increasingly important~\citep{OlympiadBench2024He,xu2025ugphysics, phan2025humanity}, especially in mathematics~\citep{MathBench2024Liu,gao2024omni}.
Although benchmarks such as GSM8K \citep{gsm8k2021cobbe} and MATH \citep{MATH2021hendrycks} are commonly used to assess these abilities, they are becoming insufficient due to rapid progress in model performance, as evidenced by accuracy exceeding 97\% in GSM8K \citep{saturated2023zhou} and 94.8\% in MATH \citep{o1}. 
While more challenging benchmarks are being introduced \citep{CollegeMath2024Tang, MathBench2024Liu}, they often remain limited in size and scope regarding undergraduate-level mathematics, which is essential due to its breadth and complexity (see Table~\ref{tab:compare_with_existing_benchmark}).
Moreover, there are growing concerns about test set contamination in these static benchmarks \citep{MATH_FUNC2024srivastava, GSM1K2024zhang, VarBench2024qian, LiveBench2024white}. 
Recent efforts \citet{VarBench2024qian, MATH_FUNC2024srivastava} have introduced dynamic benchmarks by functionalizing the original problems in GSM8K and MATH to generate randomized variations through variable disturbance. 
However, these initiatives primarily focus on elementary and competition-level mathematics (see Table~\ref{tab:compare_with_existing_benchmark}).
These limitations highlight the pressing need for a comprehensive and dynamic benchmark specifically designed to assess undergraduate-level mathematical reasoning.

\begin{figure}
    \centering
    \includegraphics[width=1\linewidth]{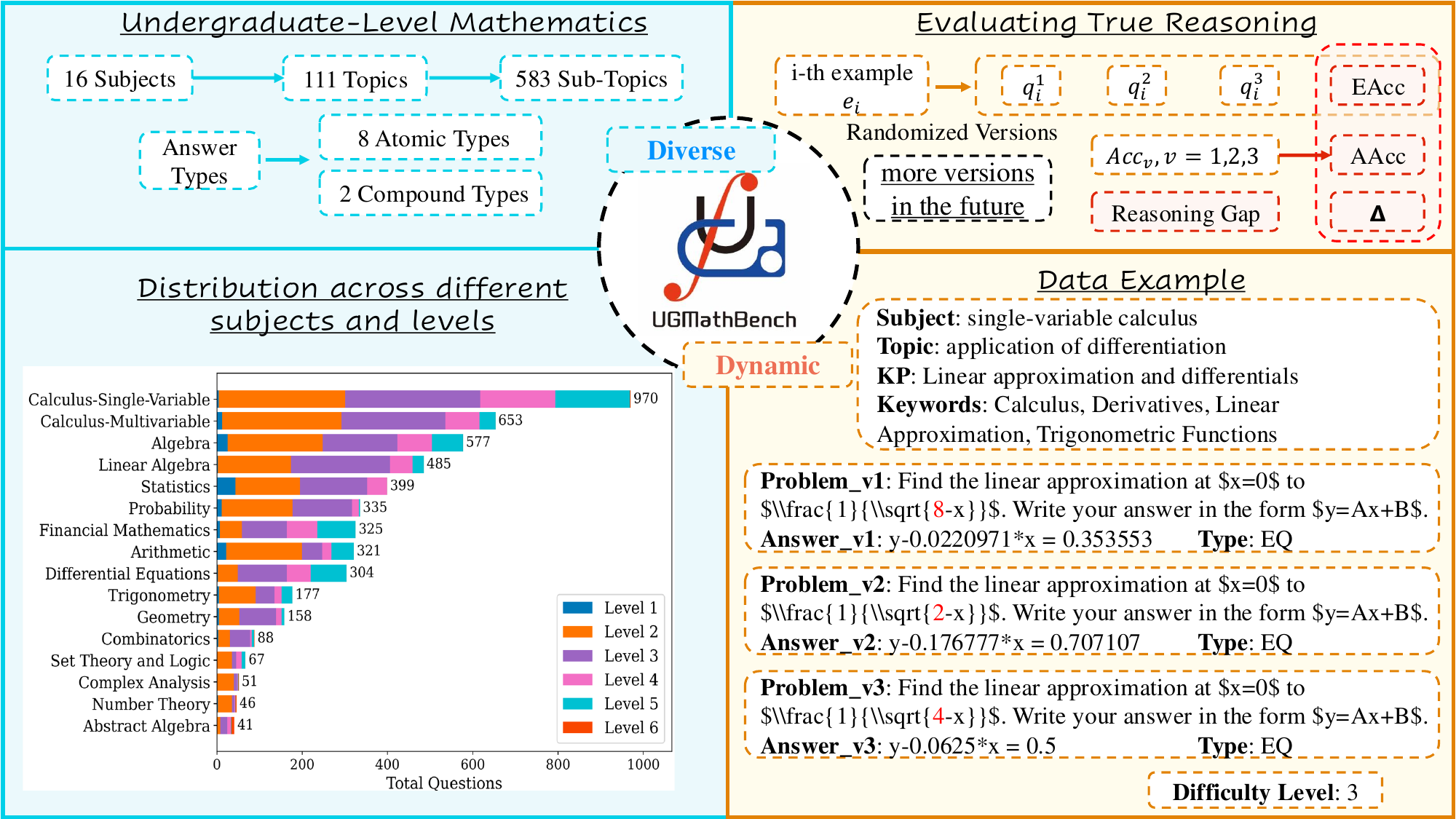}
    \caption{Overview of {\benchmark}. {\benchmark} is a diverse and dynamic benchmark specifically designed for evaluating undergraduate-level mathematics with LLMs, covering 16 distinct subjects and featuring 10 different answer types. Each problem contains three randomized versions, with EAcc and $\Delta$ rigorously assessing LLMs' true reasoning skills.}
    \label{fig:overview}
\end{figure}

In this paper, we present {\benchmark}, a diverse and dynamic benchmark designed to evaluate the mathematical reasoning capabilities of LLMs across a wide range of undergraduate-level mathematical topics, as illustrated in Figure~\ref{fig:overview}.
We meticulously collect, clean, and format undergraduate-level mathematical problems from our online homework grading system (see Appendix~\ref{app: data_source}), resulting in a benchmark comprising 5,062 problems in 16 subjects, categorized into eight atomic answer types and two compound answer types.
A key feature of {\benchmark} is the inclusion of multiple randomized versions for each problem, which aids in assessing the true reasoning abilities of LLMs through the \textbf{E}ffective \textbf{Acc}uracy (EAcc) and reasoning gap ($\Delta$) (see Section~\ref{sec: metric}).
EAcc represents the percentage of problems correctly solved across all versions, providing insights into intrinsic reasoning skills. It operates on the premise that a model capable of solving a problem through reasoning should also be able to solve all its variants under variable disturbance \citep{MATH_FUNC2024srivastava, VarBench2024qian}.
The reasoning gap, $\Delta$, is defined as the difference between the average accuracy across all versions and the EAcc, quantifying the robustness of reasoning when the original problems undergo slight modifications.
These metrics help mitigate the impact of potential test set contamination \citep{contamination12023deng, contamination22024dong, contamination32023golchin, contamination42023roberts} and ensure a more rigorous evaluation of LLMs' mathematical reasoning abilities.
These features are summarized more clearly in Figure~\ref{fig:overview} and Table~\ref{tab:compare_with_existing_benchmark}.

We conducted an extensive evaluation of the leading LLMs, including proprietary models such as OpenAI-o1 \citep{o1} and open-source models like LLaMA-3-Instruct \citep{llama3modelcard}.
Despite their advanced capabilities, the best EAcc achieved is 56.3\% by OpenAI-o1 \citep{o1} and all LLMs exhibit a large reasoning gap.
These results highlight the considerable challenges that {\benchmark} presents to current LLms in terms of mathematical reasoning, underscoring the need for future research focused on developing "large reasoning models" characterized by high EAcc and a reasoning gap $\Delta = 0$.
To summarize our key findings:

1. Even LLMs with the most advanced reasoning ability, OpenAI-o1-mini, achieves a 56.30\% EAcc on {\benchmark}, much lower on other text-only mathematical benchmarks.
2. All LLMs evaluated exhibit high reasoning gap with Robustness Efficiency (RE, the ratio between $\Delta$ and EAcc) ranging from 20.78\% to 196.6\%, pinpointing the inconsistencies of current LLMs in solving problems with variable disturbance.
3. There remains a significant discrepancy among closed-source LLMs and open-source LLMs (even specialized mathematical LLMs). Among open-source LLMs, only Qwen-2-Math-72B-Instruct and Mistral-Large-Instruct have comparable performance with GPT-4o.
4. The average EAcc varies by subject, with Arithmetic scoring 62.8\%. In contrast, Abstract Algebra, Differential Equations, and Financial Mathematics have average EAccs of less than 10\%.
5. An error analysis of OpenAI-o1-mini's performance reveals that calculation errors are a major concern. 
Even the same problem presented in different randomized versions can lead to varying types of errors.


\begin{table*}[h!]
\centering
\footnotesize
\caption{Comparison of various benchmarks. "\#Types" indicates the number of answer types in the dataset. "\#Subjects" specifies the number of mathematical subjects covered. "Dynamic" denotes whether the dataset is dynamic or static.
"\#Test" shows the number of test examples in the dataset, while "\#College" refers to the number of test examples at the college level.}
\begin{tabular}{llccccc}
\toprule
 \textbf{Dataset} & \textbf{Level}  & \textbf{\#Types} & \textbf{\#Subjects}& \textbf{Dynamic} & \textbf{\#Test} & \textbf{\#College}  \\
\midrule
GSM8K & Elementary & 1  & - & \usym{2717} & 1,319 & 0 \\
MATH & Competition & 3  & 7 & \usym{2717} & 5,000 & 0 \\
MMLU-Math & All & 1 & - & \usym{2717} & 844 & 116\\
TAL-SCQ & K12 Math & 1 & - & \usym{2717} & 1,496 & 0 \\
AGIEval-SAT-Math & High School & 2 & - & \usym{2717} & 102 & 0\\
AGIEval-Math & Competition & 2 & - & \usym{2717} & 938 & 0 \\
CollegeMath & College & 3 & 7 & \usym{2717} & 2,818 & 2,818 \\
MathBench & All & 1 & 5 & \usym{2717} & 1781 & 466\\
GSM1K & Elementary & 1 & - & \usym{1F5F8} & 1,250 & 0 \\
FN-EVAL & Competition & 3 & 7 & \usym{1F5F8} & 2,060 & 0 \\
VarBench-Math & Elementary & 1 & - & \usym{1F5F8} & 1,319 & 0 \\
LiveBench-Math & Competition & 2 & - & \usym{1F5F8} & 232 & 0 \\
\midrule
{\benchmark} & College & \textbf{10} & \textbf{16} & \textbf{\usym{1F5F8}} & \textbf{5,062} & \textbf{5,062} \\
\bottomrule
\end{tabular}
\label{tab:compare_with_existing_benchmark}
\end{table*}

\section{Related Work}

\textbf{Mathematical Benchmarks}.
Mathematical reasoning is increasingly vital for assessing the fundamental reasoning capabilities of LLMs \citep{llm4math2024Ahn}. 
Several math-related datasets have been proposed in this area \citep{mawps2016koncel, mathqa2019amini, MMLU2020hendrycks,  gsm8k2021cobbe, MATH2021hendrycks, Math23k2022chen}. 
Among these, GSM8K \citep{gsm8k2021cobbe} and MATH \citep{MATH2021hendrycks} are the most representative datasets for elementary and high school-level math reasoning, respectively. 
However, as modern LLMs become increasingly powerful, these benchmarks lack sufficient challenge forlatest LLMs.
Notably, o1 \citep{o1} achieves 94.8\% accuracy on MATH, which was previously considered highly complex. 
To better assess the mathematical reasoning abilities of current LLMs, some researchers create variants of existing benchmarks \citep{GSM-IC2023shi, R-GSM2024chen, GSM-Plus2024Li, E-GSM2024Xu}, while others propose new, more challenging math reasoning benchmarks \citep{theoremqa2023chen, SciBench2023Wang, collins2024evaluating, CollegeMath2024Tang, MathBench2024Liu, MMLU-pro2024Wang}. 
CollegeMath \citep{CollegeMath2024Tang} covers several college-level mathematics subjects with limited answer types.
In contrast, our {\benchmark} encompasses a broader range of subjects, answer types, and test examples. 
In addition, there are also several cross-modality math-related datasets \citep{geoqa2021chen, MathVista2023lu, pds2024xu, MMMU2024yue, cmmu2024he, OlympiadBench2024He, OlympicArena2024huang, MMMU-Pro2024yue}.

\textbf{Dynamic Benchmarks for Mathematical Reasoning}.
Test set contamination, wherein benchmark test data appear in a newer model's training set, significantly challenges fair LLM evaluation by artificially inflating performance \citep{contamination12023deng, contamination22024dong, contamination32023golchin, contamination42023roberts}. 
Since pretraining data often involve large corpora scraped from the Internet, any static benchmark risks data contamination \citep{GSM1K2024zhang, VarBench2024qian}. 
To mitigate this, recent benchmarks maintain private test sets \citep{GSM1K2024zhang, OlympicArena2024huang}, requiring anyone who wishes to evaluate their models to submit predictions for centralized processing before publishing results on their leaderboards. 
However, this process can be inefficient and lacks transparency for error analysis \citep{VarBench2024qian}. 
An alternative is releasing dynamic benchmarks that are periodically updated \citep{MATH_FUNC2024srivastava, VarBench2024qian, LiveBench2024white}. 
For example, \citet{MATH_FUNC2024srivastava} have functionalized a subset of the MATH dataset to regenerate new versions of the test set by reassigning variable values. 
In this vein, our {\benchmark} is a dynamic benchmark featuring different sampled values for variables by setting distinct random seeds. 
Currently, we release three snapshots for each question in {\benchmark} and plan to release new versions if leading open-source LLMs reach accuracy saturation.
\section{The {\benchmark} Benchmark}\label{sec: benchmark}

\subsection{{\benchmark} Overview}\label{sec: overview}
\begin{wraptable}{r}{0.31\textwidth}
    \centering
    \caption{Benchmark Statistics}
    \resizebox{0.31\textwidth}{!}{ 
    \begin{tabular}{l r}
    \toprule
    \textbf{Statistic} & \textbf{Number} \\
    \midrule
    Total Problems & 5562 \\
    Number of Versions & x 3 \\
    Total Subjects/topics & 16/111 \\
    Total Answer Types & 10 \\
    Total Difficulty Level & 6 \\
        \midrule
    Average Problem Tokens & 122.63 \\
    Average Number of Answers & 2.77 \\
    \bottomrule
    \end{tabular}
    }
    \label{tab:benchmark_statistics}
\end{wraptable}

We introduce the {\benchmark}, a dynamic undergraduate-level mathematical reasoning benchmark designed to thoroughly and robustly assess the mathematical reasoning ability of LLMs.
{\benchmark} enables fair evaluation through randomized versions of single problems. 
Unlike GSM1K \citep{GSM1K2024zhang}, our test set labels are publicly available, facilitating efficient evaluation and effective error analysis.
{\benchmark} covers fifteen core subject areas in undergraduate-level mathematics, including single-variable calculus, multivariable calculus, differential equations, probability, and more, encompassing a total of 111 specific topics (details in Appendix~\ref{app: problem_dist}). 
{\benchmark} comprises a set of 5,062 problems in 3 different randomized snapshots with 10 different answer types (see Appendix~\ref{app: answer_type}).
These answer types range from atomic types (e.g., numerical value, expression) to compound types (e.g., multiple answers in ordered or unordered lists), setting {\benchmark} apart from many other math-related benchmarks that focus primarily on a single answer with an atomic type.
\revise{We randomly select 100 problems to examine student performance using our grading system' records, with each problem being completed by varying numbers of students ranging from 99 to 1,537. 
The average accuracy on the first attempt is 56.5\%, while the average accuracy on the final attempt increased to 96.1\%.}

\subsection{{\benchmark} Creation}\label{sec: creation}

Our {\benchmark} creation process has three distinct phases: data collection, data cleaning \& deduplication, and answer type annotation.

\textbf{Data Collection}.
The dataset for {\benchmark} is carefully compiled from the online grading system of our institute's undergraduate courses (see Appendix~\ref{app: data_source}). 
\revise{All problems in our system are generated by programs that specify particular variable values to ensure correctness and maintain the same solution (see Appendix~\ref{app: dynamic_mechanism}).}
We gather all mathematics-related problems, resulting in 16 subjects and 111 topics in total. 
To prevent student cheating, our grading system offers randomized versions of most problems (see Figure~\ref{fig:overview}), similar to the variable disturbance approach in \citet{VarBench2024qian}. 
To create a dynamic benchmark, we exclude static problems without randomized versions, as well as those containing images, ensuring a text-only reasoning benchmark.
The collected problems are originally in HTML format.

\textbf{Data Cleaning and Deduplication}.
After collecting problems in the HTML format, we utilize the \texttt{bs4}\footnote{\href{https://pypi.org/project/beautifulsoup4/}{https://pypi.org/project/beautifulsoup4/}} and \texttt{re}\footnote{\href{https://docs.python.org/3/library/re.html}{https://docs.python.org/3/library/re.html}} Python packages to convert them into Latex.
Since no conversion process is flawless, we manually verify the converted LaTeX files against the original HTML files.
The latex files are then further organized into the format shown in Figure~\ref{fig:overview}.
After converting and cleaning all the problems, we perform deduplication within each subject based on embeddings generated by \texttt{text-embedding-ada-002} to remove duplicated problems (see Appendix~\ref{app: deduplication}).
\revise{The thresholds and the number of questions that are filtered out are given in Table~\ref{tabapp:deduplication}.}

\begin{wraptable}{r}{0.31\textwidth}
\centering
\footnotesize
\caption{Examples of eight atomic answer types.}
\resizebox{0.31\textwidth}{!}{
\begin{tabular}{lc}
\toprule
\textbf{Type}     & \textbf{Example}                         \\ \hline
Numerical Value  & \(\pi/4\)                             \\
Expression  & \(x^2 + 1\)                \\
Equation  & \(x^2+y^2=1\)                   \\
Interval & \((-\infty,-1]\)             \\ 
True/False & Yes \\
MC with single answer & A \\
MC with multiple answers & ACF \\
Open-Ended & h(1-x)\\
\bottomrule
\end{tabular}
}
\label{tab:examples of answer types}

\end{wraptable}

\textbf{Answer Type Annotation}.
Meta-information (e.g. subject, topic, subtopic, difficulty level as shown in Figure~\ref{fig:overview}) is stored in our grading system and is easily extracted along with the LaTeX files. 
The primary task is determining the answer types. 
Problems requiring definitive responses are largely categorized into two main classes: atomic and compound. 
Questions with a single required answer fall into the atomic type, while questions with multiple answers are classified as compound, represented by a list of atomic answers separated by commas. 
The atomic type can be further classified into eight types, and the compound answer lists can be either ordered or unordered, with each atomic answer fitting one of the aforementioned eight types.
Simple examples of each type are provided in Table~\ref{tab:examples of answer types}, and detailed definitions are available in Appendix~\ref{app: answer_type}.

\subsection{Evaluation Metrics}\label{sec: metric}

We denote the set of test examples in {\benchmark} by $\mathcal{D}$ with a specific test example denoted as $e_i$, where $i$ represents the index of the example.
Each example $e_i$ consists of questions presented in different randomized versions: $q_i^1, q_i^2,..., q_i^V$, where $V$ is the total number of versions\footnote{Currently, V=3 and we plan to release more versions in the future.}. 
The corresponding ground-truth answers for these versions $q_i^1, q_i^2, ..., q_i^V$ are denoted by $a_i^1, a_i^2, ..., a_i^V$.
The answer generated by an LLM $\mathcal{M}$ for a specific version of the question in the $i$-th test example is denoted by $\mathcal{M}(q_i^v)$.
Inspired by \citet{MATH_FUNC2024srivastava}, we define the following metrics to evaluate the true mathematical reasoning ability of LLM $\mathcal{M}$ in {\benchmark}.

\textbf{Accuracy of Version $\mathbf{v}$ $\text{Acc}_\mathbf{v}$} is defined as the average accuracy of model $\mathcal{M}$ on the set of questions with version $v$ in $\mathcal{D}$:
\begin{equation*}
\text{Acc}_v =\frac{\sum_{i = 1}^{|\mathcal{D}|} \mathbb{I}[\mathcal{M}(q_i^v)=a_i^v]}{|\mathcal{D}|},
\end{equation*}
where $\mathbb{I}$ is an indicator function and $|\mathcal{D}|$ denotes the number of examples in {\benchmark}. 
It assesses the performance of an LLM on the specific version $v$ from {\benchmark}.

\textbf{Average Accuracy AAcc} is defined as the mean of all $\text{Acc}_v$:
\begin{equation*}
    \text{AAcc} = \frac{\sum_{v=1}^V \text{Acc}_v}{V}.
\end{equation*}
This metric evaluates the performance across all versions of the questions.

\textbf{Effective Accuracy EAcc} is defined as the accuracy in solving a test example $e_i$ across all its $V$ versions:
\begin{equation*}
    \text{EAcc} = \frac{\sum_{i = 1}^{|\mathcal{D}|} \mathbb{I}[\mathcal{M}(q_i^v)=a_i^v, \forall v \in \{1, 2, ..., V\}]}{|\mathcal{D}|}.
\end{equation*}
If a model is able to solve a test case using proper reasoning, it should correctly solve this problem for all randomized versions. 
Thus, effective accuracy measures the fraction of test cases correctly solved across all versions $V$.
It measures true reasoning of test cases in {\benchmark}.

\textbf{Reasoning Gap $\Delta$} is defined as the percentage decrease between $\text{AAcc}$ and $\text{EAcc}$. 
It provides a measure of the robustness of reasoning, with $\Delta = 0$ being true reasoning with high robustness. 

\textbf{Robustness Efficiency (RE)} is defined as the ratio of the Reasoning Gap ($\Delta$) to the EAcc, expressed as $\text{RE} = \Delta/\text{EAcc}$.
This metric evaluates the extent of the reasoning gap relative to the model's effective reasoning ability (i.e., EAcc).
RE captures robustness by taking the effectiveness of mathematical reasoning into account , with lower values indicating superior performance in adapting to variations across different versions of problems in {\benchmark}.
Achieving a higher EAcc and a lower $\Delta$ results in a more favorable (lower) RE, reflecting improved robustness relative to "true" reasoning ablility of LLMs.

\section{Experiments}\label{sec: exp}

\subsection{Experimental Setup}\label{sec:exp_setup}

\textbf{Evaluated LLMs}. 
Our evaluation covers 23 leading LLMs, including closed-source commercial LLMs and open-source LLMs.
Based on our {\benchmark}, we provide a thorough evaluation of the mathematical reasoning capabilities of current LLMs. 
The evaluated LLMs are listed below:
\begin{itemize}
    \item For proprietary LLMs, we select OpenAI-o1-mini \citep{o1}, GPT4o \citep{gpt4o2024openai}, GPT4o-mini \citep{gpt4o2024openai}, and Claude-3-Opus \citep{claude3}.
    \item For open-source general-purpose LLMs, we evaluated the LLaMA-3-Instruct series (8B, 70B) \citep{llama3modelcard}, Qwen2-Instruct (7B, 72B)\citep{Qwen22024yang}, Yi-1.5-Chat (6B, 9B, 34B) \citep{ai2024yi}, Mistral-7B-Instruct \citep{mistral2023Jiang}, Mistral-Nemo-Instruct-2407 \citep{mistral_nemo_blog}, Mistral-Small-Instruct-2409 \citep{mistral_small_blog}, Mistral-Large-Instruct-2407 \citep{mistral_large_blog}, DeepSeek-MOE-16B-Chat \citep{dai2024deepseekmoe}, and DeepSeek-V2-Lite-Chat \citep{deepseekv2}.
    \item We also include some specialized math LLMs: DeepSeekMath-7B (-RL, -Instruct) \citep{deepseekmath2024shao}, Qwen2-Math (7B, 72B)\citep{Qwen25Math2024Yang}, Mathstral-7B \citep{mathstral2023}, and NuminaMath-7B-CoT \citep{numinamath7b}.
\end{itemize}
Details of these LLMs are provided in Appendix~\ref{app: models}.

\textbf{Evaluation Settings}.
We employ $\text{Acc}_v$ to evaluate the average performance of version $v$, AAcc to measure the average performance across all versions, EAcc to quantify true reasoning, and reasoning gap $\Delta$ to assess the robustness of reasoning (see Section~\ref{sec: metric}). 
To remove the effect of sensitivity of few-shot prompts \citep{rm0cot2021lu, fairness2023ma}, all our experiments use zero-shot prompts, tailored to different answer types for better answer extraction and rule-based matching.
Detailed prompts are given in Appendix~\ref{app: prompts}.
We use vLLM\footnote{\href{https://github.com/vllm-project/vllm}{https://github.com/vllm-project/vllm}} to speed up the evaluation process.
To maintain consistency in evaluations and facilitate reproduction, we set the maximum output length to 2,048 tokens and employ a greedy decoding strategy with temperature 0.

\subsection{Main Results}

\begin{table*}[!thb]
\centering
\footnotesize
\caption{\textbf{Main Results on {\benchmark}} (all figures are in \%). Models are classified into three categories according to their purpose and origin. The best results within each column are \textbf{bolded} and the best results of open-source Chat LLMs within a similar parameter size group are \underline{underlined}.}
\scalebox{1.0}{
\begin{tabular}{lccccccc}
\hline
\textbf{Models} & \textbf{$\text{Acc}_1$} & \textbf{$\text{Acc}_2$} & \textbf{$\text{Acc}_3$} & \textbf{AAcc} & \textbf{EAcc} & \textbf{$\Delta$} & \textbf{RE}\\
\hline
\multicolumn{7}{c}{{\textit{Closed-source LLMs}}} \\
\cdashline{1-8}
OpenAI-o1-mini-2024-09-12 & \textbf{68.02} & \textbf{67.95} & \textbf{68.04} & \textbf{68.00} & \textbf{56.30} & 11.70 & \textbf{20.78} \\
GPT-4o-2024-08-06 & 59.92 & 60.79 & 60.41 & 60.37 & 49.94 & 10.43 & 20.89 \\
GPT-4o-mini-2024-07-18 & 51.58 & 53.14 & 52.61 & 52.44 & 41.01 & 11.43 & 27.87 \\
Claude-3-Opus-20240229 & 48.62 & 50.32 & 49.47 & 49.47 & 37.00& 12.47 & 33.69 \\
\hline
\multicolumn{7}{c}{{\textit{Open-source Chat LLMs}}} \\
\cdashline{1-8}
Yi-1.5-6B-Chat & 25.33 & 26.83 & 26.59 & 26.25 & 15.23 & 11.02 & 72.33 \\
Mistral-7B-Instruct & 10.19 & 11.16 & 10.33 & 10.56 & 4.44 & \underline{6.12} & 137.6\\
Qwen2-7B-Instruct & \underline{35.60} & \underline{37.30} & \underline{36.23} & \underline{36.38} & \underline{25.15} & 11.23 & \underline{44.65} \\
LLaMA3-8B-Instruct & 16.00 & 17.01 & 16.63 & 16.55 & 8.91 & 7.64 & 85.74 \\
Yi-1.5-9B-Chat & 33.72 & 34.29 & 34.85 & 34.29 & 21.12 & 13.17 & 62.36 \\
\hline
Mistral-Nemo-Instruct-2407 & 24.53 & 25.62 & 25.09 & 25.08 & 15.43 & 9.65 & 62.57 \\ 
DeepSeek-MOE-16B-Chat & 5.59 & 5.85 & 5.97 & 5.80 & 1.96 & \underline{\textbf{3.85}} & 196.6 \\
DeepSeek-V2-Lite-Chat & 12.82 & 13.67 & 12.76 & 13.08 & 5.69 & 7.39 & 130.0 \\
Mistral-Small-Instruct-2409 & \underline{40.10} & \underline{40.52} & \underline{40.04} & \underline{40.22} & \underline{28.84} & 11.38 & \underline{39.45} \\ 
Yi-1.5-34B-Chat & 37.08 & 38.11 & 37.65 & 37.61 & 24.34 & 13.28 & 54.55 \\
\hline
LLaMA3-70B-Instruct & 33.25 & 34.35 & 33.27 & 33.62 & 23.27 & \underline{10.35} & 44.48\\
Qwen2-72B-Instruct & 47.49 & 48.56 & 47.23 & 47.76 & 35.78 & 11.98 & 33.50 \\
Mistral-Large-Instruct-2407 & \underline{55.91} & \underline{56.16} & \underline{55.97} & \underline{56.01} & \underline{45.04} & 10.97 & \underline{24.36} \\
\hline
\multicolumn{7}{c}{{\textit{Specialized Mathematical LLMs}}} \\
\cdashline{1-8}
DeepSeek-Math-7B-Instruct & 23.61 & 24.87 & 23.19 & 23.89 & 13.61 & 10.28 & 75.52 \\
DeepSeek-Math-7B-RL & 28.66 & 29.97 & 28.94 & 29.19 & 19.24 & 9.95 & 51.71 \\
NuminaMath-7B-CoT & 29.32 & 30.07 & 30.01 & 29.80 & 18.81 & 10.99 & 58.44 \\
Mathstral-7B-v0.1 & 28.57 & 28.47 & 28.51 & 28.51 & 17.94 & 10.58 & 58.96 \\
Qwen2-Math-7B-Instruct & 43.01 & 44.13 & 44.05 & 43.73 & 32.46 & 11.27 & 34.73 \\
Qwen2-Math-72B-Instruct & 56.95 & 57.05 & 57.09 & 57.03 & 45.85 & 11.18 & 24.39 \\
\hline
\end{tabular}
}
\label{tab: main_results}
\end{table*}

The overall experiment results are shown in table~\ref{tab: main_results}.
We have the following key observations:

\textbf{{\benchmark} is a challenging benchmark for evaluating the mathematical reasoning capabilities of LLMs.} 
Even LLMs with the most advanced reasoning abilities, OpenAI-o1 (mini version), achieve only 56.3\% EAcc on {\benchmark}, while most open-source LLMs, including most specialized mathematical models, struggle to reach a 30\% EAcc. 
Compared to commonly used mathematics benchmarks like MATH \citep{MATH2021hendrycks}, {\benchmark} proves to be more challenging.
For instance, OpenAI-o1-mini achieves 90\% on MATH (v.s. 56.3\% on {\benchmark}).

\textbf{Even leading LLMs still have inconsistencies when solving problems with multiple versions}.
LLMs with an AAcc greater than 20\% display a reasoning gap $\Delta$ exceeding (or near) 10\%.
All LLMs demonstrate extremely high RE on {\benchmark}, with values ranging from 20.78\% to 196.6\%.
Among the five models with the lowest RE, three of them are from OpenAI (OpenAI-o1-mini: 20.78\%; GPT-4o: 20.89\%; Mistral-Large-Instruct: 24.36\%; Qwen2-Math-72B-Instruct: 24.39\%; GPT-4o-mini: 27.87\%).
These results pinpoint the limitaions of current LLMs and urge us to develop ture "large reasoning models" with high EAcc and $\Delta = 0$.


\textbf{There remains a significant discrepancy between closed-source models and open-source LLMs.}
OpenAI-o1-mini achieves the best results across $\text{Acc}_i, i = 1, 2, 3$, as well as in AAcc and EAcc.
However, the most powerful open-source LLM evaluated, Qwen2-Math-72B-Instruct, still falls short: it shows a 10.97\% lower AAcc and a 10.45\% lower EAcc compared to OpenAI-o1-mini. 
The best-performing open-source chat model is Mistral-Large-Instruct-2407, which ranks second among all open-source LLMs. 
Only 2 out of 19 open-source LLMs exceed GPT-4o-mini in terms of EAcc, and only 3 out of 19 have an EAcc comparable to Claude-3-Opus. 
Additionally, more than half of the open-source LLMs (10 out of 19) have an EAcc smaller than 20\%.

\section{Analysis}


In this section, we conduct an in-depth analysis of the performance of the 23 LLMs evaluated on {\benchmark} by investigating the following research questions:
1. What is the relationship between EAcc and $\text{Acc}_v$? (Section~\ref{sec:analysis:relation})
2. How do model size and model series influence performance on {\benchmark}? (Section~\ref{analysis: model_size})
3.  How do LLMs perform across different subjects, difficulty levels, \revise{and, different topics} on {\benchmark}? (Section~\ref{analysis:subjects})
4. What are the typical response errors made by the best-performing LLM (OpenAI-o1-mini), and how are they distributed? (Section~\ref{analysis: error})

\subsection{Relationship Between EAcc and $\text{Acc}_v$}\label{sec:analysis:relation}

\begin{figure}
    \centering
    \includegraphics[width=1\linewidth]{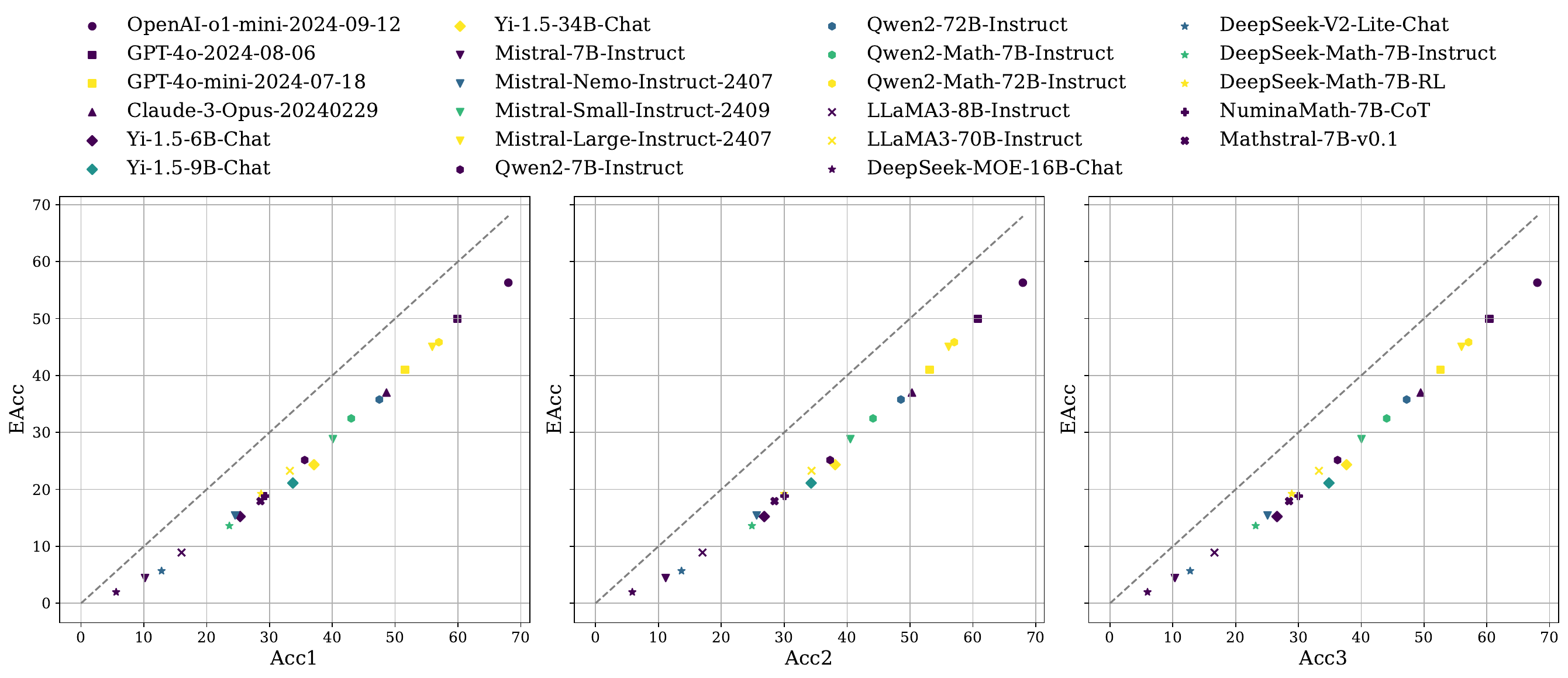}
    \caption{EAcc v.s. $\text{Acc}_v$ on {\benchmark}.}
    \label{fig:eacc_acc}
\end{figure}

To investigate the relationship between EAcc and $\text{Acc}_v$, scatter plots of EAcc against each $\text{Acc}_v$ are shown in Figure~\ref{fig:eacc_acc}.
From Figure~\ref{fig:eacc_acc}, we have the following conclusions:

\textbf{All LLMs fall below the diagonal lines}. Each LLM evaluated is represented as a point in the subfigures of Figure~\ref{fig:eacc_acc}, plotted on the axes of ($\text{Acc}_v, \text{EAcc})$.
Although LLMs exhibit small variations in accuracy across different versions, they consistently demonstrate a lower EAcc than $\text{Acc}_v$, which suggests that the accuracy of individual versions is insufficient for assessing the reasoning capabilities of LLMs. 
By considering EAcc alongside accuracy, we can gain a better understanding of how LLMs perform when solving problems that have different randomized versions. 
The discrepancy between EAcc and $\text{Acc}_v$ highlights a new inconsistency mode \citep{llm4math2024Ahn} of current LLMs: they may become inconsistent in their answers when the problem is slightly altered.

\textbf{There is an apparent trend that a high EAcc consistently leads to a high $\text{Acc}_v$}. 
A model with high EAcc is more effective in handling variable disturbances, resulting in high accuracy for each version. 
However, as EAcc becomes increasingly large, the difference between $\text{Acc}_v$ and EAcc tends to increase until it stabilizes around 10\%.

\subsection{The Effect of Model Size and Model Series}\label{analysis: model_size}

Figure~\ref{fig: model_size_and_family} has shown how EAcc and 2-RE changes with the parameter size.
We can observe that:

\begin{figure}
    \centering
    \includegraphics[width=1\linewidth]{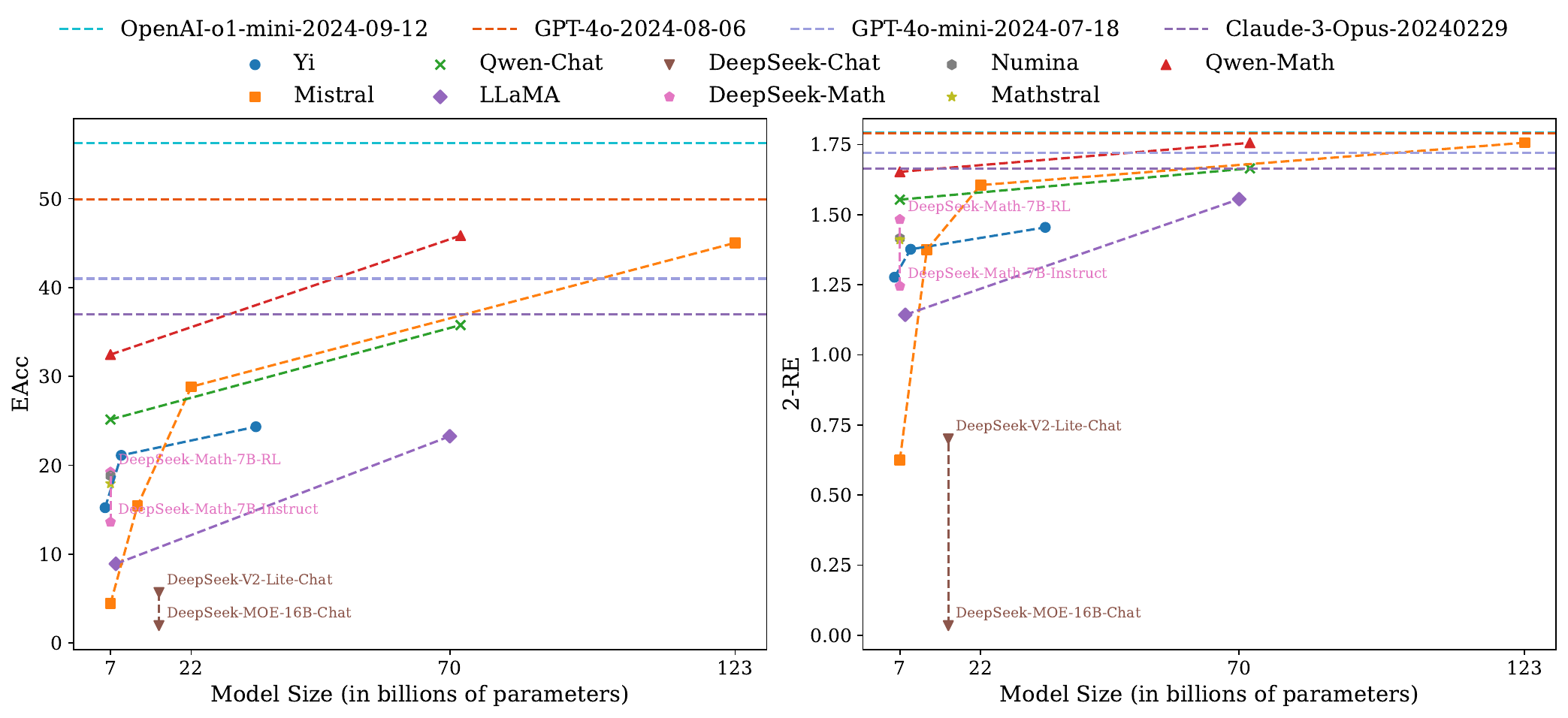}
    \caption{Left: EAcc v.s. Model Size. Right: 2-RE v.s. Model Size.
    The comparison chart of performance versus performance (EAcc and RE) on {\benchmark} for all LLMs evaluated, with models from the same series connected by lines of the same color. 
    The horizontal dotted lines represent the score of close-source LLMs.}
    \label{fig: model_size_and_family}
\end{figure}

\textbf{LLMs within the same series have shown steady improvement as the parameter size increases}.
When the model size increases from 7B to around 100B, EAcc substantially improve and RE steadily decrease for Qwen-Chat, Qwen-Math, Mistral, Deepseek-Chat, LLaMA-3-Instruct, and Yi-Chat series, indicating a steady improvement in performance in effectiveness and robustness of mathematical reasoning.

\textbf{Specialized mathematical LLMs typically outperform their general-purpose counterparts}.
For example, the Qwen2-Math series achieves significantly higher EAcc and lower RE than its general-purpose chat LLMs with the same model size. 
Among all 7B specialized mathematical LLMs, Qwen2-Math-7B-Instruct ranks first surpassing DeepSeek-Math-RL (second best) by a large margin.


\subsection{Performance Across Different Subjects and Difficulty Levels}\label{analysis:subjects}
\begin{figure}[ht]
    \centering
    \begin{subfigure}[t]{0.48\textwidth} 
        \centering
        \includegraphics[width=\textwidth]{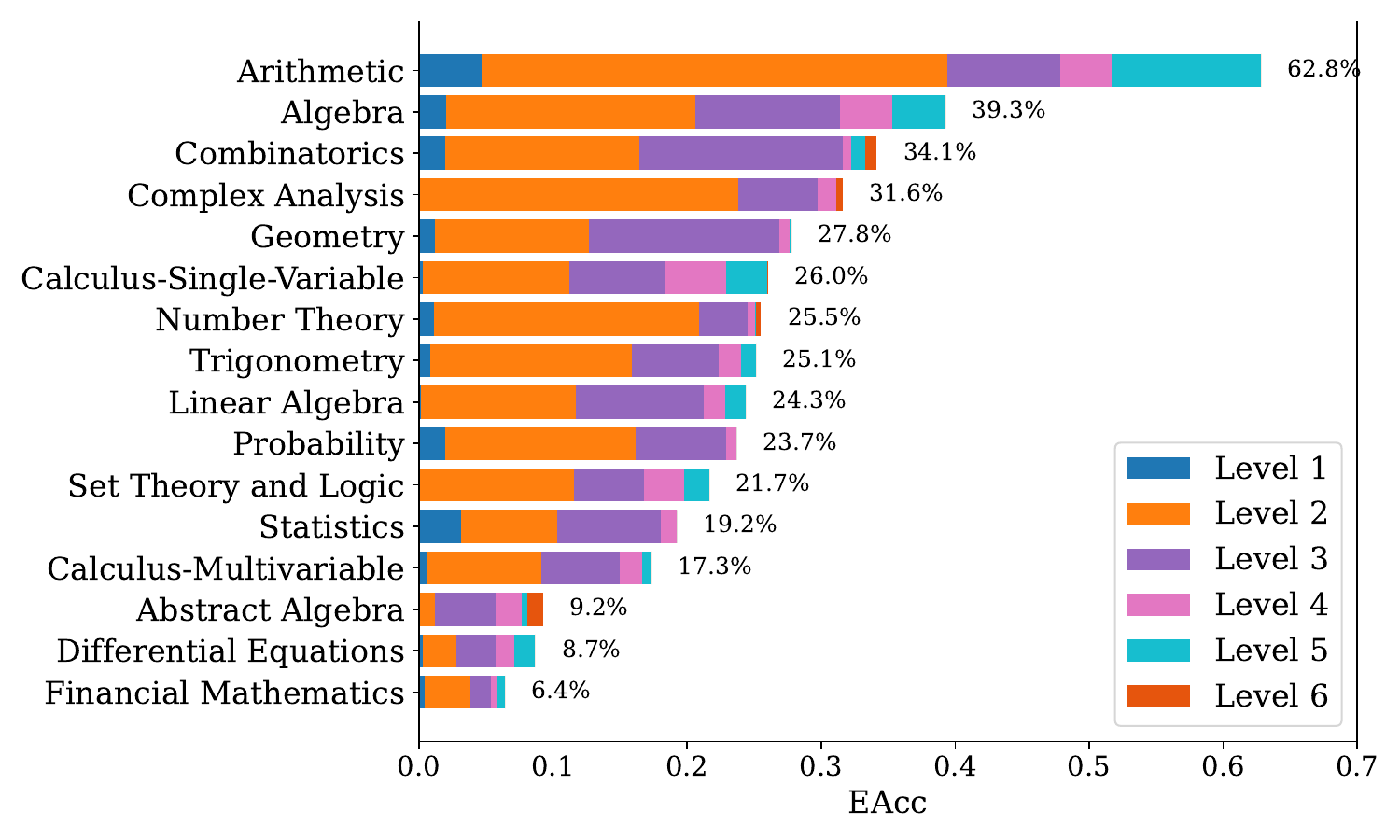}
        \caption{EAcc across subjects}
        \label{fig:acc_subject}
    \end{subfigure}
    \begin{subfigure}[t]{0.48\textwidth} 
        \centering
        \raisebox{1.5em}{ 
            \includegraphics[width=\textwidth]{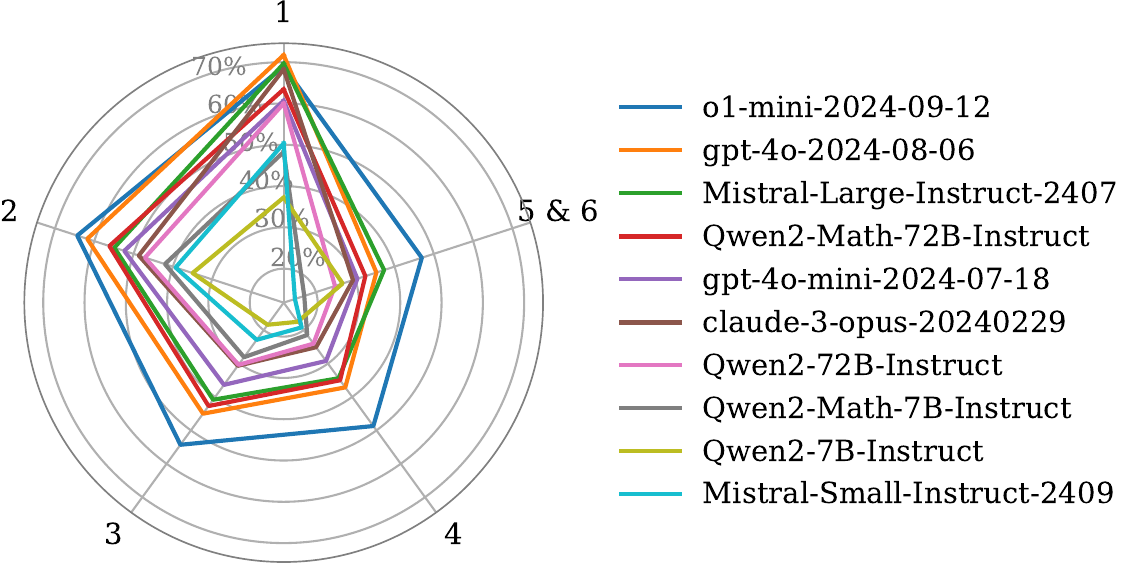}
        }
        \caption{EAcc across levels}
        \label{fig:acc_level}
    \end{subfigure}
    \caption{Relationship between EAcc, subject, and level of difficulty. (a) EAcc of different subjects, averaged across all models. Each bar consists of several segments with colors indicating their corresponding difficulty level. Notice that the length of each color segment only indicates its proportion within all problems of all levels within that subject, and is not comparable between levels or subjects.  (b) EAcc of different levels, averaged across all subjects. Only models with top-10 EAcc are included for brevity. Levels 5 and 6 are combined since level 6 has few samples.}
    \label{fig:acc_subject_level}
\end{figure}

Figure \ref{fig:acc_subject} shows the average EAcc of different models in each subject. Figure \ref{fig:acc_level} shows the averaged EAcc of all subjects with respect to different levels of difficulty.
\revise{The detailed performances across different subjects and topics for each model can be found in Appendix~\ref{app:subjects} and \ref{app:topics}.}

\textbf{The average EAcc varies across different subjects}. 
LLMs are effective at solving Arithmetic problems, achieving 62.8\% EAcc. 
In addition to Arithmetic, LLMs are also adept at Algebra, Combinatorics, and Complex Analysis (over 30\% average EAcc). The three least effective areas are Abstract Algebra, Differential Equations, and Financial Mathematics, which typically require challenging domain knowledge (less than 10\% average EAcc).

\textbf{In general, the averaged EAcc decreases as the level increases.} OpenAI's o1-mini is the strongest among all models and wins by a larger margin as the level increases. 
Mistral-Large/Small-Instruct and Qwen2 series are the most competitive open-source models on our benchmark, with EAcc comparable to leading commercial models such as GPT-4o.
However, as the level of difficulty increases, they still lag behind GPT-4o, suggesting the gap between the leading open-source LLMs and proprietary LLMs in solving difficult math problems.

\subsection{Error Analysis}\label{analysis: error}

We perform a comprehensive error analysis on OpenAI-o1-mini by randomly selecting 100 problems, each having at least one incorrectly solved version (yielding a total of 300 versions). As shown in Figure 5a, there are 231 incorrect versions, and OpenAI-o1-mini failed to solve 56\% of the problems across all versions. We then categorize these errors into six types, as illustrated in Figure 5b. Calculation errors, including both numerical and expression errors, represent the largest category, with several examples provided in Appendix~\ref{app: error}. 
We find that OpenAI-o1-mini tends to streamline its outputs to avoid generating too long responses, sometimes leading to erroneous results. 
Additionally, we encounter some "bad questions" that primarily arise due to overly complex structures (e.g. containing long tables) or inadequately
described (e.g. undefined variables in previous problems).
This is because our homework grading system (see Appendix~\ref{app: data_source}) is designed for students with a user-friendly interface, and some problems may not be suitable for LLM to solve.
\revise{In our sample of 300 versions, 19 were identified as "bad problems," giving us an estimated occurrence of approximately 2.7\% in our {\benchmark}. 
These problems do not impact our main claims, as no LLMs are able to solve these "bad questions."}
We will refine these types of problems to make them more suitable for LLM evaluation in the future.
No answer cleaning process is perfect, and improved evaluation codes continue to be released in MATH \citep{MATH2021hendrycks}.
We will also actively update our evaluation repository to improve its quality.
Inspired by recent model-based evaluation efforts~\citep{gao2024omni, xu2025ugphysics}, we have updated our code repository to integrate the model-based evaluation introduced by \citet{xu2025ugphysics}. 
Updated results will be made available there.

\begin{figure}[ht]
    \centering
    \begin{subfigure}[t]{0.4\textwidth} 
        \centering
        \includegraphics[width=\textwidth]{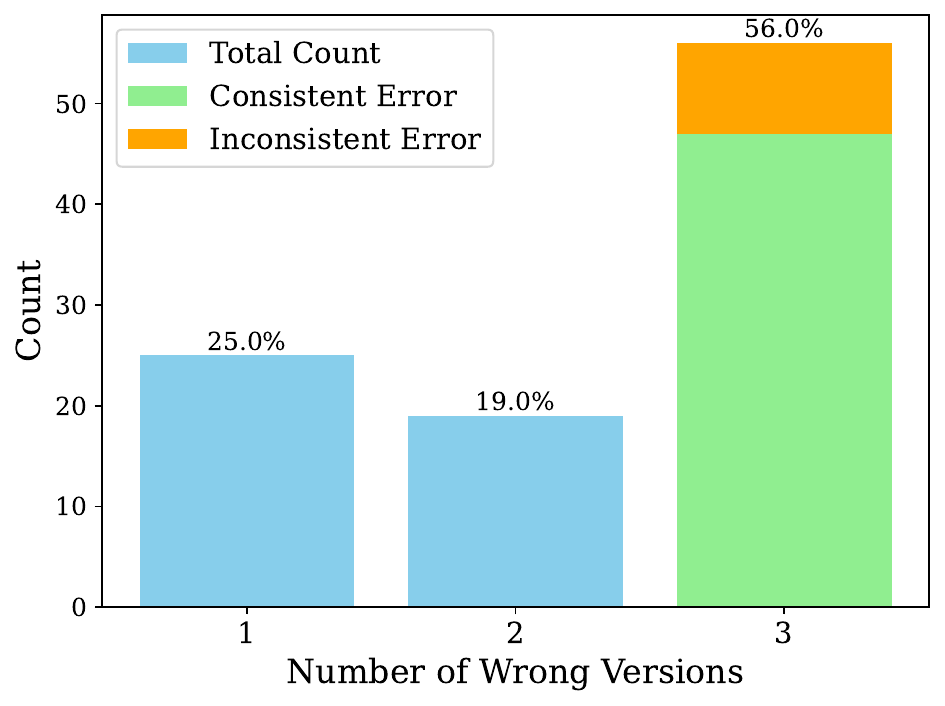}
        \caption{Count vs Number of Wrong Versions}
        \label{fig:error_bar}
    \end{subfigure}
    \begin{subfigure}[t]{0.56\textwidth} 
        \centering
        \raisebox{1.4em}{ 
            \includegraphics[width=\textwidth]{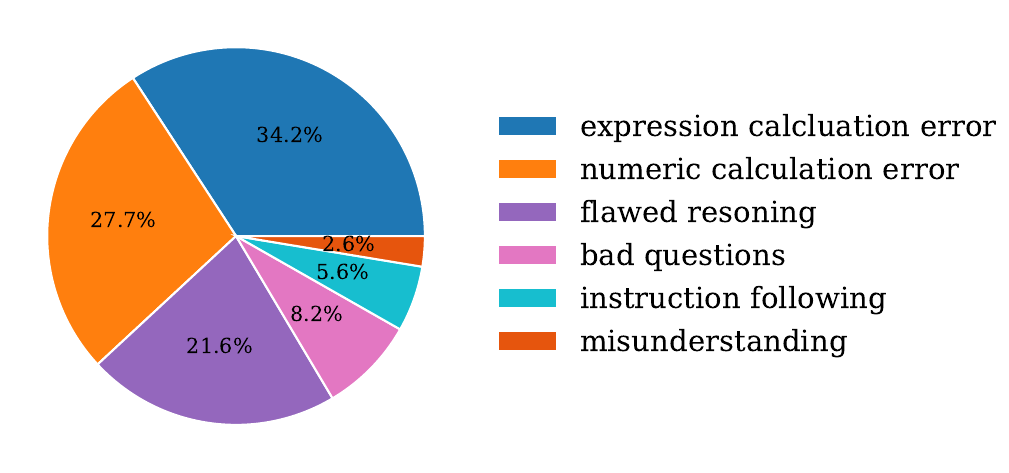}
        }
        \caption{Error Type Distribution}
        \label{fig:error_pie}
    \end{subfigure}
    \caption{Error Analysis of OpenAI-o1-mini on {\benchmark}.}
    \label{fig:error_bar_pie}
\end{figure}

Notably, we have found that even when OpenAI-o1-mini solves a problem incorrectly among all its randomized versions, the error types can be different. 
As shown in Figure~\ref{fig:error_bar}, there are around 16.1\% such inconsistent errors among the 100 problems sampled.

\subsection{\revise{Further Analysis}}\label{sec:php & contamination}

\revise{\textbf{About Self-Improvement}. To examine how LLMs perform with refinement on {\benchmark}, we conducted experiments using Progressive-Hint Prompting (PHP) \citep{PHP2023Zheng} (see Appendix~\ref{app: php}). 
Detailed results can be found in Table~\ref{tabapp: php} in Appendix~\ref{app: php}. 
Although PHP improves EAcc and AAcc for most LLMs, the enhancements are not significant, indicating considerable room for future development. 
The fine-grained results in Table~\ref{tabapp:php_gpt4o} suggest that the impact of refinement for GPT-4o varies across different subjects. 
For instance, PHP improves GPT-4o's performance in abstract algebra by 7.14\%, yet reduces its performance in probability by 2.08\% in terms of EAcc. 
Our {\benchmark} serves as an excellent testing ground for future research into refinement methods for solving undergraduate-level mathematics with LLMs.}

\begin{wraptable}{r}{0.31\textwidth}
\centering
\footnotesize
\caption{Effects of Overfitting.}
\resizebox{0.31\textwidth}{!}{
\begin{tabular}{ccccc}
\toprule
Proportion  & 5\% & 10\% & 15\% & 20\% \\
$\Delta$ & 3.43 & 3.80 & 4.50 & 4.64 \\
\bottomrule
\end{tabular}
}
\label{tab:contamination}
\end{wraptable}

\revise{\textbf{Reasoning Gap and Test Set Contamination}.
To explore how models specifically overfitting to a particular variation affect the reasoning gap, we mixed a portion of the test set from one version with MetaMathQA \citep{metamath2023yu} and then conducted supervised fine-tuning (SFT) Llama-3-8B on this data. 
Details of the SFT process are provided in Appendix~\ref{app: test_set_contamination}, and the results are presented in Table~\ref{tab:contamination}. 
As the proportion of the test set included in the training data increases, the reasoning gap ($\Delta$) also becomes more pronounced. 
This study serves as an initial investigation into test set contamination during the SFT stage.
It's worth noting that contamination at the pre-training stage is also a significant area of interest \citep{contamination22022razeghi, contamination12024jiang}.}



\section{Conclusion}\label{sec: conclusion}

Current mathematical benchmarks are often inadequate, lacking comprehensive coverage of undergraduate-level math problems or being susceptible to test-set contamination.
To fill these gaps, we propose {\benchmark}, a diverse and dynamic benchmark for undergraduate-level mathematical reasoning.
Our fine-grained analysis has pointed out the potential inconsistencies when LLMs encounter problems with slightly different versions.
We hope that our {\benchmark} can contribute to future development of "true" reasoning LLMs.


\section*{Acknowledgments}
This work was partially supported by a grant from the Research Grants Council of the Hong Kong Special Administrative Region, China (Project Reference Number: AoE/E-601/24-N).
We would like to thank all reviewers for their helpful suggestions in improving this paper.


\bibliography{iclr2025_conference}

\begin{thebibliography}{83}
\providecommand{\natexlab}[1]{#1}
\providecommand{\url}[1]{\texttt{#1}}
\expandafter\ifx\csname urlstyle\endcsname\relax
  \providecommand{\doi}[1]{doi: #1}\else
  \providecommand{\doi}{doi: \begingroup \urlstyle{rm}\Url}\fi

\bibitem[Ahn et~al.(2024)Ahn, Verma, Lou, Liu, Zhang, and Yin]{llm4math2024Ahn}
Janice Ahn, Rishu Verma, Renze Lou, Di~Liu, Rui Zhang, and Wenpeng Yin.
\newblock Large language models for mathematical reasoning: Progresses and challenges.
\newblock In Neele Falk, Sara Papi, and Mike Zhang (eds.), \emph{Proceedings of the 18th Conference of the European Chapter of the Association for Computational Linguistics: Student Research Workshop}, pp.\  225--237, St. Julian{'}s, Malta, 2024. Association for Computational Linguistics.
\newblock URL \url{https://aclanthology.org/2024.eacl-srw.17}.

\bibitem[AI et~al.(2024)AI, :, Young, Chen, Li, Huang, Zhang, Zhang, Li, Zhu, Chen, Chang, Yu, Liu, Liu, Yue, Yang, Yang, Yu, Xie, Huang, Hu, Ren, Niu, Nie, Xu, Liu, Wang, Cai, Gu, Liu, and Dai]{ai2024yi}
01. AI, :, Alex Young, Bei Chen, Chao Li, Chengen Huang, Ge~Zhang, Guanwei Zhang, Heng Li, Jiangcheng Zhu, Jianqun Chen, Jing Chang, Kaidong Yu, Peng Liu, Qiang Liu, Shawn Yue, Senbin Yang, Shiming Yang, Tao Yu, Wen Xie, Wenhao Huang, Xiaohui Hu, Xiaoyi Ren, Xinyao Niu, Pengcheng Nie, Yuchi Xu, Yudong Liu, Yue Wang, Yuxuan Cai, Zhenyu Gu, Zhiyuan Liu, and Zonghong Dai.
\newblock Yi: Open foundation models by 01.ai, 2024.

\bibitem[AI@Meta(2024)]{llama3modelcard}
AI@Meta.
\newblock Llama 3 model card.
\newblock 2024.
\newblock URL \url{https://github.com/meta-llama/llama3/blob/main/MODEL_CARD.md}.

\bibitem[Amini et~al.(2019)Amini, Gabriel, Lin, Koncel-Kedziorski, Choi, and Hajishirzi]{mathqa2019amini}
Aida Amini, Saadia Gabriel, Shanchuan Lin, Rik Koncel-Kedziorski, Yejin Choi, and Hannaneh Hajishirzi.
\newblock {M}ath{QA}: Towards interpretable math word problem solving with operation-based formalisms.
\newblock In Jill Burstein, Christy Doran, and Thamar Solorio (eds.), \emph{Proceedings of the 2019 Conference of the North {A}merican Chapter of the Association for Computational Linguistics: Human Language Technologies, Volume 1 (Long and Short Papers)}, pp.\  2357--2367, Minneapolis, Minnesota, 2019. Association for Computational Linguistics.
\newblock \doi{10.18653/v1/N19-1245}.
\newblock URL \url{https://aclanthology.org/N19-1245}.

\bibitem[{Anthropic}(2024)]{claude3}
{Anthropic}.
\newblock Claude 3 family.
\newblock \url{https://www.anthropic.com/news/claude-3-family}, 2024.
\newblock Accessed: 2024-09-23.

\bibitem[Azerbayev et~al.(2023)Azerbayev, Schoelkopf, Paster, Santos, McAleer, Jiang, Deng, Biderman, and Welleck]{llemma2023azerbayev}
Zhangir Azerbayev, Hailey Schoelkopf, Keiran Paster, Marco~Dos Santos, Stephen McAleer, Albert~Q Jiang, Jia Deng, Stella Biderman, and Sean Welleck.
\newblock Llemma: An open language model for mathematics.
\newblock \emph{ArXiv preprint}, abs/2310.10631, 2023.
\newblock URL \url{https://arxiv.org/abs/2310.10631}.

\bibitem[Beeching et~al.(2024)Beeching, Huang, Jiang, Li, Lipkin, Qina, Rasul, Shen, Soletskyi, and Tunstall]{numinamath7b}
Edward Beeching, Shengyi~Costa Huang, Albert Jiang, Jia Li, Benjamin Lipkin, Zihan Qina, Kashif Rasul, Ziju Shen, Roman Soletskyi, and Lewis Tunstall.
\newblock Numinamath 7b cot.
\newblock \url{https://huggingface.co/AI-MO/NuminaMath-7B-CoT}, 2024.

\bibitem[Bobrow et~al.(1964)]{MWP1964bobrow}
Daniel Bobrow et~al.
\newblock Natural language input for a computer problem solving system.
\newblock 1964.

\bibitem[Brown et~al.(2020)Brown, Mann, Ryder, Subbiah, Kaplan, Dhariwal, Neelakantan, Shyam, Sastry, Askell, Agarwal, Herbert{-}Voss, Krueger, Henighan, Child, Ramesh, Ziegler, Wu, Winter, Hesse, Chen, Sigler, Litwin, Gray, Chess, Clark, Berner, McCandlish, Radford, Sutskever, and Amodei]{GPT32020Brown}
Tom~B. Brown, Benjamin Mann, Nick Ryder, Melanie Subbiah, Jared Kaplan, Prafulla Dhariwal, Arvind Neelakantan, Pranav Shyam, Girish Sastry, Amanda Askell, Sandhini Agarwal, Ariel Herbert{-}Voss, Gretchen Krueger, Tom Henighan, Rewon Child, Aditya Ramesh, Daniel~M. Ziegler, Jeffrey Wu, Clemens Winter, Christopher Hesse, Mark Chen, Eric Sigler, Mateusz Litwin, Scott Gray, Benjamin Chess, Jack Clark, Christopher Berner, Sam McCandlish, Alec Radford, Ilya Sutskever, and Dario Amodei.
\newblock Language models are few-shot learners.
\newblock In Hugo Larochelle, Marc'Aurelio Ranzato, Raia Hadsell, Maria{-}Florina Balcan, and Hsuan{-}Tien Lin (eds.), \emph{Advances in Neural Information Processing Systems 33: Annual Conference on Neural Information Processing Systems 2020, NeurIPS 2020, December 6-12, 2020, virtual}, 2020.
\newblock URL \url{https://proceedings.neurips.cc/paper/2020/hash/1457c0d6bfcb4967418bfb8ac142f64a-Abstract.html}.

\bibitem[Chen et~al.(2021)Chen, Tang, Qin, Liang, Liu, Xing, and Lin]{geoqa2021chen}
Jiaqi Chen, Jianheng Tang, Jinghui Qin, Xiaodan Liang, Lingbo Liu, Eric Xing, and Liang Lin.
\newblock {G}eo{QA}: A geometric question answering benchmark towards multimodal numerical reasoning.
\newblock In Chengqing Zong, Fei Xia, Wenjie Li, and Roberto Navigli (eds.), \emph{Findings of the Association for Computational Linguistics: ACL-IJCNLP 2021}, pp.\  513--523, Online, 2021. Association for Computational Linguistics.
\newblock \doi{10.18653/v1/2021.findings-acl.46}.
\newblock URL \url{https://aclanthology.org/2021.findings-acl.46}.

\bibitem[Chen et~al.(2022)Chen, Ma, Wang, and Cohen]{Math23k2022chen}
Wenhu Chen, Xueguang Ma, Xinyi Wang, and William~W Cohen.
\newblock Program of thoughts prompting: Disentangling computation from reasoning for numerical reasoning tasks.
\newblock \emph{ArXiv preprint}, abs/2211.12588, 2022.
\newblock URL \url{https://arxiv.org/abs/2211.12588}.

\bibitem[Chen et~al.(2023)Chen, Yin, Ku, Lu, Wan, Ma, Xu, Wang, and Xia]{theoremqa2023chen}
Wenhu Chen, Ming Yin, Max Ku, Pan Lu, Yixin Wan, Xueguang Ma, Jianyu Xu, Xinyi Wang, and Tony Xia.
\newblock {T}heorem{QA}: A theorem-driven question answering dataset.
\newblock In Houda Bouamor, Juan Pino, and Kalika Bali (eds.), \emph{Proceedings of the 2023 Conference on Empirical Methods in Natural Language Processing}, pp.\  7889--7901, Singapore, 2023. Association for Computational Linguistics.
\newblock \doi{10.18653/v1/2023.emnlp-main.489}.
\newblock URL \url{https://aclanthology.org/2023.emnlp-main.489}.

\bibitem[Chen et~al.(2024)Chen, Chi, Wang, and Zhou]{R-GSM2024chen}
Xinyun Chen, Ryan~A Chi, Xuezhi Wang, and Denny Zhou.
\newblock Premise order matters in reasoning with large language models.
\newblock \emph{ArXiv preprint}, abs/2402.08939, 2024.
\newblock URL \url{https://arxiv.org/abs/2402.08939}.

\bibitem[Cobbe et~al.(2021)Cobbe, Kosaraju, Bavarian, Chen, Jun, Kaiser, Plappert, Tworek, Hilton, Nakano, et~al.]{gsm8k2021cobbe}
Karl Cobbe, Vineet Kosaraju, Mohammad Bavarian, Mark Chen, Heewoo Jun, Lukasz Kaiser, Matthias Plappert, Jerry Tworek, Jacob Hilton, Reiichiro Nakano, et~al.
\newblock Training verifiers to solve math word problems.
\newblock \emph{ArXiv preprint}, abs/2110.14168, 2021.
\newblock URL \url{https://arxiv.org/abs/2110.14168}.

\bibitem[Collins et~al.(2024)Collins, Jiang, Frieder, Wong, Zilka, Bhatt, Lukasiewicz, Wu, Tenenbaum, Hart, et~al.]{collins2024evaluating}
Katherine~M Collins, Albert~Q Jiang, Simon Frieder, Lionel Wong, Miri Zilka, Umang Bhatt, Thomas Lukasiewicz, Yuhuai Wu, Joshua~B Tenenbaum, William Hart, et~al.
\newblock Evaluating language models for mathematics through interactions.
\newblock \emph{Proceedings of the National Academy of Sciences}, 121\penalty0 (24):\penalty0 e2318124121, 2024.

\bibitem[Dai et~al.(2024)Dai, Deng, Zhao, Xu, Gao, Chen, Li, Zeng, Yu, Wu, Xie, Li, Huang, Luo, Ruan, Sui, and Liang]{dai2024deepseekmoe}
Damai Dai, Chengqi Deng, Chenggang Zhao, R.~X. Xu, Huazuo Gao, Deli Chen, Jiashi Li, Wangding Zeng, Xingkai Yu, Y.~Wu, Zhenda Xie, Y.~K. Li, Panpan Huang, Fuli Luo, Chong Ruan, Zhifang Sui, and Wenfeng Liang.
\newblock Deepseekmoe: Towards ultimate expert specialization in mixture-of-experts language models.
\newblock \emph{ArXiv preprint}, abs/2401.06066, 2024.
\newblock URL \url{https://arxiv.org/abs/2401.06066}.

\bibitem[DeepSeek-AI(2024)]{deepseekv2}
DeepSeek-AI.
\newblock Deepseek-v2: A strong, economical, and efficient mixture-of-experts language model, 2024.

\bibitem[Deng et~al.(2024)Deng, Zhao, Tang, Gerstein, and Cohan]{contamination12023deng}
Chunyuan Deng, Yilun Zhao, Xiangru Tang, Mark Gerstein, and Arman Cohan.
\newblock Investigating data contamination in modern benchmarks for large language models.
\newblock In Kevin Duh, Helena Gomez, and Steven Bethard (eds.), \emph{Proceedings of the 2024 Conference of the North American Chapter of the Association for Computational Linguistics: Human Language Technologies (Volume 1: Long Papers)}, pp.\  8706--8719, Mexico City, Mexico, 2024. Association for Computational Linguistics.
\newblock URL \url{https://aclanthology.org/2024.naacl-long.482}.

\bibitem[Dong et~al.(2024)Dong, Jiang, Liu, Jin, and Li]{contamination22024dong}
Yihong Dong, Xue Jiang, Huanyu Liu, Zhi Jin, and Ge~Li.
\newblock Generalization or memorization: Data contamination and trustworthy evaluation for large language models.
\newblock \emph{ArXiv preprint}, abs/2402.15938, 2024.
\newblock URL \url{https://arxiv.org/abs/2402.15938}.

\bibitem[Gao et~al.(2024)Gao, Song, Yang, Cai, Miao, Dong, Li, Ma, Chen, Xu, et~al.]{gao2024omni}
Bofei Gao, Feifan Song, Zhe Yang, Zefan Cai, Yibo Miao, Qingxiu Dong, Lei Li, Chenghao Ma, Liang Chen, Runxin Xu, et~al.
\newblock Omni-math: A universal olympiad level mathematic benchmark for large language models.
\newblock \emph{ArXiv preprint}, abs/2410.07985, 2024.
\newblock URL \url{https://arxiv.org/abs/2410.07985}.

\bibitem[Golchin \& Surdeanu(2023)Golchin and Surdeanu]{contamination32023golchin}
Shahriar Golchin and Mihai Surdeanu.
\newblock Time travel in llms: Tracing data contamination in large language models.
\newblock \emph{ArXiv preprint}, abs/2308.08493, 2023.
\newblock URL \url{https://arxiv.org/abs/2308.08493}.

\bibitem[Gou et~al.(2023)Gou, Shao, Gong, Yang, Huang, Duan, Chen, et~al.]{tora2023Gou}
Zhibin Gou, Zhihong Shao, Yeyun Gong, Yujiu Yang, Minlie Huang, Nan Duan, Weizhu Chen, et~al.
\newblock Tora: A tool-integrated reasoning agent for mathematical problem solving.
\newblock \emph{ArXiv preprint}, abs/2309.17452, 2023.
\newblock URL \url{https://arxiv.org/abs/2309.17452}.

\bibitem[He et~al.(2024{\natexlab{a}})He, Luo, Bai, Hu, Thai, Shen, Hu, Han, Huang, Zhang, et~al.]{OlympiadBench2024He}
Chaoqun He, Renjie Luo, Yuzhuo Bai, Shengding Hu, Zhen~Leng Thai, Junhao Shen, Jinyi Hu, Xu~Han, Yujie Huang, Yuxiang Zhang, et~al.
\newblock Olympiadbench: A challenging benchmark for promoting agi with olympiad-level bilingual multimodal scientific problems.
\newblock \emph{ArXiv preprint}, abs/2402.14008, 2024{\natexlab{a}}.
\newblock URL \url{https://arxiv.org/abs/2402.14008}.

\bibitem[He et~al.(2024{\natexlab{b}})He, Wu, Zhou, Xuan, Liu, Yang, Zhu, and Huang]{cmmu2024he}
Zheqi He, Xinya Wu, Pengfei Zhou, Richeng Xuan, Guang Liu, Xi~Yang, Qiannan Zhu, and Hua Huang.
\newblock Cmmu: A benchmark for chinese multi-modal multi-type question understanding and reasoning.
\newblock \emph{ArXiv preprint}, abs/2401.14011, 2024{\natexlab{b}}.
\newblock URL \url{https://arxiv.org/abs/2401.14011}.

\bibitem[Hendrycks et~al.(2021{\natexlab{a}})Hendrycks, Burns, Basart, Zou, Mazeika, Song, and Steinhardt]{MMLU2020hendrycks}
Dan Hendrycks, Collin Burns, Steven Basart, Andy Zou, Mantas Mazeika, Dawn Song, and Jacob Steinhardt.
\newblock Measuring massive multitask language understanding.
\newblock In \emph{9th International Conference on Learning Representations, {ICLR} 2021, Virtual Event, Austria, May 3-7, 2021}. OpenReview.net, 2021{\natexlab{a}}.
\newblock URL \url{https://openreview.net/forum?id=d7KBjmI3GmQ}.

\bibitem[Hendrycks et~al.(2021{\natexlab{b}})Hendrycks, Burns, Kadavath, Arora, Basart, Tang, Song, and Steinhardt]{MATH2021hendrycks}
Dan Hendrycks, Collin Burns, Saurav Kadavath, Akul Arora, Steven Basart, Eric Tang, Dawn Song, and Jacob Steinhardt.
\newblock Measuring mathematical problem solving with the math dataset.
\newblock \emph{ArXiv preprint}, abs/2103.03874, 2021{\natexlab{b}}.
\newblock URL \url{https://arxiv.org/abs/2103.03874}.

\bibitem[Huang et~al.(2024)Huang, Wang, Xia, Li, Zou, Xu, Fan, Ye, Chern, Ye, et~al.]{OlympicArena2024huang}
Zhen Huang, Zengzhi Wang, Shijie Xia, Xuefeng Li, Haoyang Zou, Ruijie Xu, Run-Ze Fan, Lyumanshan Ye, Ethan Chern, Yixin Ye, et~al.
\newblock Olympicarena: Benchmarking multi-discipline cognitive reasoning for superintelligent ai.
\newblock \emph{ArXiv preprint}, abs/2406.12753, 2024.
\newblock URL \url{https://arxiv.org/abs/2406.12753}.

\bibitem[Jiang et~al.(2023)Jiang, Sablayrolles, Mensch, Bamford, Chaplot, Casas, Bressand, Lengyel, Lample, Saulnier, et~al.]{mistral2023Jiang}
Albert~Q Jiang, Alexandre Sablayrolles, Arthur Mensch, Chris Bamford, Devendra~Singh Chaplot, Diego de~las Casas, Florian Bressand, Gianna Lengyel, Guillaume Lample, Lucile Saulnier, et~al.
\newblock Mistral 7b.
\newblock \emph{ArXiv preprint}, abs/2310.06825, 2023.
\newblock URL \url{https://arxiv.org/abs/2310.06825}.

\bibitem[Jiang et~al.(2024)Jiang, Liu, Zhong, Schaeffer, Ouyang, Han, and Koyejo]{contamination12024jiang}
Minhao Jiang, Ken~Ziyu Liu, Ming Zhong, Rylan Schaeffer, Siru Ouyang, Jiawei Han, and Sanmi Koyejo.
\newblock Investigating data contamination for pre-training language models.
\newblock \emph{ArXiv preprint}, abs/2401.06059, 2024.
\newblock URL \url{https://arxiv.org/abs/2401.06059}.

\bibitem[Kojima et~al.(2022)Kojima, Gu, Reid, Matsuo, and Iwasawa]{zero-shot-CoT2022Kojima}
Takeshi Kojima, Shixiang~Shane Gu, Machel Reid, Yutaka Matsuo, and Yusuke Iwasawa.
\newblock Large language models are zero-shot reasoners.
\newblock In Sanmi Koyejo, S.~Mohamed, A.~Agarwal, Danielle Belgrave, K.~Cho, and A.~Oh (eds.), \emph{Advances in Neural Information Processing Systems 35: Annual Conference on Neural Information Processing Systems 2022, NeurIPS 2022, New Orleans, LA, USA, November 28 - December 9, 2022}, 2022.
\newblock URL \url{http://papers.nips.cc/paper\_files/paper/2022/hash/8bb0d291acd4acf06ef112099c16f326-Abstract-Conference.html}.

\bibitem[Koncel-Kedziorski et~al.(2016)Koncel-Kedziorski, Roy, Amini, Kushman, and Hajishirzi]{mawps2016koncel}
Rik Koncel-Kedziorski, Subhro Roy, Aida Amini, Nate Kushman, and Hannaneh Hajishirzi.
\newblock {MAWPS}: A math word problem repository.
\newblock In Kevin Knight, Ani Nenkova, and Owen Rambow (eds.), \emph{Proceedings of the 2016 Conference of the North {A}merican Chapter of the Association for Computational Linguistics: Human Language Technologies}, pp.\  1152--1157, San Diego, California, 2016. Association for Computational Linguistics.
\newblock \doi{10.18653/v1/N16-1136}.
\newblock URL \url{https://aclanthology.org/N16-1136}.

\bibitem[Lewkowycz et~al.(2022)Lewkowycz, Andreassen, Dohan, Dyer, Michalewski, Ramasesh, Slone, Anil, Schlag, Gutman{-}Solo, Wu, Neyshabur, Gur{-}Ari, and Misra]{minerva202lewkowycz}
Aitor Lewkowycz, Anders Andreassen, David Dohan, Ethan Dyer, Henryk Michalewski, Vinay~V. Ramasesh, Ambrose Slone, Cem Anil, Imanol Schlag, Theo Gutman{-}Solo, Yuhuai Wu, Behnam Neyshabur, Guy Gur{-}Ari, and Vedant Misra.
\newblock Solving quantitative reasoning problems with language models.
\newblock In Sanmi Koyejo, S.~Mohamed, A.~Agarwal, Danielle Belgrave, K.~Cho, and A.~Oh (eds.), \emph{Advances in Neural Information Processing Systems 35: Annual Conference on Neural Information Processing Systems 2022, NeurIPS 2022, New Orleans, LA, USA, November 28 - December 9, 2022}, 2022.
\newblock URL \url{http://papers.nips.cc/paper\_files/paper/2022/hash/18abbeef8cfe9203fdf9053c9c4fe191-Abstract-Conference.html}.

\bibitem[Li et~al.(2024{\natexlab{a}})Li, Wang, Hu, Wei, Zheng, Hu, Zhang, and Peng]{xwinmath2024li}
Chen Li, Weiqi Wang, Jingcheng Hu, Yixuan Wei, Nanning Zheng, Han Hu, Zheng Zhang, and Houwen Peng.
\newblock Common 7b language models already possess strong math capabilities.
\newblock \emph{ArXiv preprint}, abs/2403.04706, 2024{\natexlab{a}}.
\newblock URL \url{https://arxiv.org/abs/2403.04706}.

\bibitem[Li et~al.(2024{\natexlab{b}})Li, Cui, Zhao, Kong, and Bi]{GSM-Plus2024Li}
Qintong Li, Leyang Cui, Xueliang Zhao, Lingpeng Kong, and Wei Bi.
\newblock Gsm-plus: A comprehensive benchmark for evaluating the robustness of llms as mathematical problem solvers.
\newblock \emph{ArXiv preprint}, abs/2402.19255, 2024{\natexlab{b}}.
\newblock URL \url{https://arxiv.org/abs/2402.19255}.

\bibitem[Ling et~al.(2017)Ling, Yogatama, Dyer, and Blunsom]{mwpmethod12017ling}
Wang Ling, Dani Yogatama, Chris Dyer, and Phil Blunsom.
\newblock Program induction by rationale generation: Learning to solve and explain algebraic word problems.
\newblock In Regina Barzilay and Min-Yen Kan (eds.), \emph{Proceedings of the 55th Annual Meeting of the Association for Computational Linguistics (Volume 1: Long Papers)}, pp.\  158--167, Vancouver, Canada, 2017. Association for Computational Linguistics.
\newblock \doi{10.18653/v1/P17-1015}.
\newblock URL \url{https://aclanthology.org/P17-1015}.

\bibitem[Liu et~al.(2024)Liu, Zheng, Qiao, Duan, Fei, Zhou, Zhang, Zhang, Lin, and Chen]{MathBench2024Liu}
Hongwei Liu, Zilong Zheng, Yuxuan Qiao, Haodong Duan, Zhiwei Fei, Fengzhe Zhou, Wenwei Zhang, Songyang Zhang, Dahua Lin, and Kai Chen.
\newblock Mathbench: Evaluating the theory and application proficiency of llms with a hierarchical mathematics benchmark.
\newblock \emph{ArXiv preprint}, abs/2405.12209, 2024.
\newblock URL \url{https://arxiv.org/abs/2405.12209}.

\bibitem[Lu et~al.(2023)Lu, Bansal, Xia, Liu, Li, Hajishirzi, Cheng, Chang, Galley, and Gao]{MathVista2023lu}
Pan Lu, Hritik Bansal, Tony Xia, Jiacheng Liu, Chunyuan Li, Hannaneh Hajishirzi, Hao Cheng, Kai-Wei Chang, Michel Galley, and Jianfeng Gao.
\newblock Mathvista: Evaluating mathematical reasoning of foundation models in visual contexts.
\newblock \emph{ArXiv preprint}, abs/2310.02255, 2023.
\newblock URL \url{https://arxiv.org/abs/2310.02255}.

\bibitem[Lu et~al.(2022)Lu, Bartolo, Moore, Riedel, and Stenetorp]{rm0cot2021lu}
Yao Lu, Max Bartolo, Alastair Moore, Sebastian Riedel, and Pontus Stenetorp.
\newblock Fantastically ordered prompts and where to find them: Overcoming few-shot prompt order sensitivity.
\newblock In Smaranda Muresan, Preslav Nakov, and Aline Villavicencio (eds.), \emph{Proceedings of the 60th Annual Meeting of the Association for Computational Linguistics (Volume 1: Long Papers)}, pp.\  8086--8098, Dublin, Ireland, 2022. Association for Computational Linguistics.
\newblock \doi{10.18653/v1/2022.acl-long.556}.
\newblock URL \url{https://aclanthology.org/2022.acl-long.556}.

\bibitem[Ma et~al.(2023)Ma, Zhang, Bian, Liu, Zhang, Zhao, Zhang, Fu, Hu, and Wu]{fairness2023ma}
Huan Ma, Changqing Zhang, Yatao Bian, Lemao Liu, Zhirui Zhang, Peilin Zhao, Shu Zhang, Huazhu Fu, Qinghua Hu, and Bingzhe Wu.
\newblock Fairness-guided few-shot prompting for large language models.
\newblock In Alice Oh, Tristan Naumann, Amir Globerson, Kate Saenko, Moritz Hardt, and Sergey Levine (eds.), \emph{Advances in Neural Information Processing Systems 36: Annual Conference on Neural Information Processing Systems 2023, NeurIPS 2023, New Orleans, LA, USA, December 10 - 16, 2023}, 2023.
\newblock URL \url{http://papers.nips.cc/paper\_files/paper/2023/hash/8678da90126aa58326b2fc0254b33a8c-Abstract-Conference.html}.

\bibitem[Mistral(2023)]{mathstral2023}
Mistral.
\newblock Mathstral.
\newblock \url{https://mistral.ai/news/mathstral/}, 2023.
\newblock Accessed: 2024-09-23.

\bibitem[Mistral(2024{\natexlab{a}})]{mistral_7b_v0_3_repo}
Mistral.
\newblock Mistral-7b-instruct-v0.3.
\newblock \url{https://huggingface.co/mistralai/Mistral-7B-Instruct-v0.3}, 2024{\natexlab{a}}.

\bibitem[Mistral(2024{\natexlab{b}})]{mistral_large_blog}
Mistral.
\newblock Mistral large 2.
\newblock \url={https://mistral.ai/news/mistral-large-2407/}, 2024{\natexlab{b}}.

\bibitem[Mistral(2024{\natexlab{c}})]{mistral_nemo_blog}
Mistral.
\newblock The future of ai: Trends and predictions.
\newblock \url{https://mistral.ai/news/mistral-nemo/}, 2024{\natexlab{c}}.
\newblock Accessed: September 29, 2024.

\bibitem[Mistral(2024{\natexlab{d}})]{mistral_small_blog}
Mistral.
\newblock mistralai/mistral-small-instruct-2409.
\newblock \url{https://huggingface.co/mistralai/Mistral-Small-Instruct-2409}, 2024{\natexlab{d}}.

\bibitem[OpenAI(2023)]{GPT42023openai}
OpenAI.
\newblock Gpt-4 technical report.
\newblock \emph{ArXiv preprint}, abs/2303.08774, 2023.
\newblock URL \url{https://arxiv.org/abs/2303.08774}.

\bibitem[{OpenAI}(2024{\natexlab{a}})]{gpt4o2024openai}
{OpenAI}.
\newblock Hello gpt-4o.
\newblock \url{https://openai.com/index/hello-gpt-4o/}, 2024{\natexlab{a}}.

\bibitem[{OpenAI}(2024{\natexlab{b}})]{o1}
{OpenAI}.
\newblock Learning to reason with llms.
\newblock \url{https://openai.com/index/learning-to-reason-with-llms/}, 2024{\natexlab{b}}.
\newblock Accessed: 2024-09-23.

\bibitem[Ouyang et~al.(2022)Ouyang, Wu, Jiang, Almeida, Wainwright, Mishkin, Zhang, Agarwal, Slama, Ray, Schulman, Hilton, Kelton, Miller, Simens, Askell, Welinder, Christiano, Leike, and Lowe]{InstructGPT2022Ouyang}
Long Ouyang, Jeffrey Wu, Xu~Jiang, Diogo Almeida, Carroll~L. Wainwright, Pamela Mishkin, Chong Zhang, Sandhini Agarwal, Katarina Slama, Alex Ray, John Schulman, Jacob Hilton, Fraser Kelton, Luke Miller, Maddie Simens, Amanda Askell, Peter Welinder, Paul~F. Christiano, Jan Leike, and Ryan Lowe.
\newblock Training language models to follow instructions with human feedback.
\newblock In Sanmi Koyejo, S.~Mohamed, A.~Agarwal, Danielle Belgrave, K.~Cho, and A.~Oh (eds.), \emph{Advances in Neural Information Processing Systems 35: Annual Conference on Neural Information Processing Systems 2022, NeurIPS 2022, New Orleans, LA, USA, November 28 - December 9, 2022}, 2022.
\newblock URL \url{http://papers.nips.cc/paper\_files/paper/2022/hash/b1efde53be364a73914f58805a001731-Abstract-Conference.html}.

\bibitem[Patel et~al.(2021)Patel, Bhattamishra, and Goyal]{SVAMP2021patel}
Arkil Patel, Satwik Bhattamishra, and Navin Goyal.
\newblock Are {NLP} models really able to solve simple math word problems?
\newblock In Kristina Toutanova, Anna Rumshisky, Luke Zettlemoyer, Dilek Hakkani-Tur, Iz~Beltagy, Steven Bethard, Ryan Cotterell, Tanmoy Chakraborty, and Yichao Zhou (eds.), \emph{Proceedings of the 2021 Conference of the North American Chapter of the Association for Computational Linguistics: Human Language Technologies}, pp.\  2080--2094, Online, 2021. Association for Computational Linguistics.
\newblock \doi{10.18653/v1/2021.naacl-main.168}.
\newblock URL \url{https://aclanthology.org/2021.naacl-main.168}.

\bibitem[Phan et~al.(2025)Phan, Gatti, Han, Li, Hu, Zhang, Shi, Choi, Agrawal, Chopra, et~al.]{phan2025humanity}
Long Phan, Alice Gatti, Ziwen Han, Nathaniel Li, Josephina Hu, Hugh Zhang, Sean Shi, Michael Choi, Anish Agrawal, Arnav Chopra, et~al.
\newblock Humanity's last exam.
\newblock \emph{ArXiv preprint}, abs/2501.14249, 2025.
\newblock URL \url{https://arxiv.org/abs/2501.14249}.

\bibitem[Qian et~al.(2024)Qian, Wan, Tang, Wang, Zhang, Chen, and Yu]{VarBench2024qian}
Kun Qian, Shunji Wan, Claudia Tang, Youzhi Wang, Xuanming Zhang, Maximillian Chen, and Zhou Yu.
\newblock Varbench: Robust language model benchmarking through dynamic variable perturbation.
\newblock \emph{ArXiv preprint}, abs/2406.17681, 2024.
\newblock URL \url{https://arxiv.org/abs/2406.17681}.

\bibitem[Razeghi et~al.(2022)Razeghi, Logan~IV, Gardner, and Singh]{contamination22022razeghi}
Yasaman Razeghi, Robert~L Logan~IV, Matt Gardner, and Sameer Singh.
\newblock Impact of pretraining term frequencies on few-shot reasoning.
\newblock \emph{ArXiv preprint}, abs/2202.07206, 2022.
\newblock URL \url{https://arxiv.org/abs/2202.07206}.

\bibitem[Roberts et~al.(2023)Roberts, Thakur, Herlihy, White, and Dooley]{contamination42023roberts}
Manley Roberts, Himanshu Thakur, Christine Herlihy, Colin White, and Samuel Dooley.
\newblock To the cutoff... and beyond? a longitudinal perspective on llm data contamination.
\newblock In \emph{The Twelfth International Conference on Learning Representations}, 2023.

\bibitem[Shao et~al.(2024)Shao, Wang, Zhu, Xu, Song, Zhang, Li, Wu, and Guo]{deepseekmath2024shao}
Zhihong Shao, Peiyi Wang, Qihao Zhu, Runxin Xu, Junxiao Song, Mingchuan Zhang, YK~Li, Y~Wu, and Daya Guo.
\newblock Deepseekmath: Pushing the limits of mathematical reasoning in open language models.
\newblock \emph{ArXiv preprint}, abs/2402.03300, 2024.
\newblock URL \url{https://arxiv.org/abs/2402.03300}.

\bibitem[Shi et~al.(2023)Shi, Chen, Misra, Scales, Dohan, Chi, Sch{\"{a}}rli, and Zhou]{GSM-IC2023shi}
Freda Shi, Xinyun Chen, Kanishka Misra, Nathan Scales, David Dohan, Ed~H. Chi, Nathanael Sch{\"{a}}rli, and Denny Zhou.
\newblock Large language models can be easily distracted by irrelevant context.
\newblock In Andreas Krause, Emma Brunskill, Kyunghyun Cho, Barbara Engelhardt, Sivan Sabato, and Jonathan Scarlett (eds.), \emph{International Conference on Machine Learning, {ICML} 2023, 23-29 July 2023, Honolulu, Hawaii, {USA}}, volume 202 of \emph{Proceedings of Machine Learning Research}, pp.\  31210--31227. {PMLR}, 2023.
\newblock URL \url{https://proceedings.mlr.press/v202/shi23a.html}.

\bibitem[Srivastava et~al.(2024)Srivastava, PV, Menon, Sukumar, Philipose, Prince, Thomas, et~al.]{MATH_FUNC2024srivastava}
Saurabh Srivastava, Anto PV, Shashank Menon, Ajay Sukumar, Alan Philipose, Stevin Prince, Sooraj Thomas, et~al.
\newblock Functional benchmarks for robust evaluation of reasoning performance, and the reasoning gap.
\newblock \emph{ArXiv preprint}, abs/2402.19450, 2024.
\newblock URL \url{https://arxiv.org/abs/2402.19450}.

\bibitem[Tang et~al.(2024)Tang, Zhang, Wan, and Wei]{CollegeMath2024Tang}
Zhengyang Tang, Xingxing Zhang, Benyou Wan, and Furu Wei.
\newblock Mathscale: Scaling instruction tuning for mathematical reasoning.
\newblock \emph{ArXiv preprint}, abs/2403.02884, 2024.
\newblock URL \url{https://arxiv.org/abs/2403.02884}.

\bibitem[Team et~al.(2023)Team, Anil, Borgeaud, Wu, Alayrac, Yu, Soricut, Schalkwyk, Dai, Hauth, et~al.]{gemini2023team}
Gemini Team, Rohan Anil, Sebastian Borgeaud, Yonghui Wu, Jean-Baptiste Alayrac, Jiahui Yu, Radu Soricut, Johan Schalkwyk, Andrew~M Dai, Anja Hauth, et~al.
\newblock Gemini: a family of highly capable multimodal models.
\newblock \emph{ArXiv preprint}, abs/2312.11805, 2023.
\newblock URL \url{https://arxiv.org/abs/2312.11805}.

\bibitem[Tong et~al.(2024)Tong, Zhang, Wang, Wu, and He]{dartmath2024tong}
Yuxuan Tong, Xiwen Zhang, Rui Wang, Ruidong Wu, and Junxian He.
\newblock Dart-math: Difficulty-aware rejection tuning for mathematical problem-solving.
\newblock \emph{ArXiv preprint}, abs/2407.13690, 2024.
\newblock URL \url{https://arxiv.org/abs/2407.13690}.

\bibitem[Wang et~al.(2023{\natexlab{a}})Wang, Hu, Lu, Zhu, Zhang, Subramaniam, Loomba, Zhang, Sun, and Wang]{SciBench2023Wang}
Xiaoxuan Wang, Ziniu Hu, Pan Lu, Yanqiao Zhu, Jieyu Zhang, Satyen Subramaniam, Arjun~R Loomba, Shichang Zhang, Yizhou Sun, and Wei Wang.
\newblock Scibench: Evaluating college-level scientific problem-solving abilities of large language models.
\newblock \emph{ArXiv preprint}, abs/2307.10635, 2023{\natexlab{a}}.
\newblock URL \url{https://arxiv.org/abs/2307.10635}.

\bibitem[Wang et~al.(2023{\natexlab{b}})Wang, Wei, Schuurmans, Le, Chi, Narang, Chowdhery, and Zhou]{CoT-SC2022Wang}
Xuezhi Wang, Jason Wei, Dale Schuurmans, Quoc~V. Le, Ed~H. Chi, Sharan Narang, Aakanksha Chowdhery, and Denny Zhou.
\newblock Self-consistency improves chain of thought reasoning in language models.
\newblock In \emph{The Eleventh International Conference on Learning Representations, {ICLR} 2023, Kigali, Rwanda, May 1-5, 2023}. OpenReview.net, 2023{\natexlab{b}}.
\newblock URL \url{https://openreview.net/pdf?id=1PL1NIMMrw}.

\bibitem[Wang et~al.(2017)Wang, Liu, and Shi]{mwpmethod2017wang}
Yan Wang, Xiaojiang Liu, and Shuming Shi.
\newblock Deep neural solver for math word problems.
\newblock In Martha Palmer, Rebecca Hwa, and Sebastian Riedel (eds.), \emph{Proceedings of the 2017 Conference on Empirical Methods in Natural Language Processing}, pp.\  845--854, Copenhagen, Denmark, 2017. Association for Computational Linguistics.
\newblock \doi{10.18653/v1/D17-1088}.
\newblock URL \url{https://aclanthology.org/D17-1088}.

\bibitem[Wang et~al.(2024)Wang, Ma, Zhang, Ni, Chandra, Guo, Ren, Arulraj, He, Jiang, et~al.]{MMLU-pro2024Wang}
Yubo Wang, Xueguang Ma, Ge~Zhang, Yuansheng Ni, Abhranil Chandra, Shiguang Guo, Weiming Ren, Aaran Arulraj, Xuan He, Ziyan Jiang, et~al.
\newblock Mmlu-pro: A more robust and challenging multi-task language understanding benchmark.
\newblock \emph{ArXiv preprint}, abs/2406.01574, 2024.
\newblock URL \url{https://arxiv.org/abs/2406.01574}.

\bibitem[Wei et~al.(2022)Wei, Wang, Schuurmans, Bosma, Ichter, Xia, Chi, Le, and Zhou]{CoT2022Wei}
Jason Wei, Xuezhi Wang, Dale Schuurmans, Maarten Bosma, Brian Ichter, Fei Xia, Ed~H. Chi, Quoc~V. Le, and Denny Zhou.
\newblock Chain-of-thought prompting elicits reasoning in large language models.
\newblock In Sanmi Koyejo, S.~Mohamed, A.~Agarwal, Danielle Belgrave, K.~Cho, and A.~Oh (eds.), \emph{Advances in Neural Information Processing Systems 35: Annual Conference on Neural Information Processing Systems 2022, NeurIPS 2022, New Orleans, LA, USA, November 28 - December 9, 2022}, 2022.
\newblock URL \url{http://papers.nips.cc/paper\_files/paper/2022/hash/9d5609613524ecf4f15af0f7b31abca4-Abstract-Conference.html}.

\bibitem[Welbl et~al.(2017)Welbl, Liu, and Gardner]{mwpmethod32017welbl}
Johannes Welbl, Nelson~F. Liu, and Matt Gardner.
\newblock Crowdsourcing multiple choice science questions.
\newblock In Leon Derczynski, Wei Xu, Alan Ritter, and Tim Baldwin (eds.), \emph{Proceedings of the 3rd Workshop on Noisy User-generated Text}, pp.\  94--106, Copenhagen, Denmark, 2017. Association for Computational Linguistics.
\newblock \doi{10.18653/v1/W17-4413}.
\newblock URL \url{https://aclanthology.org/W17-4413}.

\bibitem[White et~al.(2024)White, Dooley, Roberts, Pal, Feuer, Jain, Shwartz-Ziv, Jain, Saifullah, Naidu, et~al.]{LiveBench2024white}
Colin White, Samuel Dooley, Manley Roberts, Arka Pal, Ben Feuer, Siddhartha Jain, Ravid Shwartz-Ziv, Neel Jain, Khalid Saifullah, Siddartha Naidu, et~al.
\newblock Livebench: A challenging, contamination-free llm benchmark.
\newblock \emph{ArXiv preprint}, abs/2406.19314, 2024.
\newblock URL \url{https://arxiv.org/abs/2406.19314}.

\bibitem[Xu et~al.(2024{\natexlab{a}})Xu, Diao, Yang, and Wang]{pds2024xu}
Xin Xu, Shizhe Diao, Can Yang, and Yang Wang.
\newblock Can we verify step by step for incorrect answer detection?
\newblock \emph{ArXiv preprint}, abs/2402.10528, 2024{\natexlab{a}}.
\newblock URL \url{https://arxiv.org/abs/2402.10528}.

\bibitem[Xu et~al.(2024{\natexlab{b}})Xu, Xiao, Chao, Huang, Yang, and Wang]{E-GSM2024Xu}
Xin Xu, Tong Xiao, Zitong Chao, Zhenya Huang, Can Yang, and Yang Wang.
\newblock Can llms solve longer math word problems better?
\newblock \emph{ArXiv preprint}, abs/2405.14804, 2024{\natexlab{b}}.
\newblock URL \url{https://arxiv.org/abs/2405.14804}.

\bibitem[Xu et~al.(2025{\natexlab{a}})Xu, Xu, Xiao, Chen, Yan, Zhang, Diao, Yang, and Wang]{xu2025ugphysics}
Xin Xu, Qiyun Xu, Tong Xiao, Tianhao Chen, Yuchen Yan, Jiaxin Zhang, Shizhe Diao, Can Yang, and Yang Wang.
\newblock Ugphysics: A comprehensive benchmark for undergraduate physics reasoning with large language models.
\newblock \emph{ArXiv preprint}, abs/2502.00334, 2025{\natexlab{a}}.
\newblock URL \url{https://arxiv.org/abs/2502.00334}.

\bibitem[Xu et~al.(2025{\natexlab{b}})Xu, Xu, Chen, Yan, Liu, Chen, Wang, Yin, Wang, Shang, et~al.]{xu2025tata}
Xin Xu, Yan Xu, Tianhao Chen, Yuchen Yan, Chengwu Liu, Zaoyu Chen, Yufei Wang, Yichun Yin, Yasheng Wang, Lifeng Shang, et~al.
\newblock Teaching llms according to their aptitude: Adaptive reasoning for mathematical problem solving.
\newblock \emph{ArXiv preprint}, abs/2502.12022, 2025{\natexlab{b}}.
\newblock URL \url{https://arxiv.org/abs/2502.12022}.

\bibitem[Xue et~al.(2024)Xue, Huang, Liu, Lin, Ning, Jin, Li, and Liu]{xue2024decompose}
Shangzi Xue, Zhenya Huang, Jiayu Liu, Xin Lin, Yuting Ning, Binbin Jin, Xin Li, and Qi~Liu.
\newblock Decompose, analyze and rethink: Solving intricate problems with human-like reasoning cycle.
\newblock In \emph{The Thirty-eighth Annual Conference on Neural Information Processing Systems}, 2024.
\newblock URL \url{https://openreview.net/forum?id=NPKZF1WDjZ}.

\bibitem[Yan et~al.(2024)Yan, Jiang, Liu, Cao, Xu, Cai, Shao, et~al.]{yan2024s}
Yuchen Yan, Jin Jiang, Yang Liu, Yixin Cao, Xin Xu, Xunliang Cai, Jian Shao, et~al.
\newblock S $^{3}$ c-math: Spontaneous step-level self-correction makes large language models better mathematical reasoners.
\newblock \emph{ArXiv preprint}, abs/2409.01524, 2024.
\newblock URL \url{https://arxiv.org/abs/2409.01524}.

\bibitem[Yan et~al.(2025)Yan, Shen, Liu, Jiang, Xu, Zhang, Shao, and Zhuang]{yan2025mathfimer}
Yuchen Yan, Yongliang Shen, Yang Liu, Jin Jiang, Xin Xu, Mengdi Zhang, Jian Shao, and Yueting Zhuang.
\newblock Mathfimer: Enhancing mathematical reasoning by expanding reasoning steps through fill-in-the-middle task.
\newblock \emph{ArXiv preprint}, abs/2502.11684, 2025.
\newblock URL \url{https://arxiv.org/abs/2502.11684}.

\bibitem[Yang et~al.(2024{\natexlab{a}})Yang, Yang, Hui, Zheng, Yu, Zhou, Li, Li, Liu, Huang, et~al.]{Qwen22024yang}
An~Yang, Baosong Yang, Binyuan Hui, Bo~Zheng, Bowen Yu, Chang Zhou, Chengpeng Li, Chengyuan Li, Dayiheng Liu, Fei Huang, et~al.
\newblock Qwen2 technical report.
\newblock \emph{ArXiv preprint}, abs/2407.10671, 2024{\natexlab{a}}.
\newblock URL \url{https://arxiv.org/abs/2407.10671}.

\bibitem[Yang et~al.(2024{\natexlab{b}})Yang, Zhang, Hui, Gao, Yu, Li, Liu, Tu, Zhou, Lin, et~al.]{Qwen25Math2024Yang}
An~Yang, Beichen Zhang, Binyuan Hui, Bofei Gao, Bowen Yu, Chengpeng Li, Dayiheng Liu, Jianhong Tu, Jingren Zhou, Junyang Lin, et~al.
\newblock Qwen2. 5-math technical report: Toward mathematical expert model via self-improvement.
\newblock \emph{ArXiv preprint}, abs/2409.12122, 2024{\natexlab{b}}.
\newblock URL \url{https://arxiv.org/abs/2409.12122}.

\bibitem[Yu et~al.(2023)Yu, Jiang, Shi, Yu, Liu, Zhang, Kwok, Li, Weller, and Liu]{metamath2023yu}
Longhui Yu, Weisen Jiang, Han Shi, Jincheng Yu, Zhengying Liu, Yu~Zhang, James~T Kwok, Zhenguo Li, Adrian Weller, and Weiyang Liu.
\newblock Metamath: Bootstrap your own mathematical questions for large language models.
\newblock \emph{ArXiv preprint}, abs/2309.12284, 2023.
\newblock URL \url{https://arxiv.org/abs/2309.12284}.

\bibitem[Yue et~al.(2023)Yue, Qu, Zhang, Fu, Huang, Sun, Su, and Chen]{mammoth2023yue}
Xiang Yue, Xingwei Qu, Ge~Zhang, Yao Fu, Wenhao Huang, Huan Sun, Yu~Su, and Wenhu Chen.
\newblock Mammoth: Building math generalist models through hybrid instruction tuning.
\newblock \emph{ArXiv preprint}, abs/2309.05653, 2023.
\newblock URL \url{https://arxiv.org/abs/2309.05653}.

\bibitem[Yue et~al.(2024{\natexlab{a}})Yue, Ni, Zhang, Zheng, Liu, Zhang, Stevens, Jiang, Ren, Sun, et~al.]{MMMU2024yue}
Xiang Yue, Yuansheng Ni, Kai Zhang, Tianyu Zheng, Ruoqi Liu, Ge~Zhang, Samuel Stevens, Dongfu Jiang, Weiming Ren, Yuxuan Sun, et~al.
\newblock Mmmu: A massive multi-discipline multimodal understanding and reasoning benchmark for expert agi.
\newblock In \emph{Proceedings of the IEEE/CVF Conference on Computer Vision and Pattern Recognition}, pp.\  9556--9567, 2024{\natexlab{a}}.

\bibitem[Yue et~al.(2024{\natexlab{b}})Yue, Zheng, Ni, Wang, Zhang, Tong, Sun, Yin, Yu, Zhang, et~al.]{MMMU-Pro2024yue}
Xiang Yue, Tianyu Zheng, Yuansheng Ni, Yubo Wang, Kai Zhang, Shengbang Tong, Yuxuan Sun, Ming Yin, Botao Yu, Ge~Zhang, et~al.
\newblock Mmmu-pro: A more robust multi-discipline multimodal understanding benchmark.
\newblock \emph{ArXiv preprint}, abs/2409.02813, 2024{\natexlab{b}}.
\newblock URL \url{https://arxiv.org/abs/2409.02813}.

\bibitem[Zhang et~al.(2024)Zhang, Da, Lee, Robinson, Wu, Song, Zhao, Raja, Slack, Lyu, et~al.]{GSM1K2024zhang}
Hugh Zhang, Jeff Da, Dean Lee, Vaughn Robinson, Catherine Wu, Will Song, Tiffany Zhao, Pranav Raja, Dylan Slack, Qin Lyu, et~al.
\newblock A careful examination of large language model performance on grade school arithmetic.
\newblock \emph{ArXiv preprint}, abs/2405.00332, 2024.
\newblock URL \url{https://arxiv.org/abs/2405.00332}.

\bibitem[Zhang et~al.(2023)Zhang, Yang, Yuan, and Yao]{CR2023Zhang}
Yifan Zhang, Jingqin Yang, Yang Yuan, and Andrew Chi-Chih Yao.
\newblock Cumulative reasoning with large language models.
\newblock \emph{ArXiv preprint}, abs/2308.04371, 2023.
\newblock URL \url{https://arxiv.org/abs/2308.04371}.

\bibitem[Zheng et~al.(2023)Zheng, Liu, Xie, Li, and Li]{PHP2023Zheng}
Chuanyang Zheng, Zhengying Liu, Enze Xie, Zhenguo Li, and Yu~Li.
\newblock Progressive-hint prompting improves reasoning in large language models.
\newblock \emph{ArXiv preprint}, abs/2304.09797, 2023.
\newblock URL \url{https://arxiv.org/abs/2304.09797}.

\bibitem[Zhou et~al.(2023)Zhou, Wang, Lu, Shi, Luo, Qin, Lu, Jia, Song, Zhan, et~al.]{saturated2023zhou}
Aojun Zhou, Ke~Wang, Zimu Lu, Weikang Shi, Sichun Luo, Zipeng Qin, Shaoqing Lu, Anya Jia, Linqi Song, Mingjie Zhan, et~al.
\newblock Solving challenging math word problems using gpt-4 code interpreter with code-based self-verification.
\newblock \emph{ArXiv preprint}, abs/2308.07921, 2023.
\newblock URL \url{https://arxiv.org/abs/2308.07921}.

\end{thebibliography}
\bibliographystyle{iclr2025_conference}

\clearpage
\appendix
\section*{\revise{Limitations}}

\revise{This work has several limitations. 
First, {\benchmark} focuses on text-only reasoning, whereas some undergraduate-level math problems require images for their solutions. Developing a multimodal benchmark for undergraduate-level mathematics will be future work. 
Second, {\benchmark} is designed as an English-language benchmark. 
Extending {\benchmark} to support multiple languages could be an interesting avenue for future research.
Third, the number of problems in certain subjects is limited. 
Expanding these subjects would be valuable.}

\section{Detailed Statistics of {\benchmark}}\label{app: statistics}

\subsection{\revise{Intrinsic Mechanism of Dynamic Problems}}\label{app: dynamic_mechanism}

\revise{A key feature of our {\benchmark} is its dynamic nature. 
In this appendix, we detail how this is achieved. All problems in our homework grading system (see Appendix~\ref{app: data_source}) are stored as programs written in Program Generation, an established programming language for mathematics. 
These programs strictly specify conditions to ensure that the generated variations of each problem do not fundamentally alter their nature, required solution approach, difficulty level, or underlying knowledge points.
One such program is shown in Listing~\ref{lst:pg_int_tech}, and two different versions of this problem it generates are given in Figure~\ref{fig:wwexample}.
The relationship between the first and second variables is defined by \textit{\$$expnt = -1 + 2 * a;$}, which maintains the consistency of the concepts, techniques, and solutions involved in different versions of each problem.}

\subsection{Distribution of Problems}\label{app: problem_dist}

Our {\benchmark} covers various subjects in undergraduate-level mathematics. 
The detailed topics and the number of subtopics of each topic across different subjects are listed in Table~\ref{tabapp:topics} and \ref{tabapp:topics1}.
There are 111 topics and 583 subtopics across 16 subjects in total.
Furthermore, the distribution information of our benchmark on different subjects and difficulty level is presented in Table~\ref{tabapp:detailed distribution}.
Note that there are problems with missing difficulty level in our online homework grading system (see Appendix~\ref{app: data_source}) and we remain as is for consistency. 
The keywords of our {\benchmark} are shown in Figure~\ref{fig:word_cloud}.
\revise{The detailed results across different subjects and topics are discussed in Appendix~\ref{app:subjects} and \ref{app:topics}.}

\begin{table}[hbp]
\centering
\footnotesize
\caption{Topics of each subject and corresponding number of subtopics included in {\benchmark}.}
\scalebox{0.8}{
\begin{tabular}[tbp]{lp{7cm}c}
\toprule
\textbf{Subject}  & \textbf{Topics} & \textbf{\# Sub-Topics} \\ 
\midrule
Arithmetic & Integers &11 \\ &Fractions/rational numbers &12 \\ &Decimals &9 \\ &Percents &3 \\ &Irrational numbers &1 \\ &Other bases &3 \\ &Units &2  \\
\hline
Algebra & Algebra of real numbers and simplifying expressions&8 \\ &Absolute value expressions and functions &3 \\ &Properties of exponents, rational exponents and radicals&2 \\ &Cartesian coordinate system&4 \\ &Factoring &5 \\ & Functions&8 \\ &Transformations of functions and graphs & 6 \\ & Linear equations and functions &9 \\ &Quadratic equations and functions &8\\ & Operations on polynomial and rational expressions&7 \\ &Polynomial equations and functions &7 \\ &Variation and power functions &5 \\ &Systems of equations and inequalities &1 \\ & Functions with fractional exponents and radical functions& 1\\ &Rational equations and functions &6 \\ &Inverse functions &3 \\ &Exponential and logarithmic expressions and functions &8 \\ &Finite sequences and series &4 \\ &Conic sections &3 \\
\hline
Set theory and logic &Operations on sets &5 \\ &Relations between sets &2 \\ & Functions &2 \\ & Propositional logic& 4 \\ & First order logic &3 \\ &Pattern matching &1  \\
\hline
Trigonometry & Geometric and algebraic foundations for trigonometry &3 \\ &Trigonometric functions &8 \\ &Triangle trigonometry & 4 \\ &Analytic trigonometry  & 7 \\ & Polar coordinates \& vectors &2 \\
\hline
Combinatorics &Counting&8 \\ &Recurrence relations&3 \\
\hline
Geometry & Shapes & 4 \\ & Circle geometry &1 \\ &Vector geometry &7\\
\hline
Calculus single-variable & Calculus of vector valued functions& 6 \\ &Concepts for multivariable functions &6 \\ &Differentiation of multivariable functions &7 \\ &Integration of multivariable functions &9 \\ &Vector fields &1 \\ &Vector calculus &6 \\ &Fundamental theorems &4 \\
\hline
Calculus multivariable & Limits and continuity& 14 \\ &Differentiation &13 \\ &Applications of differentiation &18 \\ &Integrals &4 \\ &Techniques of integration &8 \\ &Applications of integration &16 \\ &Infinite sequences and series &18 \\ &Parametric &6 \\ &Polar &3 \\
\hline
Linear Algebra &Systems of linear equations&7 \\ &Matrices & 8 \\ & Matrix factorizations &5 \\ &Euclidean spaces & 8 \\ & Abstract vector spaces &7 \\ &Eigenvalues and eigenvectors &5 \\ &Inner products &6 \\ &Linear transformations &6 \\ &Determinants &3 \\
\hline
Number Theory &Divisibility &4 \\ & Congruences &5 \\ &Diophantine equations &1 \\
\hline
Financial Mathematics &Annuities &5 \\ & Bonds &3 \\ &Equations of value &2 \\ &Interest & 6 \\ & Options&5 \\ &Expected and contingent payments &2 \\ &Equities &2 \\
\bottomrule
\end{tabular}}
\label{tabapp:topics}
\end{table}

\begin{table}[tbp]
\centering
\footnotesize
\caption{Topics of each subject and corresponding number of subtopics included in {\benchmark} (Con't).}
\scalebox{0.8}{
\begin{tabular}{lp{7cm}c}
\toprule
\textbf{Subject}  & \textbf{Topics} & \textbf{\# Sub-Topics} \\ 
\midrule
Probability & Sample Space & 6 \\ &Random variables &5 \\ &Discrete distributions & 6\\ &Continuous distributions &5 \\ &Laws, theory &2 \\ &Several variables &2 \\
\hline
Statistics &Experimental design &1 \\ &Sample survey methods &2 \\ &Exploratory data analysis/descriptive statistics &5 \\ &Sampling distributions &3 \\ &Confidence intervals &9 \\ &Hypothesis tests &13 \\ &Simple linear regression &7 \\ &Time series & 2 \\ & Point estimation &1 \\ &Multiple regression &5 \\ & Generalized linear methods &1\\
\hline
Complex Analysis &Arithmetic& 8\\ &Complex equations &1 \\ &Complex plane &1 \\ &Complex functions &2 \\ &Analytic functions &3 \\ &Series and residues &4\\
\hline
Differential Equations & Introductory concepts &2 \\ &First order differential equations  &13 \\ &Higher order differential equations &9 \\ &Laplace transforms &6 \\ &Systems of differential equations &4 \\ &Numerical methods &1 \\ &Series solutions &3 \\ & Partial differential equations &7\\
\hline
Abstract Algebra & Groups&7 \\ & Rings&4 \\ &Fields and polynomials &1 \\
\hline
Total count &111 & 583 \\
\bottomrule
\end{tabular}}
\label{tabapp:topics1}
\end{table}

\begin{table}[!t]
\centering
\footnotesize
\caption{Statistics of {\benchmark} across different subjects and difficulty levels.}
\scalebox{0.9}{
\begin{tabular}{lccccccc}
\toprule
& \textbf{Level 1} & \textbf{Level 2} & \textbf{Level 3} & \textbf{Level 4} & \textbf{Level 5} & \textbf{Level 6} & \textbf{All} \\
\midrule
Arithmetic &22 &179 &47 &21 &53 &0 & 322 \\
Algebra & 25& 233& 176 &80 &73 &0 &583\\
Set theory and logic &0 &35& 11& 12& 9& 0& 69 \\
Trigonometry &5 &86&44&17&25&0&178 \\
Combinatorics & 3& 28& 47& 4& 5& 1 & 88\\
Geometry & 5& 48& 86 & 13& 6 & 0 &161\\
Calculus single-variable &4 & 297& 317& 175& 176& 1& 982 \\
Calculus multivariable &13 &279 &244 &80 &37 &0 &654 \\
Linear Algebra & 2& 172& 232& 53& 26& 0& 498 \\
Number Theory & 2& 33& 7& 1& 1& 2& 46 \\
Financial Mathematics & 7&52&105&72&89&0&346 \\
Probability & 11& 167& 139& 16& 2& 0& 336 \\
Statistics &44 &151& 158& 46& 0 &0 &401 \\
Complex Analysis &0 &40 &7 &2 &1 &1 &51 \\
Differential Equations &2 &47 &115 &56 &84 &0 &305 \\
Abstract Algebra & 0& 8 &16 &9 &1 &7 &42\\
\midrule
Grand Total & 145& 1,855& 1,751& 657& 588& 12 &5,062 \\
\bottomrule
\end{tabular}
}
\label{tabapp:detailed distribution}
\end{table}

\subsection{Answer Types}\label{app: answer_type}
By carefully reviewing a large collection of problems and referring to various past benchmarks \citep{OlympiadBench2024He, OlympicArena2024huang}, we classify all answers to be two main categories: atomic and compound.
There are 8 atomic types and 2 compound types.
Each compound type is composed of a list of atomic ones.
These types are designed to encompass a wide range of problems. 
Detailed definitions for each answer type can be found in Table~\ref{tabapp:AnswerTypes}.

\begin{table}[!t]

\centering
\footnotesize
\caption{Answer Types and Definitions}

\begin{tabular}{lp{9cm}}
\toprule

\textbf{Answer Type} & \textbf{Definition}  \\
\midrule
\multicolumn{2}{l}{{\textit{Atomic Types}}} \\
\midrule
Numerical Value (NV) & Problems where the answer is a numerical value, including special values such as $\pi$, $e$, $\sqrt{7}$, $\sin{\pi/8}$, etc., represented in LaTeX. \\

\midrule
Expression (EX) & Problems requiring an expression containing variables, e.g., $8x^2 + x + 1$, represented in LaTeX. \\

\midrule
Equation (EQ) & Problems requiring an equation containing variables, e.g., $y = 2x +1$ represented in LaTeX. \\

\midrule
Interval (INT) & Problems where the answer is a range of values, e.g., $(-\infty, 2) \cup (3, \infty)$ represented as an interval in LaTeX. \\

\midrule
True/False (TF) & Problems where the answer is either True or False, Yes or No, T or F, Y or N, etc. \\

\midrule
MC with Single answer (MCS) & Multiple-Choice (MC) problems with only one correct option (e.g., one out of four, one out of five, etc.). The options can be captical letters (ABCD) or any other string according to the problems (independent or dependent, etc.).\\

\midrule
MC with multiple answers (MCM) & Multiple-Choice (MC) problems with multiple correct options (e.g., two out of four, two out of five, two out of six, etc.). The options can be captical letters (ABCD) or any other string according to the problems (independent or dependent, etc.).\\

\midrule
Open-Ended (OE) & Problems whose answers can be a term, name, or any other string that satisfies the description of the problem, for example, the name of the variable or function that occurs in the problem (which should be treated differently with EX). \\

\midrule
\multicolumn{2}{l}{{\textit{Compound Types}}} \\

\midrule
Ordered List (OL) & Problems where the answer is an ordered list, e.g. a coordinate ( $(1, 2, 3)$, $(2t, t^2)$, etc.). \\
\midrule
Unordered List (UOL) & Problems where the answer is an unordered list, e.g., a set or multiple solutions for an equation. \\
\bottomrule
\end{tabular}%
\label{tabapp:AnswerTypes}
\end{table}

\begin{figure}
    \centering
    \includegraphics[width=1\linewidth]{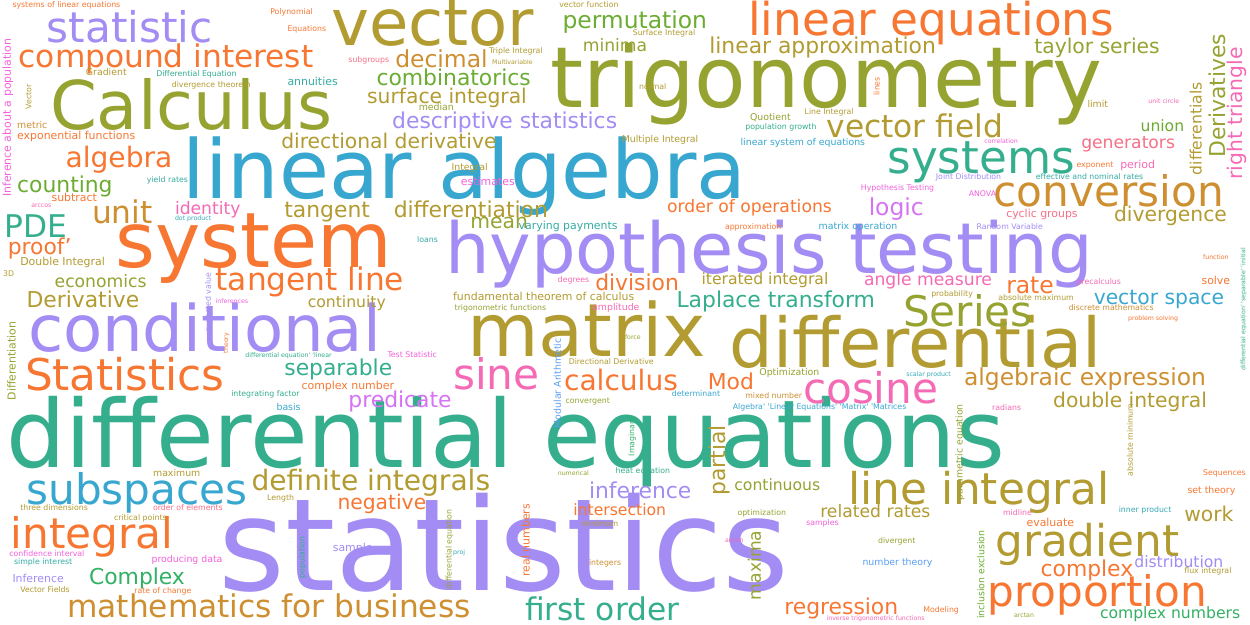}
    \caption{Word Cloud of Keywords in {\benchmark}.}
    \label{fig:word_cloud}
\end{figure}
\section{{\benchmark} Creation}\label{app: creation}

\subsection{Data Source}\label{app: data_source}
 {\benchmark} originates from questions in the online homework grading system of our institute, utilizing WebWork\footnote{\href{https://www.webwork.org}{https://www.webwork.org}}, an open-source online platform licensed under \href{https://www.gnu.org/licenses/gpl-3.0.en.html}{GNU}.
 Widely employed for assigning mathematics and science homework in educational settings, WebWork benefits from collaborative contributions by educators across various institutions.

Each question in WebWork is tagged with keywords related to concepts and difficulty level based on Bloom's taxonomy\footnote{\href{https://webwork.maa.org/wiki/Problem_Levels}{https://webwork.maa.org/wiki/Problem\_Levels}}, which helps simplify statistical analysis and cognitive assessment. 
To prevent cheating from each other, WebWork is able to generate tailored problem sets with different random seeds, making it popular among educational institutions.

\begin{figure}
    \centering
    \includegraphics[width=1\linewidth]{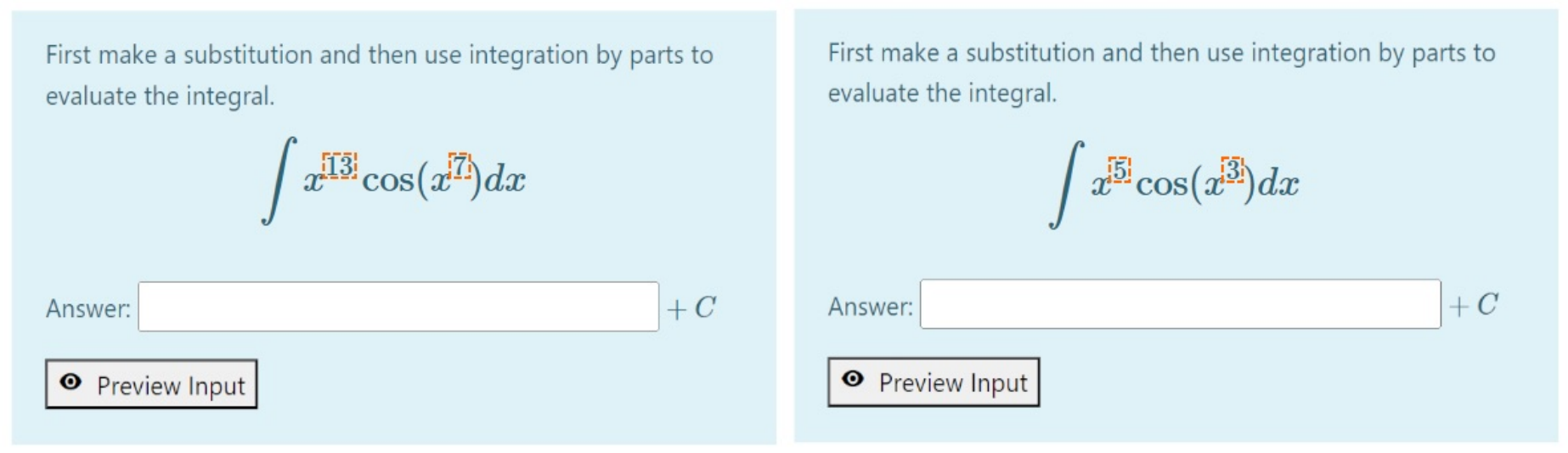}
    \caption{An Example of Our Online Homework Grading System Based on WebWork. The dashed red box highlights the differences between randomized versions of the problem.}
    \label{fig:wwexample}
\end{figure}

\begin{figure}
    \centering
    \includegraphics[width=1\linewidth]{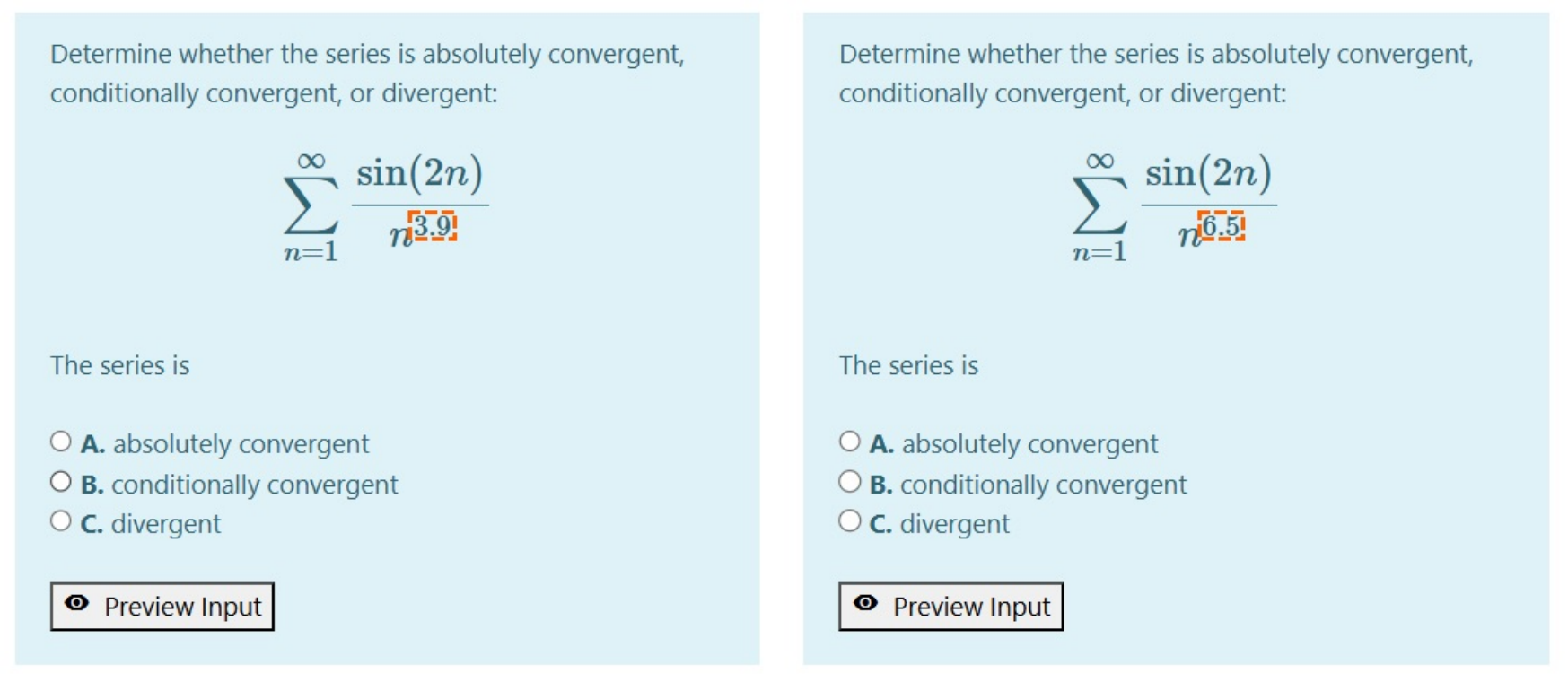}
    \caption{Another Example of Our Online Homework Grading System Based on WebWork.}
    \label{fig:wwexample_series}
\end{figure}

\revise{As mentioned in Appendix~\ref{app: dynamic_mechanism}, our problem generation programs will remain the main problem structure and knowledge points.
One such example is illustrated in Figure~\ref{fig:wwexample} and the corresponding program is given in Listing~\ref{lst:pg_int_tech}.
The dashed red box highlights the differences between randomized versions of the problem.
For this specific example, the second random variable is initialized through \textit{$a=random(2, 10)$}. 
The relationship between the first and second variables is defined by \textit{$expnt = -1 + 2 * a$}.
This meticulous setup maintains the consistency of the concepts, techniques, and solutions involved in different versions of each problem.
Another example is demonstrated in Figure~\ref{fig:wwexample_series}, where 'n' is initialized to a randomly chosen exponent from 2 to 10 in increments of 0.1.}


All problems within WebWork are stored in the Problem Generation language and are presented in HTML format with JavaScript and external resources. This poses challenges for human interpretation and LLMs analysis, urging us to clean and re-format them in Latex (see Section~\ref{sec: creation}).



\begin{lstlisting}[caption={Problem Generator Code for the Problem in Figure ~\ref{fig:wwexample}}, label={lst:pg_int_tech}]
$a=random(2, 10); # all-inclusive integers between 2 and 10
$expnt = -1+2*$a;
$ans = "1/$a*x^$a*sin(x^$a)+1/$a*cos(x^$a)"; #Right answer

TEXT(beginproblem());
BEGIN_TEXT
(*@\textit{First make a substitution and then use integration by parts to evaluate the integral.}@*)
$BR
\[ \int x^{$expnt} \cos(x^$a) dx \]
$BR
Answer: \{ans_rule(40)\}  \(+\) \(C\)
END_TEXT

#Compare the right answer to the input
ANS(fun_cmp($ans, mode=>'antider'));
\end{lstlisting}

\subsection{Deduplication}\label{app: deduplication}
After converting and cleaning all the problems, we perform deduplication within each subject.
More specifically, we adhere to the following steps:
First, we transform each question into a vector with dimension 1536 using the embedding model \texttt{text-embedding-ada-002}, which is the most capable 2nd generation embedding model of OpenAI\footnote{The information can be found at \href{https://openai.com/index/new-embedding-models-and-api-updates/}{https://openai.com/index/new-embedding-models-and-api-updates/}}.
We then calculate pairwise cosine similarities using the embeddings in the previous step. 
Finally, a threshold is selected based on manual inspection within each subject, and problems that have a cosine similarity higher than that threshold with existing problems are excluded.
\revise{The thresholds and the number of questions filtered out for different subjects are presented in Table~\ref{tabapp:deduplication}. 
We use subject-agnostic thresholds and filter out 9,382 questions in total.}

\begin{table}[!t]
\centering
\footnotesize
\caption{Thresholds of deduplication and the number of questions that filtered out in {\benchmark}.}
\scalebox{1.0}{
\begin{tabular}{lcccc}
\toprule
& \textbf{Threshold} & \textbf{Before}  & \textbf{After} & \textbf{\# Filter Out} \\
\midrule
Arithmetic& 0.92 & 845 & 322 & 523 \\
Algebra & 0.86 & 4663 & 583 & 4080 \\
Set theory and logic & 0.96 & 94 & 69 & 25 \\
Trigonometry &  0.91 & 399 & 178 & 221 \\
Combinatorics & 0.89 & 140 & 88 & 52\\
Geometry & 0.92 & 267 & 161 & 106\\
Calculus single-variable & 0.90 & 3234 & 982 & 2252\\
Calculus multivariable & 0.94 & 866 & 654 & 212 \\
Linear Algebra & 0.93 & 1235 & 498 & 737 \\
Number Theory & 0.92 & 68 & 46 & 22 \\
Financial Mathematics & 0.92 & 653 &346 & 307 \\
Probability & 0.91 & 453 & 336 & 117 \\
Statistics & 0.92 & 599 &401 & 198 \\
Complex Analysis & 0.95 & 129 & 51 & 78 \\
Differential Equations & 0.92 & 741 &305 
 & 436\\
Abstract Algebra & 0.94 & 58 & 42 & 16 \\
\midrule
Grand Total & - & 14444 &5,062 & 9382 \\
\bottomrule
\end{tabular}
}
\label{tabapp:deduplication}
\end{table}

\section{Detailed Experimental Setup}\label{app: exp_setup}

\subsection{Evaluated LLMs}\label{app: models}


A variety of LLMs are covered in our evaluation, including closed-source commercial models and open-source models, general-purpose models and models dedicated for math problem solving. Closed-source LLMs are as follows:
\begin{itemize}
    \item \textbf{o1-preview} \citep{o1}: An early preview of OpenAI's o1 model, designed to reason about hard problems using broad general knowledge about the world. We used o1-preview-2024-09-12 for our evaluation.
    \item \textbf{GPT-4o} \citep{gpt4o2024openai}: GPT-4o is multimodal, and has the same high intelligence as GPT-4 Turbo but is much more efficient.
    \item \textbf{Claude-3-Opus} \citep{claude3}: Anthropic's most intelligent model, claimed to outperform its peers on most of the common evaluation benchmarks for AI systems.
\end{itemize}

We evaluated the following open-source general-purpose LLMs on our benchmark:
\begin{itemize}
    \item \textbf{Llama-3-Instruct} \citep{llama3modelcard}:  \href{https://www.llama.com/llama3/license/}{LLaMA 3 Community License}.
    \item \textbf{Mistral-7B-Instruct-v0.3} \citep{mistral_7b_v0_3_repo}: \href{https://www.apache.org/licenses/LICENSE-2.0}{Apache 2.0}
    \item \textbf{Mistral-Nemo-Instruct-2407} \citep{mistral_nemo_blog}: \href{https://www.apache.org/licenses/LICENSE-2.0}{Apache 2.0}
    \item \textbf{Mistral-Small-Instruct-2409} \citep{mistral_small_blog}: \href{https://mistral.ai/licenses/MRL-0.1.md}{MRL License}.
    \item \textbf{Mistral-Large-Instruct-2407} \citep{mistral_large_blog}: \href{https://mistral.ai/licenses/MRL-0.1.md}{MRL License}.
    \item \textbf{Qwen2-Instruct} \citep{Qwen22024yang}: Qwen2 series are developed with dedication to math and coding. We used 7B and 72B models. 7B models are licensed under \href{https://www.apache.org/licenses/LICENSE-2.0}{Apache 2.0}, while 72B models are under \href{https://huggingface.co/Qwen/Qwen2-72B/blob/main/LICENSE}{Tongyi Qianwen License}.
    \item \textbf{Yi-1.5-Chat} \citep{ai2024yi}: Yi-1.5 delivers stronger performance in coding, math, reasoning, and instruction-following capability compared to its predecessor. We used 6B, 9B, 34B variants. Yi-1.5 series are licensed under \href{https://www.apache.org/licenses/LICENSE-2.0}{Apache 2.0}.
    \item \textbf{DeepSeek-V2-Lite-Chat} \citep{deepseekv2}: model under \href{https://github.com/deepseek-ai/DeepSeek-V2/blob/main/LICENSE-MODEL}{Model License} code under \href{https://github.com/deepseek-ai/DeepSeek-V2/blob/main/LICENSE-CODE}{MIT License}.
    \item \textbf{deepseek-moe-16b-chat} \citep{dai2024deepseekmoe}: model under \href{https://github.com/deepseek-ai/DeepSeek-MoE/blob/main/LICENSE-MODEL}{Model License}, code under \href{https://github.com/deepseek-ai/DeepSeek-MoE/blob/main/LICENSE-CODE}{MIT License}.
\end{itemize}

The following specialized math LLMs are evaluated in our study:
\begin{itemize}
    \item \textbf{DeepSeekMath-7B} \citep{deepseekmath2024shao}: DeepSeekMath is initialized with DeepSeek-Coder-v1.5 7B and continues pre-training on math-related tokens. We tested both DeepSeekMath-7B-RL and DeepSeekMath-7B-Instruct variants. Models are under \href{https://github.com/deepseek-ai/DeepSeek-Math/blob/main/LICENSE-MODEL}{Model License} while code is under \href{https://github.com/deepseek-ai/DeepSeek-Math/blob/main/LICENSE-CODE}{MIT License}.
    \item \textbf{Qwen2-Math} \citep{Qwen25Math2024Yang}: Qwen2-Math is a series of specialized math language models built upon the Qwen2 LLMs. We evaluated 7B and 72B variants. They are under the same license as Qwen2-Instruct series.
    \item \textbf{Mathstral-7B} \citep{mathstral2023}: Mathstral stands on the shoulders of Mistral 7B and specializes in STEM subjects. This model is published under \href{https://www.apache.org/licenses/LICENSE-2.0}{Apache 2.0}.
    \item \textbf{Numinamath-7B-CoT} \citep{numinamath7b}: This model is finetuned from DeepSeekMath-7B-base with two stages of supervised fine-tuning to solve math problems using chain of thought (CoT). It is licensed under \href{https://www.apache.org/licenses/LICENSE-2.0}{Apache 2.0}.
\end{itemize}

\subsection{Evaluation Prompts}\label{app: prompts}

The evaluation prompts in our experiments are given in Table~\ref{tabapp:prompts}, where detailed answer type descriptions are given in Table~\ref{tabapp:answer type instructions}.
Following \citet{OlympiadBench2024He, OlympicArena2024huang}, these prompts are specially designed for different subjects and answer types for better evaluation. 
Note that, for chat models, we will apply \href{https://huggingface.co/docs/transformers/main/en/chat_templating}{chat template} for better evaluation.

\begin{table}[!t]
\centering
\footnotesize
\caption{Evaluation prompts for problems with single answer or multiple answers. \{problem\} is the specific problem to evaluate. \{subject\} denotes the subject this problem belongs to and all subjects are given in Table~\ref{tabapp:topics}. \{answer\_type\_description\} are specified in Table~\ref{tabapp:answer type instructions}. \{number\_of\_answers\} stands for the number of answers in the problem evaluated.}

\begin{tabular}{p{12cm}}
\toprule
Evaluation Prompt for Single Answer \\
\midrule
The following is an undergraduate-level mathematical problem in \{subject\}. You need to solve the problem by completing all placeholders [ANS]. \\ \\
This problem involves only one placeholders [ANS] to be completed. The answer type is \{answer\_type\_description\}. \\ \\
Problem: \\ \{problem\} \\ \\
All mathematical formulas and symbols you output should be represented with LaTeX. Please end your response with: "The final answer is $\boxed{\text{ANSWER}}$, where ANSWER should be your final answer. \\

\midrule
Evaluation Prompt for Multiple Answers \\
\midrule
The following is an undergraduate-level mathematical problem in \{subject\}. You need to solve the problem by completing all placeholders [ANS]. \\ \\
This problem involves \{number\_of\_answers\} placeholders [ANS] to be completed. Their answer types are, in order, \{answer\_type\_description\}. \\ \\
Problem: \\ \{problem\} \\ \\
All mathematical formulas and symbols you output should be represented with LaTeX. Please end your response with: "The final answer is $\boxed{\text{ANSWER}}$, where ANSWER should be the sequence of your final answers, separated by commas. \\

\bottomrule
\end{tabular}%

\label{tabapp:prompts}
\end{table}

\begin{table}[!t]
\centering
\footnotesize
\caption{Descriptions of answer types included in evaluation prompts, where \{options\} is the specific options from the multiple choice question evaluated.}

\scalebox{1.0}{
\begin{tabular}{lp{9cm}}
\toprule

\textbf{Answer Type} & \textbf{Answer Type Description} \\

\midrule
NV & a numerical value without units \\

\midrule
EX & an expression \\

\midrule
EQ & an equation \\

\midrule
INT & a range interval \\

\midrule
TF & either True or False\\

\midrule
MCS & one option for a multiple choice question with options \{options\} \\

\midrule
MCM & more than one option concatenated without space or commas of a multiple choice question with options \{options\}, for example: BD \\

\midrule
OE & a word, phrase, term or string that satisfies the requirements of the problem \\

\midrule
OL & an ordered list of answers surrounded by parentheses with any answer types, for example $(1, x^2, True)$, where "ordered list" means changing the order of elements results in different answers \\

\midrule
UOL & an unordered list of answers surrounded by parentheses with any answer types, for example, $(1, x^2, True)$, where "unordered list" means changing the order of elements results in the same answer \\

\bottomrule
\end{tabular}%
}
\label{tabapp:answer type instructions}
\end{table}

\section{\revise{Results Across Different Subjects}}\label{app:subjects}

\revise{The detailed results across different subjects are given in Table~\ref{tabapp: abstract_algebra}, \ref{tabapp: algebra}, \ref{tabapp: arithmetic}, \ref{tabapp: calculus_multivariable}, \ref{tabapp: calculus_singlevariable}, \ref{tabapp: combinatorics}, \ref{tabapp: complex_analysis}, \ref{tabapp: differential_equations}, \ref{tabapp: financial_mathematics}, \ref{tabapp: geometry}, \ref{tabapp: linear_algebra}, \ref{tabapp: number_theory}, \ref{tabapp: probability}, \ref{tabapp: set_theory}, \ref{tabapp: statistics}, and \ref{tabapp: trigonometry}.
From the results, we have the following observations:
\begin{itemize}
    \item OpenAI-o1-mini achieves the best results across nearly all subjects, although GPT-4o sometimes excels in terms of RE.
    \item For open-source LLMs, Qwen-2-Math-72B-instruct achieves the best results in almost all subjects. However, Mistral-Large-instruct-2407 outperforms Qwen-2-Math-72B-instruct in Algebra.
    \item Some LLMs even achieve zero EAcc in certain subjects. For instance, Mistral-7B-Instruct get zero EAcc in Set Theory and Logic, and seven LLMs exhibit zero EAcc in Abstract Algebra.
    \item The variation in the reasoning gap differs significantly across subjects, providing more fine-grained information on how different LLMs perform across various domains.
\end{itemize}}

\begin{table*}[!thb]
\centering
\footnotesize
\caption{\textbf{Results on Abstract Algebra} (all figures are in \%).
The best results within each column are \textbf{bolded} and the best results of open-source Chat LLMs within a similar parameter size group are \underline{underlined}.}
\scalebox{1.0}{
\begin{tabular}{lccccccc}
\hline
\textbf{Models} & \textbf{$\text{Acc}_1$} & \textbf{$\text{Acc}_2$} & \textbf{$\text{Acc}_3$} & \textbf{AAcc} & \textbf{EAcc} & \textbf{$\Delta$} & \textbf{RE}\\
\hline
\multicolumn{7}{c}{{\textit{Closed-source LLMs}}} \\
\cdashline{1-8}
OpenAI-o1-mini-2024-09-12 & \textbf{76.19} & \textbf{73.81} & \textbf{71.42} & \textbf{73.81} & \textbf{57.14} & 16.67 & \textbf{29.17} \\
GPT-4o-2024-08-06 & 42.86 & 50.00 & 52.38 & 48.41 & 28.57 & 19.84 & 69.44 \\
GPT-4o-mini-2024-07-18 & 26.19 & 35.71 & 38.09 & 33.33 & 14.29 & 19.04 & 133.2 \\
Claude-3-Opus-20240229 & 23.81 & 38.10 & 26.19 & 29.37 & 11.90& 17.47 & 146.8 \\
\hline
\multicolumn{7}{c}{{\textit{Open-source Chat LLMs}}} \\
\cdashline{1-8}
Yi-1.5-6B-Chat & 4.76 & 2.38 & 9.52 & 5.56 & 0.00 & 5.56 & $\infty$ \\
Mistral-7B-Instruct & 2.38 & 2.38 & 0.00 & 1.59 & 0.00 & \underline{\textbf{1.59}} & $\infty$\\
Qwen2-7B-Instruct & 4.76 & \underline{16.67} & \underline{19.05} & \underline{13.49} & \underline{4.76} & 8.73 & \underline{183.4} \\
LLaMA3-8B-Instruct & 0.00 & 7.14 & 9.52 & 5.56 & 0.00 & 5.56 & $\infty$ \\
Yi-1.5-9B-Chat & \underline{7.14} & 14.29 & 11.90 & 11.11 & 0.00 & 11.11 & $\infty$ \\
\hline
Mistral-Nemo-Instruct-2407 & 0.00 & 9.52 & 4.76 & 4.76 & 0.00 & 4.76 & $\infty$ \\ 
DeepSeek-MOE-16B-Chat & 0.00 & 2.38 & 2.38 & 1.59 & 0.00 & \underline{\textbf{1.59}} & $\infty$ \\
DeepSeek-V2-Lite-Chat & 0.00 & 2.38 & 2.38 & 1.59 & 0.00 & \underline{\textbf{1.59}} & $\infty$ \\
Mistral-Small-Instruct-2409 & 9.52 & \underline{19.05} & \underline{21.43} & \underline{16.67} & \underline{7.14} & 9.53 & \underline{133.5} \\ 
Yi-1.5-34B-Chat & \underline{19.05} & 9.52 & 9.52 & 12.70 & 0.00 & 12.70 & $\infty$ \\
\hline
LLaMA3-70B-Instruct & 19.05 & 16.67 & 19.04 & 18.25 & 4.76 & \underline{13.49} & 283.4\\
Qwen2-72B-Instruct & 28.57 & \underline{47.62} & 30.95 & 35.71 & 14.29 & 21.42 & 149.9 \\
Mistral-Large-Instruct-2407 & \underline{35.71} & 35.71 & \underline{45.24} & \underline{38.89} & \underline{21.43} & 17.46 & \underline{81.47} \\
\hline
\multicolumn{7}{c}{{\textit{Specialized Mathematical LLMs}}} \\
\cdashline{1-8}
DeepSeek-Math-7B-Instruct & 7.14 & 9.52 & 4.76 & 7.14 & 0.00 & 7.14 & $\infty$ \\
DeepSeek-Math-7B-RL & 4.76 & 19.05 & 11.90 & 11.90 & 2.38 & 9.52 & 400.0 \\
NuminaMath-7B-CoT & 7.14 & 16.67 & 7.14 & 10.32 & 0.00 & 10.32 & $\infty$ \\
Mathstral-7B-v0.1 & 9.52 & 4.76 & 7.14 & 7.14 & 0.00 & 7.14 & $\infty$ \\
Qwen2-Math-7B-Instruct & 23.81 & 33.33 & 28.57 & 28.57 & 14.29 & 14.28 & 99.93 \\
Qwen2-Math-72B-Instruct & 45.24 & 40.48 & 42.86 & 42.86 & 26.19 & 16.67 & 63.65 \\
\hline
\end{tabular}
}

\label{tabapp: abstract_algebra}
\end{table*}

\begin{table*}[!thb]
\centering
\footnotesize
\caption{\textbf{Results on Algebra} (all figures are in \%).
The best results within each column are \textbf{bolded} and the best results of open-source Chat LLMs within a similar parameter size group are \underline{underlined}.
} 
\scalebox{1.0}{
\begin{tabular}{lccccccc}
\hline
\textbf{Models} & \textbf{$\text{Acc}_1$} & \textbf{$\text{Acc}_2$} & \textbf{$\text{Acc}_3$} & \textbf{AAcc} & \textbf{EAcc} & \textbf{$\Delta$} & \textbf{RE}\\
\hline
\multicolumn{7}{c}{{\textit{Closed-source LLMs}}} \\
\cdashline{1-8}
OpenAI-o1-mini-2024-09-12 & \textbf{76.16} & \textbf{75.47} & \textbf{74.44} & \textbf{75.36} & \textbf{66.72} & 8.64 & 12.95 \\
GPT-4o-2024-08-06 & 70.50 & 71.53 & 72.04 & 71.36 & 64.67 & \textbf{6.69} & \textbf{10.34} \\
GPT-4o-mini-2024-07-18 & 67.41 & 67.24 & 66.38 & 67.01 & 59.52 & 7.49 & 12.58 \\
Claude-3-Opus-20240229 & 63.29 & 66.38 & 63.81 & 64.49 & 54.20& 10.29 & 18.99 \\
\hline
\multicolumn{7}{c}{{\textit{Open-source Chat LLMs}}} \\
\cdashline{1-8}
Yi-1.5-6B-Chat & 38.77 & 37.22 & 40.48 & 38.82 & 25.21 & 13.61 & 53.99 \\
Mistral-7B-Instruct & 19.73 & 21.96 & 19.55 & 20.41 & 11.32 & \underline{9.09} & 80.30\\
Qwen2-7B-Instruct & \underline{51.63} & \underline{53.00} & \underline{54.03} & \underline{52.89} & \underline{41.85} & 11.04 & \underline{26.38}\\
LLaMA3-8B-Instruct & 30.53 & 33.28 & 30.19 & 31.33 & 18.35 & 12.98 & 70.74 \\
Yi-1.5-9B-Chat & 46.83 & 48.89 & 48.37 & 48.03 & 33.79 & 14.24 & 42.14 \\
\hline
Mistral-Nemo-Instruct-2407 & 39.79 & 43.05 & 39.79 & 40.88 & 28.82 & 12.06 & 41.85 \\ 
DeepSeek-MOE-16B-Chat & 12.01 & 12.35 & 10.98 & 11.78 & 4.97 & \underline{6.81} & 137.0 \\
DeepSeek-V2-Lite-Chat & 23.33 & 24.01 & 24.53 & 23.96 & 11.32 & 12.64 & 111.7 \\
Mistral-Small-Instruct-2409 & \underline{58.32} & \underline{57.46} & \underline{56.95} & \underline{57.58} & \underline{47.86} & 9.72 & \underline{20.31} \\ 
Yi-1.5-34B-Chat & 53.34 & 53.00 & 52.66 & 53.00 & 40.31 & 12.69 & 31.48 \\
\hline
LLaMA3-70B-Instruct & 50.09 & 51.80 & 51.11 & 51.00 & 40.48 & 10.52 & 25.99\\
Qwen2-72B-Instruct & 63.81 & 63.46 & 64.49 & 63.92 & 54.72 & 9.20 & 16.81 \\
Mistral-Large-Instruct-2407 & \underline{68.44} & \underline{68.10} & \underline{67.92} & \underline{68.15} & \underline{59.86} & \underline{8.29} & \underline{13.85} \\
\hline
\multicolumn{7}{c}{{\textit{Specialized Mathematical LLMs}}} \\
\cdashline{1-8}
DeepSeek-Math-7B-Instruct & 40.31 & 38.77 & 38.25 & 39.11 & 26.07 & 13.04 & 50.02 \\
DeepSeek-Math-7B-RL & 44.43 & 47.51 & 44.94 & 45.63 & 34.31 & 11.32 & 32.99 \\
NuminaMath-7B-CoT & 45.80 & 45.45 & 46.14 & 45.80 & 34.65 & 11.15 & 32.18 \\
Mathstral-7B-v0.1 & 45.97 & 43.91 & 43.57 & 44.48 & 32.25 & 12.23 & 37.92 \\
Qwen2-Math-7B-Instruct & 56.09 & 57.46 & 57.63 & 57.06 & 47.86 & 9.20 & 19.22 \\
Qwen2-Math-72B-Instruct & 67.58 & 67.75 & 67.24 & 67.52 & 58.83 & 8.69 & 14.77 \\
\hline
\end{tabular}
}

\label{tabapp: algebra}
\end{table*}

\begin{table*}[!thb]
\centering
\footnotesize
\caption{\textbf{Main Results on Arithmetic} (all figures are in \%).
The best results within each column are \textbf{bolded} and the best results of open-source Chat LLMs within a similar parameter size group are \underline{underlined}.} 
\scalebox{1.0}{
\begin{tabular}{lccccccc}
\hline
\textbf{Models} & \textbf{$\text{Acc}_1$} & \textbf{$\text{Acc}_2$} & \textbf{$\text{Acc}_3$} & \textbf{AAcc} & \textbf{EAcc} & \textbf{$\Delta$} & \textbf{RE}\\
\hline
\multicolumn{7}{c}{{\textit{Closed-source LLMs}}} \\
\cdashline{1-8}
OpenAI-o1-mini-2024-09-12 & \textbf{93.48} & \textbf{92.86} & \textbf{94.10} & \textbf{93.48} & \textbf{90.37} & \textbf{3.11} & \textbf{3.44} \\
GPT-4o-2024-08-06 & 91.93 & 91.30 & 92.24 & 91.82 & 87.27 & 4.55 & 5.21 \\
GPT-4o-mini-2024-07-18 & 89.44 & 87.27 & 90.06 & 88.92 & 83.23 & 5.69 & 6.84 \\
Claude-3-Opus-20240229 & 84.78 & 85.71 & 86.96 & 85.82 & 78.26& 7.56 & 9.66 \\
\hline
\multicolumn{7}{c}{{\textit{Open-source Chat LLMs}}} \\
\cdashline{1-8}
Yi-1.5-6B-Chat & 63.98 & 64.91 & 67.70 & 65.53 & 50.00 & 15.53 & 31.06\\
Mistral-7B-Instruct & 30.43 & 34.78 & 33.54 & 32.92 & 15.22 & 17.70 & 116.3\\
Qwen2-7B-Instruct & \underline{77.64} & \underline{78.26} & \underline{78.57} & \underline{78.16} & \underline{68.94} & \underline{9.22} & \underline{13.37}\\
LLaMA3-8B-Instruct & 51.24 & 53.42 & 54.97 & 53.21 & 33.54 & 19.67 & 58.65 \\
Yi-1.5-9B-Chat & 73.60 & 73.60 & 70.81 & 72.67 & 59.32 & 13.35 & 22.51 \\
\hline
Mistral-Nemo-Instruct-2407 & 69.88 & 70.19 & 72.67 & 70.91 & 55.28 & 15.63 & 28.27 \\ 
DeepSeek-MOE-16B-Chat & 26.09 & 30.43 & 31.68 & 29.40 & 12.42 & 16.98 & 136.7 \\
DeepSeek-V2-Lite-Chat & 50.31 & 53.42 & 49.07 & 50.93 & 32.92 & 18.01& 54.71 \\
Mistral-Small-Instruct-2409 & \underline{82.61} & \underline{84.78} & \underline{81.68} & \underline{83.02} & \underline{73.29} & \underline{9.73} & \underline{13.28} \\ 
Yi-1.5-34B-Chat & 75.47 & 74.84 & 75.47 & 75.26& 62.42 & 12.84 & 20.57 \\
\hline
LLaMA3-70B-Instruct & 79.81 & 82.61 & 83.85 & 82.09 & 72.36 & 9.73 & 13.45\\
Qwen2-72B-Instruct & 88.51 & 88.20 & 89.44& 88.72 & \underline{82.61} & \underline{6.11} & \underline{7.40} \\
Mistral-Large-Instruct-2407 & \underline{90.37} & \underline{89.13} & \underline{90.06} & \underline{89.86} & \underline{82.61} & 7.25 & 8.78 \\
\hline
\multicolumn{7}{c}{{\textit{Specialized Mathematical LLMs}}} \\
\cdashline{1-8}
DeepSeek-Math-7B-Instruct & 62.73 & 68.63 & 65.22 & 65.53 & 49.69 & 15.84 & 31.88 \\
DeepSeek-Math-7B-RL & 74.53 & 76.71 & 76.71 & 75.98 & 67.08 & 8.90 & 13.27 \\
NuminaMath-7B-CoT & 74.22 & 75.78 & 77.33 & 75.78 & 62.42 & 13.36& 21.40 \\
Mathstral-7B-v0.1 & 71.74 & 72.67 & 72.36 & 72.26 & 59.63 & 12.63 & 21.18 \\
Qwen2-Math-7B-Instruct & 88.51 & 84.78 & 86.65 & 86.65 & 78.88 & 7.77 & 9.85 \\
Qwen2-Math-72B-Instruct & 90.68 & 90.68 & 90.99 & 90.79 & 86.96 & 3.83 & 4.40 \\
\hline
\end{tabular}
}

\label{tabapp: arithmetic}
\end{table*}

\begin{table*}[!thb]
\centering
\footnotesize
\caption{\textbf{Main Results on Calculus - multivariable} (all figures are in \%).
The best results within each column are \textbf{bolded} and the best results of open-source Chat LLMs within a similar parameter size group are \underline{underlined}.}
\scalebox{1.0}{
\begin{tabular}{lccccccc}
\hline
\textbf{Models} & \textbf{$\text{Acc}_1$} & \textbf{$\text{Acc}_2$} & \textbf{$\text{Acc}_3$} & \textbf{AAcc} & \textbf{EAcc} & \textbf{$\Delta$} & \textbf{RE}\\
\hline
\multicolumn{7}{c}{{\textit{Closed-source LLMs}}} \\
\cdashline{1-8}
OpenAI-o1-mini-2024-09-12 & \textbf{59.33} & \textbf{60.09} & \textbf{59.94} & \textbf{59.79} & \textbf{48.32} & 11.47 & \textbf{23.74} \\
GPT-4o-2024-08-06 & 50.00 & 49.69 & 49.24 & 49.64 & 38.23 & 11.41 & 29.85 \\
GPT-4o-mini-2024-07-18 & 42.81 & 44.80 & 43.27 & 43.63 & 32.11 & 11.52 & 35.88 \\
Claude-3-Opus-20240229 & 35.93 & 39.14 & 35.93 & 37.00 & 24.01 & 12.99 & 54.10 \\
\hline
\multicolumn{7}{c}{{\textit{Open-source Chat LLMs}}} \\
\cdashline{1-8}
Yi-1.5-6B-Chat & 20.34 & 22.48 & 20.03 & 20.95 & 10.40 & 10.55 & 101.4 \\
Mistral-7B-Instruct & 5.05 & 3.98 & 4.59 & 4.54 & 1.22 & \underline{3.32} & 272.1 \\
Qwen2-7B-Instruct & 25.38 & 25.23 & 26.15 & 25.59 & 16.06 & 9.53 & \underline{59.34} \\
LLaMA3-8B-Instruct & 6.27 & 7.80 & 7.03 & 7.03 & 2.45 & 4.58 & 186.9 \\
Yi-1.5-9B-Chat & \underline{28.75}& \underline{28.75} & \underline{29.51} & \underline{29.00} & \underline{16.67} & 12.33 & 73.97 \\
\hline
Mistral-Nemo-Instruct-2407 & 13.61 & 12.39 & 10.55 & 12.18 & 5.81 & 6.37 & 109.6 \\
DeepSeek-MOE-16B-Chat & 1.68 & 0.46 & 1.22 & 1.12 & 0.00 & \textbf{\underline{1.12}} & $\infty$ \\
DeepSeek-V2-Lite-Chat & 5.05 & 4.89 & 3.52 & 4.49 & 1.38 & 3.11 & 225.4 \\
Mistral-Small-Instruct-2409 & \underline{27.98} & \underline{29.51} & \underline{27.68} & \underline{28.39} & \underline{17.28} & 11.11 & \underline{64.29} \\
Yi-1.5-34B-Chat & 26.45 & 27.98 & 26.15 & 26.86 & 14.22 & 12.64 & 88.89\\
\hline
LLaMA3-70B-Instruct & 20.79 & 23.39 & 20.95 & 21.71 & 12.23 & \underline{9.48} & 77.51 \\
Qwen2-72B-Instruct & 35.02 & 37.92 & 35.32 & 36.09 & 23.55 & 12.54 & 53.25\\
Mistral-Large-Instruct-2407 & \underline{46.94} & \underline{49.08} & \underline{47.71} & \underline{47.91} & \underline{36.70} & 11.21 & \underline{30.54} \\
\hline
\multicolumn{7}{c}{{\textit{Specialized Mathematical LLMs}}} \\
\cdashline{1-8}
DeepSeek-Math-7B-Instruct & 11.16 & 12.54 & 12.23 & 11.98 & 5.35 & 6.63 & 123.9 \\
DeepSeek-Math-7B-RL & 18.81 & 17.58 & 16.06 & 17.48 & 8.56 & 8.92 & 104.2 \\
NuminaMath-7B-CoT & 20.49 & 20.80 & 22.02 & 21.10 & 11.47 & 9.63 & 83.96 \\
Mathstral-7B-v0.1 & 16.36 & 17.89 & 17.89 & 17.38 & 9.63 & 7.75 & 80.48 \\
Qwen2-Math-7B-Instruct & 35.78 & 36.09 & 33.33 & 35.07 & 24.77 & 10.30 & 41.58\\
Qwen2-Math-72B-Instruct & 48.47 & 49.39 & 51.22 & 49.69 & 39.45 & 10.24 & 25.96\\
\hline
\end{tabular}
}

\label{tabapp: calculus_multivariable}
\end{table*}

\begin{table*}[!thb]
\centering
\footnotesize
\caption{\textbf{Main Results on Calculus - single variable} (all figures are in \%).
The best results within each column are \textbf{bolded} and the best results of open-source Chat LLMs within a similar parameter size group are \underline{underlined}.} 
\scalebox{1.0}{
\begin{tabular}{lccccccc}
\hline
\textbf{Models} & \textbf{$\text{Acc}_1$} & \textbf{$\text{Acc}_2$} & \textbf{$\text{Acc}_3$} & \textbf{AAcc} & \textbf{EAcc} & \textbf{$\Delta$} & \textbf{RE}\\
\hline
\multicolumn{7}{c}{{\textit{Closed-source LLMs}}} \\
\cdashline{1-8}
OpenAI-o1-mini-2024-09-12 & \textbf{72.81} & \textbf{73.32} & \textbf{72.61} & \textbf{72.91} & \textbf{59.37} & 13.54 & 22.81 \\
GPT-4o-2024-08-06 & 63.65 & 65.48 & 65.48 & 64.87 & 53.05 & 11.82 & \textbf{22.28} \\
GPT-4o-mini-2024-07-18 & 54.68 & 57.43 & 55.91 & 56.01 & 42.26 & 13.75 & 32.53 \\
Claude-3-Opus-20240229 & 47.96 & 51.32 & 50.61 & 49.97 & 34.73 & 15.24 & 43.88 \\
\hline
\multicolumn{7}{c}{{\textit{Open-source Chat LLMs}}} \\
\cdashline{1-8}
Yi-1.5-6B-Chat & 27.39 & 31.57 & 30.96 & 29.97 & 16.80& 13.17 & 78.39 \\
Mistral-7B-Instruct & 8.86 & 10.29 & 8.96 & 9.37 & 4.38 & \underline{4.99} & 113.9 \\
Qwen2-7B-Instruct & \underline{39.51} & \underline{41.96} & \underline{39.41} & \underline{40.29} & \underline{27.70} & 12.59 & \underline{45.45} \\
LLaMA3-8B-Instruct & 13.14 & 13.95 & 13.44 & 13.51 & 7.03 & 6.48 & 92.18 \\
Yi-1.5-9B-Chat & 37.27 & 37.88 & 38.19 & 37.78 & 23.42 & 14.36 & 61.32 \\
\hline
Mistral-Nemo-Instruct-2407 & 21.28 & 23.22 & 22.20 & 22.23 & 12.32 & 9.91 & 80.44 \\
DeepSeek-MOE-16B-Chat & 4.58 & 4.48 & 4.48 & 4.51 & 1.22 & \textbf{\underline{3.29}} & 269.67 \\
DeepSeek-V2-Lite-Chat & 11.30 & 12.53 & 12.02 & 11.95 & 4.38 & 7.57 & 172.83 \\
Mistral-Small-Instruct-2409 & \underline{39.41} & 40.73 & \underline{39.51} & 39.88 & \underline{26.37} & 13.51 & \underline{51.23}\\
Yi-1.5-34B-Chat & \underline{39.41} & \underline{41.34} & 39.21 & \underline{39.99} & 24.95 & 15.04 & 60.28 \\
\hline
LLaMA3-70B-Instruct & 32.89 & 36.46 & 34.21 & 34.52 & 22.81 & \underline{11.71} & 51.34\\
Qwen2-72B-Instruct & 47.86 & 50.20 & 48.88 & 48.98 & 33.50 & 15.48 & 46.21\\
Mistral-Large-Instruct-2407 & \underline{57.33} & \underline{59.17} & \underline{59.57} & \underline{58.69} & \underline{45.93} & 12.76 & \underline{27.78} \\
\hline
\multicolumn{7}{c}{{\textit{Specialized Mathematical LLMs}}} \\
\cdashline{1-8}
DeepSeek-Math-7B-Instruct & 23.83 & 27.39 & 23.83 & 25.02 & 13.75 & 11.27 & 81.96 \\
DeepSeek-Math-7B-RL & 30.04 & 30.65 & 30.65 & 30.45 & 20.06 & 10.39 & 51.79 \\
NuminaMath-7B-CoT & 32.08 & 32.89 & 32.79 & 32.59 & 20.16 & 12.43 & 61.66 \\
Mathstral-7B-v0.1 & 27.70 & 27.19 & 29.53 & 28.14 & 16.29 & 11.85 & 72.74 \\
Qwen2-Math-7B-Instruct & 46.03 & 49.49 & 49.49 & 48.34 & 35.74 & 12.60 & 35.25\\
Qwen2-Math-72B-Instruct & 61.71 & 61.81 & 60.08 & 61.20 & 49.29 & 11.91 & 24.16\\
\hline
\end{tabular}
}

\label{tabapp: calculus_singlevariable}
\end{table*}

\begin{table*}[!thb]
\centering
\footnotesize
\caption{\textbf{Main Results on Combinatorics} (all figures are in \%). Models are classified into three categories according to their purpose and origin. The best results within each column are \textbf{bolded} and the best results of open-source Chat LLMs within a similar parameter size group are \underline{underlined}.}
\scalebox{1.0}{
\begin{tabular}{lccccccc}
\hline
\textbf{Models} & \textbf{$\text{Acc}_1$} & \textbf{$\text{Acc}_2$} & \textbf{$\text{Acc}_3$} & \textbf{AAcc} & \textbf{EAcc} & \textbf{$\Delta$} & \textbf{RE}\\
\hline
\multicolumn{7}{c}{{\textit{Closed-source LLMs}}} \\
\cdashline{1-8}
OpenAI-o1-mini-2024-09-12 & \textbf{90.91} & \textbf{92.05} & \textbf{95.45} & \textbf{92.80} & \textbf{88.64} & \textbf{4.16} & \textbf{4.69} \\
GPT-4o-2024-08-06 & 77.27 & 77.27 & 78.41 & 77.65 & 61.36 & 16.29 & 26.55 \\
GPT-4o-mini-2024-07-18 & 70.45 & 69.32 & 73.86 & 71.21 & 56.82 & 14.39 & 25.33 \\
Claude-3-Opus-20240229 & 56.82 & 65.91 & 68.18 & 63.64 & 50.00 & 13.64 & 27.28 \\
\hline
\multicolumn{7}{c}{{\textit{Open-source Chat LLMs}}} \\
\cdashline{1-8}
Yi-1.5-6B-Chat & 25.00 & 34.09 & 29.55 & 29.55 & 18.18 & 11.37 & 62.54 \\
Mistral-7B-Instruct & 10.23 & 12.50 & 11.36 & 11.36 & 2.27 & 9.09 & 400.4\\
Qwen2-7B-Instruct & \underline{39.77} & \underline{45.45} & \underline{44.32} & \underline{43.18} & \underline{28.41} & 14.77 & \underline{51.99} \\
LLaMA3-8B-Instruct & 22.73 & 23.86 & 23.86 & 23.48 & 14.77 & \underline{8.71} & 58.97 \\
Yi-1.5-9B-Chat & 32.95 & 43.18 & \underline{44.32} & 40.15 & 25.00 & 15.15 & 60.60\\
\hline
Mistral-Nemo-Instruct-2407 & 26.14 & 30.68 & 32.95 & 29.92 & 18.18 & 11.74 & 64.58 \\
DeepSeek-MOE-16B-Chat & 6.82 & 11.36 & 7.95 & 8.71 & 2.27 & \underline{6.44} & 283.7 \\
DeepSeek-V2-Lite-Chat & 17.05 & 22.73 & 12.50& 17.42 & 6.82 & 10.60 & 155.43 \\
Mistral-Small-Instruct-2409 & \underline{46.59} & \underline{59.09} & 50.00 & \underline{51.89} & \underline{38.64} & 13.25 & \underline{34.29} \\
Yi-1.5-34B-Chat & 37.50  & 48.86 & \underline{52.27} & 46.21 & 27.27 & 18.94 & 69.45 \\
\hline
LLaMA3-70B-Instruct & 47.73 & 52.27& 45.45 & 48.48 & 37.50 & \underline{10.98} & 29.28 \\
Qwen2-72B-Instruct & 60.23 & 63.64 & 60.23 & 61.36 & 47.73 & 13.63 & 28.56\\
Mistral-Large-Instruct-2407 & \underline{69.32} & \underline{72.73} & \underline{71.59} & \underline{71.21} & \underline{57.95} & 13.26 & \underline{22.88} \\
\hline
\multicolumn{7}{c}{{\textit{Specialized Mathematical LLMs}}} \\
\cdashline{1-8}
DeepSeek-Math-7B-Instruct & 26.14 & 37.50 & 27.27 & 30.30 & 17.05 & 13.25 & 77.71 \\
DeepSeek-Math-7B-RL & 30.68 & 42.05 & 39.77 & 37.50 & 26.14 & 11.36 & 43.46 \\
NuminaMath-7B-CoT & 30.68 & 35.23 & 37.50 & 34.47 & 23.86 & 10.61 & 44.47 \\
Mathstral-7B-v0.1 & 34.09 & 38.64 & 35.23 & 35.98 & 26.14 & 9.84 & 37.64 \\
Qwen2-Math-7B-Instruct & 62.50 & 62.50 & 64.77 & 63.26 & 47.73 & 15.53 & 32.54\\
Qwen2-Math-72B-Instruct & 76.14 & 76.14 & 75.00 & 75.76 & 62.50 & 13.26 & 21.22\\
\hline
\end{tabular}
}

\label{tabapp: combinatorics}
\end{table*}

\begin{table*}[!thb]
\centering
\footnotesize
\caption{\textbf{Main Results on Complex analysis} (all figures are in \%).
The best results within each column are \textbf{bolded} and the best results of open-source Chat LLMs within a similar parameter size group are \underline{underlined}.} 
\scalebox{1.0}{
\begin{tabular}{lccccccc}
\hline
\textbf{Models} & \textbf{$\text{Acc}_1$} & \textbf{$\text{Acc}_2$} & \textbf{$\text{Acc}_3$} & \textbf{AAcc} & \textbf{EAcc} & \textbf{$\Delta$} & \textbf{RE}\\
\hline
\multicolumn{7}{c}{{\textit{Closed-source LLMs}}} \\
\cdashline{1-8}
OpenAI-o1-mini-2024-09-12 & \textbf{74.51} & \textbf{74.51} & \textbf{76.47} & \textbf{75.16} & \textbf{64.71} & 10.45 & 16.15 \\
GPT-4o-2024-08-06 & 66.67 & 70.59 & 70.59 & 69.28 & 58.82 & 10.46 & 17.78 \\
GPT-4o-mini-2024-07-18 & 62.75 & 62.75 & 62.75 & 62.75 & 50.98 & 11.77 & 23.09 \\
Claude-3-Opus-20240229 & 47.06 & 49.02 & 50.98 & 49.02 & 37.25 & 11.77 & 31.60 \\
\hline
\multicolumn{7}{c}{{\textit{Open-source Chat LLMs}}} \\
\cdashline{1-8}
Yi-1.5-6B-Chat & 35.29 & 37.25 & 31.37 & 34.64 & 23.53 & 11.11 & 47.22 \\
Mistral-7B-Instruct & 11.76 & 11.76 & 9.80 & 11.11 & 5.88 & 5.23 & 88.95 \\
Qwen2-7B-Instruct & \underline{45.10} & \underline{50.98} & \underline{43.14} & \underline{46.41} & \underline{35.29} & 11.12 & \underline{31.51}\\
LLaMA3-8B-Instruct & 19.61 & 17.65 & 17.65 & 18.30 & 13.73 & \underline{4.57} & 33.28\\
Yi-1.5-9B-Chat & 39.22 & 39.22 & 39.22 & 39.22 & 27.45 & 11.77 & 42.88 \\
\hline
Mistral-Nemo-Instruct-2407 & 29.41& 25.49 & 25.49 & 26.80 & 19.61 & 7.19 & 36.66\\
DeepSeek-MOE-16B-Chat & 1.96 & 3.92 & 1.96 & 2.61& 0.00 & \textbf{\underline{2.61}} & $\infty$ \\
DeepSeek-V2-Lite-Chat & 17.65 & 13.73 & 15.69 & 15.69 & 7.84 & 7.85 & 100.1\\
Mistral-Small-Instruct-2409 & \underline{47.06} & \underline{49.02} & \underline{43.14} & \underline{46.41} & \underline{37.25} & 9.16 & \underline{24.59} \\
Yi-1.5-34B-Chat & \underline{47.06} & 45.10 & 39.22 & 43.79 & 25.49 & 18.30 & 71.79 \\
\hline
LLaMA3-70B-Instruct & 43.14 & 33.33 & 37.25 & 37.91 & 25.49 & 12.42 & 48.72 \\
Qwen2-72B-Instruct & 52.94 & 62.75 & 62.75 & 59.48 & 47.06 & 12.42 & 26.39\\
Mistral-Large-Instruct-2407 & \underline{64.71} & \underline{64.71} & \underline{60.78} & \underline{63.40} & \underline{52.94} & \underline{10.46} & \underline{19.76} \\
\hline
\multicolumn{7}{c}{{\textit{Specialized Mathematical LLMs}}} \\
\cdashline{1-8}
DeepSeek-Math-7B-Instruct & 35.29 & 29.41 & 27.45 & 30.72 & 23.53 & 7.19 & 30.56 \\
DeepSeek-Math-7B-RL & 35.29 & 41.18 & 47.06 & 41.18 & 27.45 & 13.73 & 50.02 \\
NuminaMath-7B-CoT & 33.33 & 35.29 & 37.25 & 35.29 & 21.57 & 13.72 & 63.61 \\
Mathstral-7B-v0.1 & 37.25 & 29.41 & 31.37 & 32.68 & 15.69 & 16.99 & 108.3 \\
Qwen2-Math-7B-Instruct & 56.86 & 49.02 & 52.94 & 52.94 & 41.18 & 11.76 & 28.56\\
Qwen2-Math-72B-Instruct & 70.59 & 72.55 & 72.55 & 71.90 & 64.71 & 7.19 & \textbf{11.11}\\
\hline
\end{tabular}
}

\label{tabapp: complex_analysis}
\end{table*}

\begin{table*}[!thb]
\centering
\footnotesize
\caption{\textbf{Main Results on Differential equations} (all figures are in \%). Models are classified into three categories according to their purpose and origin. The best results within each column are \textbf{bolded} and the best results of open-source Chat LLMs within a similar parameter size group are \underline{underlined}.}
\scalebox{1.0}{
\begin{tabular}{lccccccc}
\hline
\textbf{Models} & \textbf{$\text{Acc}_1$} & \textbf{$\text{Acc}_2$} & \textbf{$\text{Acc}_3$} & \textbf{AAcc} & \textbf{EAcc} & \textbf{$\Delta$} & \textbf{RE}\\
\hline
\multicolumn{7}{c}{{\textit{Closed-source LLMs}}} \\
\cdashline{1-8}
OpenAI-o1-mini-2024-09-12 & \textbf{29.84} & \textbf{29.18} & \textbf{29.51} & \textbf{29.51} & \textbf{22.62} & 6.89 & 30.46 \\
GPT-4o-2024-08-06 & 23.93 & 24.59 & 25.57 & 24.70 & 19.34 & 5.36 & \textbf{27.71} \\
GPT-4o-mini-2024-07-18 & 20.00 & 22.30 & 19.67 & 20.66 & 16.07 & 4.59 & 28.56 \\
Claude-3-Opus-20240229 & 18.69 & 18.03 & 20.33 & 19.02 & 13.11 & 5.91 & 45.08 \\
\hline
\multicolumn{7}{c}{{\textit{Open-source Chat LLMs}}} \\
\cdashline{1-8}
Yi-1.5-6B-Chat & 6.23 & 9.51 & 4.59 & 6.78 & 2.95 & 3.83 & 129.8 \\
Mistral-7B-Instruct & 3.28 & 3.93 & 4.59 & 3.93 & 1.64 & \underline{2.29} & 139.6 \\
Qwen2-7B-Instruct & \underline{11.80}& 12.13 & 11.48 & 11.80 & \underline{7.54} & 4.26 & \underline{56.50}\\
LLaMA3-8B-Instruct & 4.26 & 5.90 & 5.57 & 5.25 & 2.62 & 2.63 & 100.4\\
Yi-1.5-9B-Chat & \underline{11.80} & \underline{13.44} & \underline{12.13} & \underline{12.46} & 6.23 & 6.23 & 100.0 \\
\hline
Mistral-Nemo-Instruct-2407 & 8.52 & 7.87 & 9.18 & 8.52 & 4.26 & 4.26 & 100.0\\
DeepSeek-MOE-16B-Chat & 0.66& 1.31& 0.00 & 0.66 & 0.00 & \textbf{\underline{0.66}} & $\infty$ \\
DeepSeek-V2-Lite-Chat & 2.62 & 2.62 & 1.97 & 2.40 & 0.33 & 2.07 & 627.3 \\
Mistral-Small-Instruct-2409 & \underline{15.74} & \underline{15.74} & \underline{14.43} & \underline{15.30} & \underline{9.84} & 5.46 & \underline{55.49} \\
Yi-1.5-34B-Chat & 12.13 & 15.08 & 12.13 & 13.11 & 7.21 & 5.90 & 81.83 \\
\hline
LLaMA3-70B-Instruct & 12.79 & 13.44 & 12.13 & 12.79 & 7.54 & \underline{5.25} & 69.63 \\
Qwen2-72B-Instruct & 19.34 & 20.00 & 18.36 & 19.23 & 13.77 & 5.46 & 39.65\\
Mistral-Large-Instruct-2407 & \underline{22.95} & \underline{25.90} & \underline{24.92} & \underline{24.59} & \underline{18.69} & 5.90 & \underline{31.57} \\
\hline
\multicolumn{7}{c}{{\textit{Specialized Mathematical LLMs}}} \\
\cdashline{1-8}
DeepSeek-Math-7B-Instruct & 6.89 & 7.54 & 5.90 & 6.78 & 1.97 & 4.81 & 244.2 \\
DeepSeek-Math-7B-RL & 8.52 & 10.82 & 9.84 & 9.73 & 5.25 & 4.48 & 85.33\\
NuminaMath-7B-CoT & 10.82 & 10.16 & 7.87 & 9.62 & 4.26 & 5.36 & 125.8\\
Mathstral-7B-v0.1 & 10.49 & 10.49 & 9.51 & 10.16 & 4.92 & 5.24 & 106.5\\
Qwen2-Math-7B-Instruct & 15.41 & 18.69 & 18.03 & 17.38 & 12.13 & 5.25 & 43.28\\
Qwen2-Math-72B-Instruct & 22.95 & 22.95 & 22.95 & 22.95 & 17.38 & 5.57 & 32.05\\
\hline
\end{tabular}
}

\label{tabapp: differential_equations}
\end{table*}

\begin{table*}[!thb]
\centering
\footnotesize
\caption{\textbf{Main Results on Financial mathematics} (all figures are in \%). Models are classified into three categories according to their purpose and origin. The best results within each column are \textbf{bolded} and the best results of open-source Chat LLMs within a similar parameter size group are \underline{underlined}.}
\scalebox{1.0}{
\begin{tabular}{lccccccc}
\hline
\textbf{Models} & \textbf{$\text{Acc}_1$} & \textbf{$\text{Acc}_2$} & \textbf{$\text{Acc}_3$} & \textbf{AAcc} & \textbf{EAcc} & \textbf{$\Delta$} & \textbf{RE}\\
\hline
\multicolumn{7}{c}{{\textit{Closed-source LLMs}}} \\
\cdashline{1-8}
OpenAI-o1-mini-2024-09-12 & \textbf{46.53} & \textbf{49.13} & \textbf{51.16} & \textbf{48.94} & \textbf{28.32} & 20.62 & 72.81 \\
GPT-4o-2024-08-06 & 28.61 & 29.77 & 29.77 & 29.38 & 18.21& 11.17 & 61.34 \\
GPT-4o-mini-2024-07-18 & 16.19 & 17.63 & 19.65 & 17.82 & 9.54 & 8.28 & 86.79 \\
Claude-3-Opus-20240229 & 19.65 & 21.39 & 20.52 & 20.52 & 10.98 & 9.54 & 86.89 \\
\hline
\multicolumn{7}{c}{{\textit{Open-source Chat LLMs}}} \\
\cdashline{1-8}
Yi-1.5-6B-Chat & 5.49 & 6.94 & 6.94 & 6.45 & 3.18 & 3.27 & \underline{102.8}\\
Mistral-7B-Instruct & 1.16 & 3.47 & 2.02 & 2.22 & 0.87& \underline{1.35} & 155.2\\
Qwen2-7B-Instruct & 8.67 & 12.43 & 8.96 & 10.02 & 3.76 & 6.26 & 166.5 \\
LLaMA3-8B-Instruct & 2.60 & 2.02 & 3.18 & 2.60 & 1.16 & 1.44 & 124.1 \\
Yi-1.5-9B-Chat & \underline{9.25} & \underline{14.45} & \underline{12.43} & \underline{12.04} & \underline{4.05} & 7.99 & 197.2\\
\hline
Mistral-Nemo-Instruct-2407 & 4.62 & 5.49 & 6.07 & 5.39 & 1.73 & 3.66 & 211.6 \\
DeepSeek-MOE-16B-Chat & 0.58 & 0.87 & 1.16 & 0.87 & 0.00 & \textbf{\underline{0.87}} & $\infty$ \\
DeepSeek-V2-Lite-Chat & 3.76 & 3.76 & 3.76 & 3.76 & 1.45 & 2.31 & 159.3 \\
Mistral-Small-Instruct-2409 & \underline{12.43} & 11.85 & 13.01 & 12.43 & \underline{5.78} & 6.65 & \underline{115.1}\\
Yi-1.5-34B-Chat & 11.85 & \underline{14.45} & \underline{13.58} & \underline{13.29} & 5.20 & 8.09 & 155.6\\
\hline
LLaMA3-70B-Instruct & 8.09 & 10.12 & 9.25 & 9.15 & 4.62 & \underline{4.53} & 98.05 \\
Qwen2-72B-Instruct & 15.03 & 18.50 & 16.47 & 16.67 & 8.96 & 7.71 & 86.05\\
Mistral-Large-Instruct-2407 & \underline{22.54} & \underline{26.59} & \underline{21.97} & \underline{23.70} & \underline{15.32} & 8.38 & \underline{\textbf{54.70}} \\
\hline
\multicolumn{7}{c}{{\textit{Specialized Mathematical LLMs}}} \\
\cdashline{1-8}
DeepSeek-Math-7B-Instruct & 4.91 & 6.07 & 4.62 & 5.20 & 2.31 & 2.89 & 125.1\\
DeepSeek-Math-7B-RL & 5.49 & 7.51 & 5.78 & 6.26 & 2.02 & 4.24 & 209.9 \\
NuminaMath-7B-CoT & 8.67 & 9.25 & 9.25 & 9.06 & 3.76 & 5.30 & 141.0\\
Mathstral-7B-v0.1 & 4.91 & 9.83 &7.23 & 7.32 & 3.18 & 4.14 & 130.2\\
Qwen2-Math-7B-Instruct & 10.69 & 17.05 & 15.32 & 14.35 & 5.49 & 8.86 & 161.38\\
Qwen2-Math-72B-Instruct & 28.90 & 26.88 & 29.19 & 28.32 & 16.47 & 11.85 & 71.95\\
\hline
\end{tabular}
}

\label{tabapp: financial_mathematics}
\end{table*}

\begin{table*}[!thb]
\centering
\footnotesize
\caption{\textbf{Main Results on Geometry} (all figures are in \%). Models are classified into three categories according to their purpose and origin. The best results within each column are \textbf{bolded} and the best results of open-source Chat LLMs within a similar parameter size group are \underline{underlined}.}

\scalebox{1.0}{
\begin{tabular}{lccccccc}
\hline
\textbf{Models} & \textbf{$\text{Acc}_1$} & \textbf{$\text{Acc}_2$} & \textbf{$\text{Acc}_3$} & \textbf{AAcc} & \textbf{EAcc} & \textbf{$\Delta$} & \textbf{RE}\\
\hline
\multicolumn{7}{c}{{\textit{Closed-source LLMs}}} \\
\cdashline{1-8}
OpenAI-o1-mini-2024-09-12 & \textbf{71.43} & \textbf{67.08} & \textbf{71.43} & \textbf{69.98} & \textbf{58.39} & 11.59 & 19.85\\
GPT-4o-2024-08-06 & 64.60 & 63.98 & 63.98 & 64.18 & 51.55 & 12.63 & 24.50\\
GPT-4o-mini-2024-07-18 & 60.87 & 55.90 & 59.63 & 58.80 & 44.10 & 14.70 & 33.33\\
Claude-3-Opus-20240229 & 57.76 & 49.69 & 49.07 & 52.17 & 37.89 & 14.28 & 37.69\\
\hline
\multicolumn{7}{c}{{\textit{Open-source Chat LLMs}}} \\
\cdashline{1-8}
Yi-1.5-6B-Chat & 32.92 & 27.33 & 30.43 & 30.23 & 18.01 & 12.22 & 67.85\\
Mistral-7B-Instruct & 8.70 & 8.70 & 6.83 & 8.07 & 3.11 & \underline{4.96} & 159.5\\
Qwen2-7B-Instruct & \underline{50.31} & \underline{39.75} & \underline{42.86} & \underline{44.31} & \underline{29.19} & 15.12 & \underline{51.80}\\
LLaMA3-8B-Instruct & 24.84 & 19.88 & 19.88 & 21.53 & 9.94 & 11.59 & 116.6\\
Yi-1.5-9B-Chat & 38.51 & 37.89 & 36.02 & 37.47 & 22.36 & 15.11 & 67.58\\
\hline
Mistral-Nemo-Instruct-2407 & 31.68 & 27.95 & 28.57 & 29.40 & 16.77 & 12.63 & 75.31\\
DeepSeek-MOE-16B-Chat & 4.97 & 3.73 & 6.21 & 4.97 & 1.86 & \underline{\textbf{3.11}} & 167.2\\
DeepSeek-V2-Lite-Chat & 11.18 & 14.29 & 14.91 & 13.46 & 4.97 & 8.49 & 170.8\\
Mistral-Small-Instruct-2409 & \underline{44.72} & \underline{47.21} & \underline{46.58} & \underline{46.17} & \underline{32.92} & 13.25 & \underline{40.25}\\
Yi-1.5-34B-Chat & 43.48 & 37.89 & 44.72 & 42.03 & 27.33 & 14.70 & 53.79\\
\hline
LLaMA3-70B-Instruct & 39.13 & 38.51 & 39.75 & 39.13 & 25.47 & 13.66 & 53.63\\
Qwen2-72B-Instruct & 59.01 & 49.69 & 53.42 & 54.04 & 40.99 & 13.05 & 31.84\\
Mistral-Large-Instruct-2407 & \underline{61.49} & \underline{57.76} & \underline{60.25} & \underline{59.83} & \underline{50.31} & \underline{9.52} & \underline{\textbf{18.92}}\\
\hline
\multicolumn{7}{c}{{\textit{Specialized Mathematical LLMs}}} \\
\cdashline{1-8}
DeepSeek-Math-7B-Instruct & 29.19 & 24.22 & 24.22 & 25.88 & 14.29 & 11.59 & 81.11\\
DeepSeek-Math-7B-RL & 34.78 & 29.19 & 29.81 & 31.26 & 19.25 & 12.01 & 62.39\\
NuminaMath-7B-CoT & 33.54 & 31.68 & 28.57 & 31.26 & 19.25 & 12.01 & 62.39\\
Mathstral-7B-v0.1 & 39.13 & 32.92 & 37.89 & 36.65 & 24.22 & 12.43 & 51.32\\
Qwen2-Math-7B-Instruct & 52.80 & 42.86 & 45.96 & 47.20 & 34.16 & 13.04 & 38.17\\
Qwen2-Math-72B-Instruct & 67.08 & 61.49 & 64.60 & 64.39 & 53.42 & 10.97 & 20.54\\
\hline
\end{tabular}
}
\label{tabapp: geometry}
\end{table*}

\begin{table*}[!thb]
\centering
\footnotesize
\caption{\textbf{Main Results on Linear algebra} (all figures are in \%). Models are classified into three categories according to their purpose and origin. The best results within each column are \textbf{bolded} and the best results of open-source Chat LLMs within a similar parameter size group are \underline{underlined}.}
\scalebox{1.0}{
\begin{tabular}{lccccccc}
\hline
\textbf{Models} & \textbf{$\text{Acc}_1$} & \textbf{$\text{Acc}_2$} & \textbf{$\text{Acc}_3$} & \textbf{AAcc} & \textbf{EAcc} & \textbf{$\Delta$} & \textbf{RE}\\
\hline
\multicolumn{7}{c}{{\textit{Closed-source LLMs}}} \\
\cdashline{1-8}
OpenAI-o1-mini-2024-09-12 & \textbf{74.90} & \textbf{74.90} & \textbf{75.90} & \textbf{75.23} & \textbf{65.86} & 9.37 & \textbf{14.23}\\
GPT-4o-2024-08-06 & 63.86 & 61.65 & 61.65 & 62.38 & 51.61 & 10.77 & 20.87\\
GPT-4o-mini-2024-07-18 & 57.63 & 57.03 & 56.22 & 56.96 & 43.37 & 13.59 & 31.34\\
Claude-3-Opus-20240229 & 51.20 & 48.80 & 49.00 & 49.67 & 37.15 & 12.52 & 33.70\\
\hline
\multicolumn{7}{c}{{\textit{Open-source Chat LLMs}}} \\
\cdashline{1-8}
Yi-1.5-6B-Chat & 23.69 & 23.69 & 21.49 & 22.96 & 12.45 & 10.51 & 84.42\\
Mistral-7B-Instruct & 8.03 & 7.43 & 7.23 & 7.56 & 2.61 & \underline{4.95} & 189.7\\
Qwen2-7B-Instruct & \underline{36.35} & \underline{34.54} & \underline{32.73} & \underline{34.54} & \underline{22.29} & 12.25 & \underline{54.96}\\
LLaMA3-8B-Instruct & 13.25 & 12.45 & 12.65 & 12.78 & 6.43 & 6.35 & 98.76\\
Yi-1.5-9B-Chat & 34.14 & 31.93 & 31.93 & 32.66 & 18.47 & 14.19 & 76.83\\
\hline
Mistral-Nemo-Instruct-2407 & 23.69 & 20.48 & 23.90 & 22.69 & 11.24 & 11.45 & 101.9\\
DeepSeek-MOE-16B-Chat & 2.01 & 2.41 & 3.61 & 2.68 & 0.80 & \underline{\textbf{1.88}} & 235.0\\
DeepSeek-V2-Lite-Chat & 9.24 & 8.23 & 7.43 & 8.30 & 1.81 & 6.49 & 358.6\\
Mistral-Small-Instruct-2409 & \underline{43.17} & \underline{38.35} & \underline{41.57} & \underline{41.03} & \underline{28.51} & 12.52 & \underline{43.91}\\
Yi-1.5-34B-Chat & 35.34 & 31.93 & 34.94 & 34.07 & 20.08 & 13.99 & 69.67\\
\hline
LLaMA3-70B-Instruct & 34.34 & 30.72 & 29.32 & 31.46 & 19.28 & 12.18 & 63.17\\
Qwen2-72B-Instruct & 52.81 & 51.00 & 49.80 & 51.20 & 38.15 & 13.05 & 34.21\\
Mistral-Large-Instruct-2407 & \underline{62.25} & \underline{59.04} & \underline{58.43} & \underline{59.91} & \underline{48.59} & \underline{11.32} & \underline{23.30}\\
\hline
\multicolumn{7}{c}{{\textit{Specialized Mathematical LLMs}}} \\
\cdashline{1-8}
DeepSeek-Math-7B-Instruct & 20.08 & 21.08 & 18.07 & 19.75 & 9.44 & 10.31 & 109.2\\
DeepSeek-Math-7B-RL & 26.91 & 26.51 & 23.09 & 25.50 & 15.46 & 10.04 & 64.94\\
NuminaMath-7B-CoT & 25.70 & 25.50 & 27.11 & 26.10 & 14.66 & 11.44 & 78.04\\
Mathstral-7B-v0.1 & 29.92 & 23.90 & 24.70 & 26.17 & 13.86 & 12.31 & 88.82\\
Qwen2-Math-7B-Instruct & 46.39 & 43.17 & 43.57 & 44.38 & 32.73 & 11.65 & 35.59\\
Qwen2-Math-72B-Instruct & 61.45 & 59.24 & 62.85 & 61.18 & 47.59 & 13.59 & 28.56\\
\hline
\end{tabular}
}

\label{tabapp: linear_algebra}
\end{table*}

\begin{table*}[!thb]
\centering
\footnotesize
\caption{\textbf{Main Results on Number theory} (all figures are in \%). Models are classified into three categories according to their purpose and origin. The best results within each column are \textbf{bolded} and the best results of open-source Chat LLMs within a similar parameter size group are \underline{underlined}.}
\scalebox{1.0}{
\begin{tabular}{lccccccc}
\hline
\textbf{Models} & \textbf{$\text{Acc}_1$} & \textbf{$\text{Acc}_2$} & \textbf{$\text{Acc}_3$} & \textbf{AAcc} & \textbf{EAcc} & \textbf{$\Delta$} & \textbf{RE}\\
\hline
\multicolumn{7}{c}{{\textit{Closed-source LLMs}}} \\
\cdashline{1-8}
OpenAI-o1-mini-2024-09-12 & \textbf{80.43} & \textbf{76.09} & \textbf{73.91} & \textbf{76.81} & \textbf{58.70} & 18.11 & 30.85\\
GPT-4o-2024-08-06 & 60.87 & 63.04 & 67.39 & 63.77 & 56.52 & 7.25 & \textbf{12.83}\\
GPT-4o-mini-2024-07-18 & 60.87 & 58.70 & 58.70 & 59.42 & 52.17 & 7.25 & 13.90\\
Claude-3-Opus-20240229 & 47.83 & 54.35 & 56.52 & 52.90 & 39.13 & 13.77 & 35.19\\
\hline
\multicolumn{7}{c}{{\textit{Open-source Chat LLMs}}} \\
\cdashline{1-8}
Yi-1.5-6B-Chat & 26.09 & 26.09 & 19.57 & 23.91 & 13.04 & 10.87 & 83.36\\
Mistral-7B-Instruct & 6.52 & 6.52 & 8.70 & 7.25 & 2.17 & \underline{5.08} & 234.1\\
Qwen2-7B-Instruct & \underline{34.78} & \underline{39.13} & \underline{41.30} & \underline{38.41} & \underline{23.91} & 14.50 & \underline{60.64}\\
LLaMA3-8B-Instruct & 19.57 & 21.74 & 15.22 & 18.84 & 8.70 & 10.14 & 116.6\\
Yi-1.5-9B-Chat & \underline{34.78} & 32.61 & 34.78 & 34.06 & \underline{23.91} & 10.15 & 42.45\\
\hline
Mistral-Nemo-Instruct-2407 & 17.39 & 28.26 & 28.26 & 24.64 & 8.70 & 15.94 & 183.2\\
DeepSeek-MOE-16B-Chat & 6.52 & 6.52 & 2.17 & 5.07 & 2.17 & \underline{\textbf{2.90}} & 133.6\\
DeepSeek-V2-Lite-Chat & 15.22 & 13.04 & 15.22 & 14.49 & 4.35 & 10.14 & 233.1\\
Mistral-Small-Instruct-2409 & \underline{45.65} & \underline{34.78} & 41.30 & \underline{40.58} & \underline{30.43} & 10.15 & 33.36\\
Yi-1.5-34B-Chat & 36.96 & \underline{34.78} & \underline{47.83} & 39.86 & 26.09 & 13.77 & \underline{52.78}\\
\hline
LLaMA3-70B-Instruct & 30.43 & 23.91 & 36.96 & 30.43 & 15.22 & 15.21 & 99.93\\
Qwen2-72B-Instruct & 47.83 & \underline{58.70} & 43.48 & 50.00 & 36.96 & \underline{13.04} & 35.28\\
Mistral-Large-Instruct-2407 & \underline{58.70} & \underline{58.70} & \underline{58.70} & \underline{58.70} & \underline{45.65} & 13.05 & \underline{28.59}\\
\hline
\multicolumn{7}{c}{{\textit{Specialized Mathematical LLMs}}} \\
\cdashline{1-8}
DeepSeek-Math-7B-Instruct & 23.91 & 28.26 & 30.43 & 27.54 & 15.22 & 12.32 & 80.95\\
DeepSeek-Math-7B-RL & 21.74 & 26.09 & 26.09 & 24.64 & 13.04 & 11.60 & 88.96\\
NuminaMath-7B-CoT & 17.39 & 32.61 & 32.61 & 27.54 & 13.04 & 14.50 & 111.2\\
Mathstral-7B-v0.1 & 28.26 & 30.43 & 30.43 & 29.71 & 17.39 & 12.32 & 70.85\\
Qwen2-Math-7B-Instruct & 39.13 & 43.48 & 43.48 & 42.03 & 30.43 & 11.60 & 38.12\\
Qwen2-Math-72B-Instruct & 60.87 & 54.35 & 63.04 & 59.42 & 50.00 & 9.42 & 18.84\\
\hline
\end{tabular}
}

\label{tabapp: number_theory}
\end{table*}

\begin{table*}[!thb]
\centering
\footnotesize
\caption{\textbf{Main Results on Probability} (all figures are in \%). Models are classified into three categories according to their purpose and origin. The best results within each column are \textbf{bolded} and the best results of open-source Chat LLMs within a similar parameter size group are \underline{underlined}.}
\scalebox{1.0}{
\begin{tabular}{lccccccc}
\hline
\textbf{Models} & \textbf{$\text{Acc}_1$} & \textbf{$\text{Acc}_2$} & \textbf{$\text{Acc}_3$} & \textbf{AAcc} & \textbf{EAcc} & \textbf{$\Delta$} & \textbf{RE}\\
\hline
\multicolumn{7}{c}{{\textit{Closed-source LLMs}}} \\
\cdashline{1-8}
OpenAI-o1-mini-2024-09-12 & \textbf{66.07} & \textbf{68.45} & \textbf{68.15} & \textbf{67.56} & 51.49 & 16.07 & 31.21\\
GPT-4o-2024-08-06 & 66.07 & 66.07 & 65.77 & 65.97 & \textbf{54.76} & 11.21 & \textbf{20.47}\\
GPT-4o-mini-2024-07-18 & 50.89 & 56.85 & 52.38 & 53.37 & 39.88 & 13.49 & 33.83\\
Claude-3-Opus-20240229 & 56.55 & 59.23 & 59.82 & 58.53 & 44.94 & 13.59 & 30.24\\
\hline
\multicolumn{7}{c}{{\textit{Open-source Chat LLMs}}} \\
\cdashline{1-8}
Yi-1.5-6B-Chat & 20.24 & 22.62 & 21.43 & 21.43 & 10.42 & 11.01 & 105.7\\
Mistral-7B-Instruct & 10.12 & 11.01 & 10.42 & 10.52 & 3.57 & \underline{6.95} & 194.7\\
Qwen2-7B-Instruct & \underline{31.85} & \underline{39.58} & \underline{33.63} & \underline{35.02} & \underline{20.54} & 14.48 & \underline{70.50}\\
LLaMA3-8B-Instruct & 16.37 & 16.96 & 15.48 & 16.27 & 9.23 & 7.04 & 76.27\\
Yi-1.5-9B-Chat & 28.27 & 30.06 & 31.85 & 30.06 & 13.69 & 16.37 & 119.6\\
\hline
Mistral-Nemo-Instruct-2407 & 23.81 & 26.79 & 25.00 & 25.20 & 15.18 & 10.02 & 66.01\\
DeepSeek-MOE-16B-Chat & 5.65 & 4.76 & 4.17 & 4.86 & 1.19 & \underline{\textbf{3.67}} & 308.4\\
DeepSeek-V2-Lite-Chat & 9.82 & 11.61 & 10.42 & 10.62 & 3.27 & 7.35 & 224.8\\
Mistral-Small-Instruct-2409 & \underline{39.58} & \underline{43.15} & \underline{43.45} & \underline{42.06} & \underline{27.68} & 14.38 & 51.95\\
Yi-1.5-34B-Chat & 37.20 & 41.67 & 38.39 & 39.09 & 23.51 & 15.58 & \underline{66.27}\\
\hline
LLaMA3-70B-Instruct & 35.12 & 36.31 & 34.23 & 35.22 & 22.02 & \underline{13.20} & 59.95\\
Qwen2-72B-Instruct & 48.51 & 48.51 & 44.05 & 47.02 & 33.04 & 13.98 & 42.31\\
Mistral-Large-Instruct-2407 & \underline{59.82} & \underline{60.12} & \underline{58.33} & \underline{59.42} & \underline{45.54} & 13.88 & \underline{30.48}\\
\hline
\multicolumn{7}{c}{{\textit{Specialized Mathematical LLMs}}} \\
\cdashline{1-8}
DeepSeek-Math-7B-Instruct & 21.73 & 22.92 & 21.43 & 22.02 & 10.71 & 11.31 & 105.6\\
DeepSeek-Math-7B-RL & 24.40 & 25.89 & 26.19 & 25.50 & 14.88 & 10.62 & 71.37\\
NuminaMath-7B-CoT & 25.60 & 26.49 & 25.30 & 25.79 & 13.39 & 12.40 & 92.61\\
Mathstral-7B-v0.1 & 27.38 & 29.46 & 25.30 & 27.38 & 14.88 & 12.50 & 84.01\\
Qwen2-Math-7B-Instruct & 36.61 & 44.35 & 39.88 & 40.28 & 25.89 & 14.39 & 55.58\\
Qwen2-Math-72B-Instruct & 58.33 & 62.20 & 59.52 & 60.02 & 44.05 & 15.97 & 36.25\\
\hline
\end{tabular}
}

\label{tabapp: probability}
\end{table*}

\begin{table*}[!thb]
\centering
\footnotesize
\caption{\textbf{Main Results on Set theory and logic} (all figures are in \%). Models are classified into three categories according to their purpose and origin. The best results within each column are \textbf{bolded} and the best results of open-source Chat LLMs within a similar parameter size group are \underline{underlined}.}
\scalebox{1.0}{
\begin{tabular}{lccccccc}
\hline
\textbf{Models} & \textbf{$\text{Acc}_1$} & \textbf{$\text{Acc}_2$} & \textbf{$\text{Acc}_3$} & \textbf{AAcc} & \textbf{EAcc} & \textbf{$\Delta$} & \textbf{RE}\\
\hline
\multicolumn{7}{c}{{\textit{Closed-source LLMs}}} \\
\cdashline{1-8}
OpenAI-o1-mini-2024-09-12 & \textbf{78.26} & 73.91 & \textbf{79.71} & \textbf{77.29} & \textbf{66.67} & 10.62 & \textbf{15.93}\\
GPT-4o-2024-08-06 & 75.36 & \textbf{76.81} & 68.12 & 73.43 & 59.42 & 14.01 & 23.58\\
GPT-4o-mini-2024-07-18 & 59.42 & 57.97 & 59.42 & 58.94 & 40.58 & 18.36 & 45.24\\
Claude-3-Opus-20240229 & 56.52 & 60.87 & 62.32 & 59.90 & 43.48 & 16.42 & 37.76\\
\hline
\multicolumn{7}{c}{{\textit{Open-source Chat LLMs}}} \\
\cdashline{1-8}
Yi-1.5-6B-Chat & 14.49 & 14.49 & 18.84 & 15.94 & 7.25 & 8.69 & 119.9\\
Mistral-7B-Instruct & 2.90 & 5.80 & 5.80 & 4.83 & 0.00 & \underline{4.83} & $\infty$\\
Qwen2-7B-Instruct & 20.29 & \underline{36.23} & 24.64 & \underline{27.05} & \underline{13.04} & 14.01 & 107.4\\
LLaMA3-8B-Instruct & 8.70 & 15.94 & 8.70 & 11.11 & 5.80 & 5.31 & \underline{91.55}\\
Yi-1.5-9B-Chat & \underline{24.64} & 18.84 & \underline{26.09} & 23.19 & 5.80 & 17.39 & 299.8\\
\hline
Mistral-Nemo-Instruct-2407 & 24.64 & 33.33 & 26.09 & 28.02 & 18.84 & 9.18 & 48.73\\
DeepSeek-MOE-16B-Chat & 2.90 & 1.45 & 1.45 & 1.93 & 0.00 & \underline{\textbf{1.93}} & $\infty$\\
DeepSeek-V2-Lite-Chat & 4.35 & 7.25 & 5.80 & 5.80 & 1.45 & 4.35 & 300.0\\
Mistral-Small-Instruct-2409 & \underline{47.83} & \underline{44.93} & 40.58 & \underline{44.44} & \underline{30.43} & 14.01 & \underline{46.04}\\
Yi-1.5-34B-Chat & 30.43 & 40.58 & \underline{43.48} & 38.16 & 21.74 & 16.42 & 75.53\\
\hline
LLaMA3-70B-Instruct & 28.99 & 26.09 & 24.64 & 26.57 & 15.94 & \underline{10.63} & 66.69\\
Qwen2-72B-Instruct & 52.17 & 55.07 & 49.28 & 52.17 & 36.23 & 15.94 & 44.00\\
Mistral-Large-Instruct-2407 & \underline{59.42} & \underline{60.87} & \underline{57.97} & \underline{59.42} & \underline{44.93} & 14.49 & \underline{32.25}\\
\hline
\multicolumn{7}{c}{{\textit{Specialized Mathematical LLMs}}} \\
\cdashline{1-8}
DeepSeek-Math-7B-Instruct & 21.74 & 18.84 & 13.04 & 17.87 & 8.70 & 9.17 & 105.4\\
DeepSeek-Math-7B-RL & 23.19 & 23.19 & 24.64 & 23.67 & 10.14 & 13.53 & 133.4\\
NuminaMath-7B-CoT & 17.39 & 26.09 & 18.84 & 20.77 & 8.70 & 12.07 & 138.7\\
Mathstral-7B-v0.1 & 23.19 & 26.09 & 20.29 & 23.19 & 14.49 & 8.70 & 60.04\\
Qwen2-Math-7B-Instruct & 31.88 & 39.13 & 27.54 & 32.85 & 18.84 & 14.01 & 74.36\\
Qwen2-Math-72B-Instruct & 60.87 & 56.52 & 55.07 & 57.49 & 40.58 & 16.91 & 41.67\\
\hline
\end{tabular}
}

\label{tabapp: set_theory}
\end{table*}

\begin{table*}[!thb]
\centering
\footnotesize
\caption{\textbf{Main Results on Statistics} (all figures are in \%). Models are classified into three categories according to their purpose and origin. The best results within each column are \textbf{bolded} and the best results of open-source Chat LLMs within a similar parameter size group are \underline{underlined}.}
\scalebox{1.0}{
\begin{tabular}{lccccccc}
\hline
\textbf{Models} & \textbf{$\text{Acc}_1$} & \textbf{$\text{Acc}_2$} & \textbf{$\text{Acc}_3$} & \textbf{AAcc} & \textbf{EAcc} & \textbf{$\Delta$} & \textbf{RE}\\
\hline
\multicolumn{7}{c}{{\textit{Closed-source LLMs}}} \\
\cdashline{1-8}
OpenAI-o1-mini-2024-09-12 & \textbf{64.09} & \textbf{62.34} & \textbf{61.85} & \textbf{62.76} & \textbf{49.88} & 12.88 & 25.82\\
GPT-4o-2024-08-06 & 59.10 & 62.09 & 60.10 & 60.43 & 49.38 & 11.05 & \textbf{22.38}\\
GPT-4o-mini-2024-07-18 & 43.14 & 46.38 & 48.63 & 46.05 & 32.92 & 13.13 & 39.88\\
Claude-3-Opus-20240229 & 52.12 & 53.12 & 50.37 & 51.87 & 37.16 & 14.71 & 39.59\\
\hline
\multicolumn{7}{c}{{\textit{Open-source Chat LLMs}}} \\
\cdashline{1-8}
Yi-1.5-6B-Chat & 15.96 & 16.96 & 21.45 & 18.12 & 6.23 & 11.89 & 190.9\\
Mistral-7B-Instruct & 11.97 & 11.72 & 11.47 & 11.72 & 2.99 & 8.73 & 291.0\\
Qwen2-7B-Instruct & \underline{26.68} & \underline{27.93} & 29.18 & \underline{27.93} & \underline{13.97} & 13.96 & 99.93\\
LLaMA3-8B-Instruct & 10.72 & 13.47 & 16.21 & 13.47 & 5.99 & \underline{7.48} & 124.9\\
Yi-1.5-9B-Chat & 26.43 & 23.94 & \underline{31.67} & 27.35 & \underline{13.97} & 13.38 & \underline{95.78}\\
\hline
Mistral-Nemo-Instruct-2407 & 22.94 & 27.18 & 24.94 & 25.02 & 14.71 & 10.31 & 70.09\\
DeepSeek-MOE-16B-Chat & 3.24 & 3.49 & 4.74 & 3.82 & 1.00 & \underline{\textbf{2.82}} & 282.0\\
DeepSeek-V2-Lite-Chat & 8.98 & 8.48 & 9.23 & 8.89 & 2.74 & 6.15 & 224.5\\
Mistral-Small-Instruct-2409 & \underline{37.66} & 35.91 & 35.66 & \underline{36.41} & \underline{23.69} & 12.72 & \underline{53.69}\\
Yi-1.5-34B-Chat & 32.92 & \underline{37.41} & \underline{37.91} & 36.08 & 21.45 & 14.63 & 68.21\\
\hline
LLaMA3-70B-Instruct & 22.44 & 20.70 & 21.45 & 21.53 & 11.97 & \underline{9.56} & 79.87\\
Qwen2-72B-Instruct & 43.14 & 44.39 & 44.89 & 44.14 & 31.42 & 12.72 & 40.48\\
Mistral-Large-Instruct-2407 & \underline{54.36} & \underline{50.12} & \underline{53.62} & \underline{52.70} & \underline{39.90} & 12.80 & \underline{32.08}\\
\hline
\multicolumn{7}{c}{{\textit{Specialized Mathematical LLMs}}} \\
\cdashline{1-8}
DeepSeek-Math-7B-Instruct & 18.95 & 16.96 & 18.95 & 18.29 & 6.73 & 11.56 & 171.8\\
DeepSeek-Math-7B-RL & 21.70 & 24.44 & 22.69 & 22.94 & 9.23 & 13.71 & 148.5\\
NuminaMath-7B-CoT & 17.71 & 18.20 & 18.95 & 18.29 & 6.48 & 11.81 & 182.3\\
Mathstral-7B-v0.1 & 19.70 & 22.69 & 21.70 & 21.36 & 9.73 & 11.63 & 119.5\\
Qwen2-Math-7B-Instruct & 31.42 & 31.17 & 37.66 & 33.42 & 17.71 & 15.71 & 88.71\\
Qwen2-Math-72B-Instruct & 47.38 & 50.87 & 47.63 & 48.63 & 33.92 & 14.71 & 43.37\\
\hline
\end{tabular}
}

\label{tabapp: statistics}
\end{table*}

\begin{table*}[!thb]
\centering
\footnotesize
\caption{\textbf{Main Results on Trigonometry} (all figures are in \%). Models are classified into three categories according to their purpose and origin. The best results within each column are \textbf{bolded} and the best results of open-source Chat LLMs within a similar parameter size group are \underline{underlined}.}
\scalebox{1.0}{
\begin{tabular}{lccccccc}
\hline
\textbf{Models} & \textbf{$\text{Acc}_1$} & \textbf{$\text{Acc}_2$} & \textbf{$\text{Acc}_3$} & \textbf{AAcc} & \textbf{EAcc} & \textbf{$\Delta$} & \textbf{RE}\\
\hline
\multicolumn{7}{c}{{\textit{Closed-source LLMs}}} \\
\cdashline{1-8}
OpenAI-o1-mini-2024-09-12 & \textbf{75.84} & \textbf{74.16} & \textbf{69.10} & \textbf{73.03} & \textbf{56.74} & 16.29 & 28.71\\
GPT-4o-2024-08-06 & 67.98 & 74.16 & 66.29 & 69.48 & 51.69 & 17.79 & 34.42\\
GPT-4o-mini-2024-07-18 & 52.25 & 58.99 & 55.06 & 55.43 & 37.64 & 17.79 & 47.26\\
Claude-3-Opus-20240229 & 53.93 & 52.81 & 53.37 & 53.37 & 37.64 & 15.73 & 41.79\\
\hline
\multicolumn{7}{c}{{\textit{Open-source Chat LLMs}}} \\
\cdashline{1-8}
Yi-1.5-6B-Chat & 24.16 & 24.72 & 20.79 & 23.22 & 11.24 & 11.98 & 106.6\\
Mistral-7B-Instruct & 6.74 & 7.87 & 6.18 & 6.93 & 1.69 & \underline{5.24} & 310.0\\
Qwen2-7B-Instruct & \underline{36.52} & \underline{41.01} & \underline{42.13} & \underline{39.89} & \underline{25.84} & 14.05 & \underline{54.37}\\
LLaMA3-8B-Instruct & 14.61 & 12.92 & 13.48 & 13.67 & 4.49 & 9.18 & 204.5\\
Yi-1.5-9B-Chat & 32.02 & 30.34 & 32.02 & 31.46 & 15.73 & 15.73 & 100.0\\
\hline
Mistral-Nemo-Instruct-2407 & 23.03 & 23.60 & 24.72 & 23.78 & 11.80 & 11.98 & 101.5\\
DeepSeek-MOE-16B-Chat & 3.93 & 3.93 & 4.49 & 4.12 & 0.00 & \underline{\textbf{4.12}} & $\infty$\\
DeepSeek-V2-Lite-Chat & 10.67 & 15.73 & 11.80 & 12.73 & 3.37 & 9.36 & 277.7\\
Mistral-Small-Instruct-2409 & 38.76 & \underline{41.01} & \underline{45.51} & \underline{41.76} & \underline{27.53} & 14.23 & \underline{51.69}\\
Yi-1.5-34B-Chat & \underline{44.38} & 39.33 & 37.64 & 40.45 & 25.28 & 15.17 & 60.01\\
\hline
LLaMA3-70B-Instruct & 33.71 & 36.52 & 34.83 & 35.02 & 23.03 & \underline{11.99} & 52.06\\
Qwen2-72B-Instruct & 52.25 & 50.56 & 50.00 & 50.94 & 35.39 & 15.55 & 43.94\\
Mistral-Large-Instruct-2407 & \underline{65.73} & \underline{64.61} & \underline{66.85} & \underline{65.73} & \underline{50.00} & 15.73 & \underline{31.46}\\
\hline
\multicolumn{7}{c}{{\textit{Specialized Mathematical LLMs}}} \\
\cdashline{1-8}
DeepSeek-Math-7B-Instruct & 26.40 & 28.09 & 29.78 & 28.09 & 11.24 & 16.85 & 149.9\\
DeepSeek-Math-7B-RL & 32.02 & 33.71 & 36.52 & 34.08 & 20.22 & 13.86 & 68.55\\
NuminaMath-7B-CoT & 33.71 & 34.83 & 30.34 & 32.96 & 17.42 & 15.54 & 89.21\\
Mathstral-7B-v0.1 & 30.34 & 31.46 & 34.27 & 32.02 & 18.54 & 13.48 & 72.71\\
Qwen2-Math-7B-Instruct & 53.93 & 50.00 & 51.69 & 51.87 & 38.76 & 13.11 & 33.82\\
Qwen2-Math-72B-Instruct & 62.92 & 65.17 & 63.48 & 63.86 & 50.00 & 13.86 & \textbf{27.72}\\
\hline
\end{tabular}
}

\label{tabapp: trigonometry}
\end{table*}

\section{Error Analysis}\label{app: error}

We perform error analysis in Section~\ref{analysis: error}, and here we showcase several examples of various error types in Table~\ref{tabapp:error example 1}, \ref{tabapp:error example 2}, and \ref{tabapp:error example 3}.

\begin{figure}[ht]
    \centering
    \includegraphics[width=0.7\linewidth]{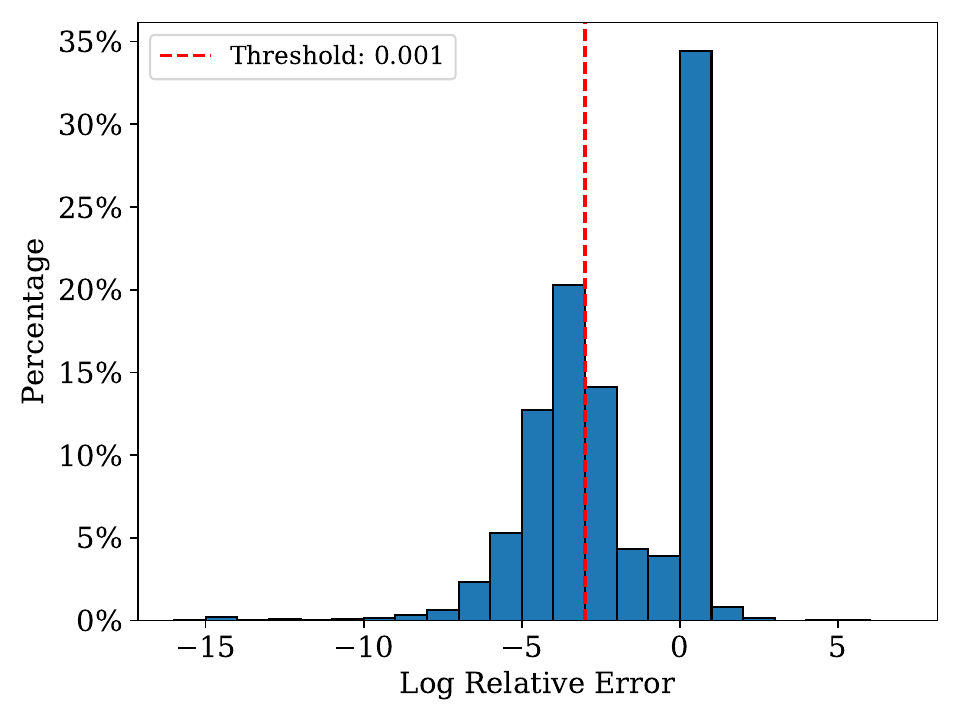}
    \caption{Relative Error of Numeric Answers excluding ones that are identical to the ground truth. Red dotted line indicates the tolerance threshold when we evaluate model answers.}
    \label{fig:numeric_error_hist}
\end{figure}

\revise{The distribution of the relative error for OpenAI-o1-mini numerical values is illustrated in Figure~\ref{fig:numeric_error_hist}.
We excluded numerical answers identical to the ground truth, as their logarithmic relative error would be negative infinity.}

\begin{table}[!t]
\centering
\footnotesize
\caption{An example of Linear algebra with expression calculation error.}

\scalebox{1.0}{
\begin{tabular}{p{10cm}}
\toprule
\textbf{Problem}: 
You have boxes of five types. They are characterized by their length $L$, their width $W$, their height $H$, their age $A$, and their price $P$, as given in the following table:\\
{\centering \includegraphics[width=0.45\linewidth]{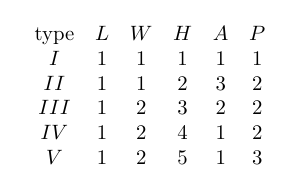}\par}
So for example, boxes of type IV measure 1 by 2 by 4 feet, are a year old, and cost \$2 each.  
Suppose the sum of the lengths of your boxes is 25, the sum of their widths is 40, the sum of their heights is 75, the sum of their ages is 43, and the total price of those boxes is 51. You have [ANS] boxes of type I, [ANS] boxes of type II, [ANS] boxes of type III, [ANS] boxes of type IV, [ANS] boxes of type V.\\
\midrule
\textbf{o1-mini Completion}: 
To solve the system of equations based on the given constraints, we define the variables as follows:
{\centering \includegraphics[width=0.55\linewidth]{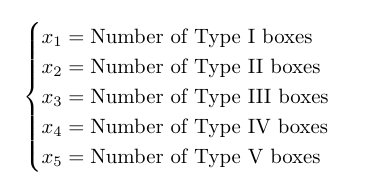}\par}
Using the provided data, we set up the following equations:
{\centering \includegraphics[width=0.6\linewidth]{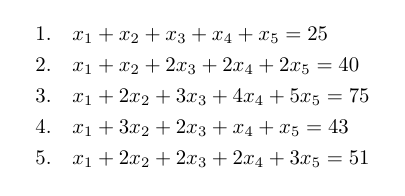}\par}
By systematically solving these equations, we find the values of the variables:
{\centering \includegraphics[width=0.45\linewidth]{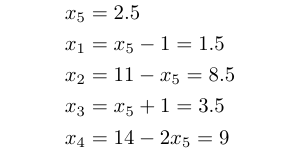}\par}
\textcolor{blue}{[It skipped detailed steps and obtained incorrect results in equation manipulation]} \\

**The final answers are \boxed{1.5, 8.5, 3.5, 9, 2.5}**\\

\midrule
\textbf{Ground-Truth Answer}: 4, 6, 6, 4, 5 \\

\midrule
\textbf{Model Output}: 1.5, 8.5, 3.5, 9, 2.5 \\ 
\bottomrule
\end{tabular}%
}
\label{tabapp:error example 1}
\end{table}

\begin{table}[!t]
\centering
\footnotesize
\caption{An example of Financial mathematics with numeric calculation error.}

\scalebox{1.0}{
\begin{tabular}{p{10cm}}
\toprule
\textbf{Problem}: 
How long will it take for \$3100 compounded semiannually at an annual rate of $1\%$ to amount to \$7000? [ANS] years \\
\midrule
\textbf{o1-mini Completion}: 
To determine how long it will take for \$3100 to grow to \$7000 with semiannual compounding at an annual rate of \(1\%\), we can use the compound interest formula:

\[
A = P \left(1 + \frac{r}{n}\right)^{nt}
\]

Where:
- \(A = 7000\) (the final amount)
- \(P = 3100\) (the principal amount)
- \(r = 0.01\) (the annual interest rate)
- \(n = 2\) (the number of compounding periods per year)
- \(t\) is the time in years

Plugging in the known values:

\[
7000 = 3100 \left(1 + \frac{0.01}{2}\right)^{2t}
\]

\[
\frac{7000}{3100} = \left(1.005\right)^{2t}
\]

\[
2.258064516 = \left(1.005\right)^{2t}
\]

Taking the natural logarithm of both sides:

\[
\ln(2.258064516) = 2t \cdot \ln(1.005)
\]

\[
t = \frac{\ln(2.258064516)}{2 \cdot \ln(1.005)}
\]

Calculating the values:

\[
\ln(2.258064516) \approx \textcolor{red}{0.8154244}
\]
\begin{center}
\textcolor{blue}{[should be 0.8145080376]}
\end{center}
\[
\ln(1.005) \approx \textcolor{red}{0.00498752}
\]
\begin{center}
\textcolor{blue}{[should be 0.00498754141]}
\end{center}
\[
t \approx \textcolor{red}{\frac{0.8154244}{2 \cdot 0.00498752} \approx 81.56}
\]
\begin{center}
\textcolor{blue}{\scalebox{2}{[}With the wrong parameters, should be $\displaystyle \frac{0.8154244}{2 \cdot 0.00498752}=81.74647921$\scalebox{2}{]}}
\end{center}
\begin{center}
\textcolor{blue}{\scalebox{2}{[}With the right parameters, should be $\displaystyle \frac{0.8145080376}{2 \cdot 0.00498754141}=81.6542615$\scalebox{2}{]}}
\end{center}

The final answer is \boxed{81.56}. \\
\midrule
\textbf{Ground-Truth Answer}: 81.6542614995239 \\

\midrule
\textbf{Model Output}: 81.56 \\
\bottomrule
\end{tabular}%
}
\label{tabapp:error example 2}
\end{table}

\begin{table}[!t]
\centering
\footnotesize
\caption{An example of Financial mathematics with flawed reasoning.}

\scalebox{1.0}{
\begin{tabular}{p{10cm}}
\toprule
\textbf{Problem}: 
If you make a deposit into a bank account, at what interest rate (compounded weekly) should you invest if you would like to double your investment in $78$ weeks? [ANS] $\%$ (Note: Your answer should be accurate to two decimal places) \\
\midrule
\textbf{o1-mini Completion}: 
To determine the interest rate \( r \) (compounded weekly) required to double your investment in \( 78 \) weeks, we can use the compound interest formula:

\[
FV = PV \times \textcolor{red}{\left(1 + \frac{r}{100}\right)^n}
\]
\[
\textcolor{blue}{\left[\text{here should be } \left(1+\frac{r}{52}\right)^n \text{ for 52 weeks a year}\right]}
\]

Where:
- \( FV \) is the future value
- \( PV \) is the present value
- \( r \) is the weekly interest rate (in percentage)
- \( n \) is the number of weeks

Given that you want to double your investment:

\[
2 = \left(1 + \frac{r}{100}\right)^{78}
\]

To solve for \( r \), take the natural logarithm of both sides:

\[
\ln(2) = 78 \times \ln\left(1 + \frac{r}{100}\right)
\]

\[
\ln\left(1 + \frac{r}{100}\right) = \frac{\ln(2)}{78}
\]

\[
1 + \frac{r}{100} = e^{\frac{\ln(2)}{78}}
\]

\[
\frac{r}{100} = e^{\frac{\ln(2)}{78}} - 1
\]

\[
r = \left(e^{\frac{\ln(2)}{78}} - 1\right) \times 100
\]

Calculating the above:

\[
r \approx \left(e^{\frac{0.693147}{78}} - 1\right) \times 100 \approx 0.89\%
\]

The final answer is \boxed{0.89}. \\
\midrule
\textbf{Ground-Truth Answer}: 46.4157 \\

\midrule
\textbf{Model Output}: 0.89 \\
\bottomrule
\end{tabular}%
}
\label{tabapp:error example 3}
\end{table}

\section{\revise{Results Across Different Topics}}\label{app:topics}
\begin{figure}[ht!]
    \centering
    \includegraphics[width=0.8\paperwidth,keepaspectratio]{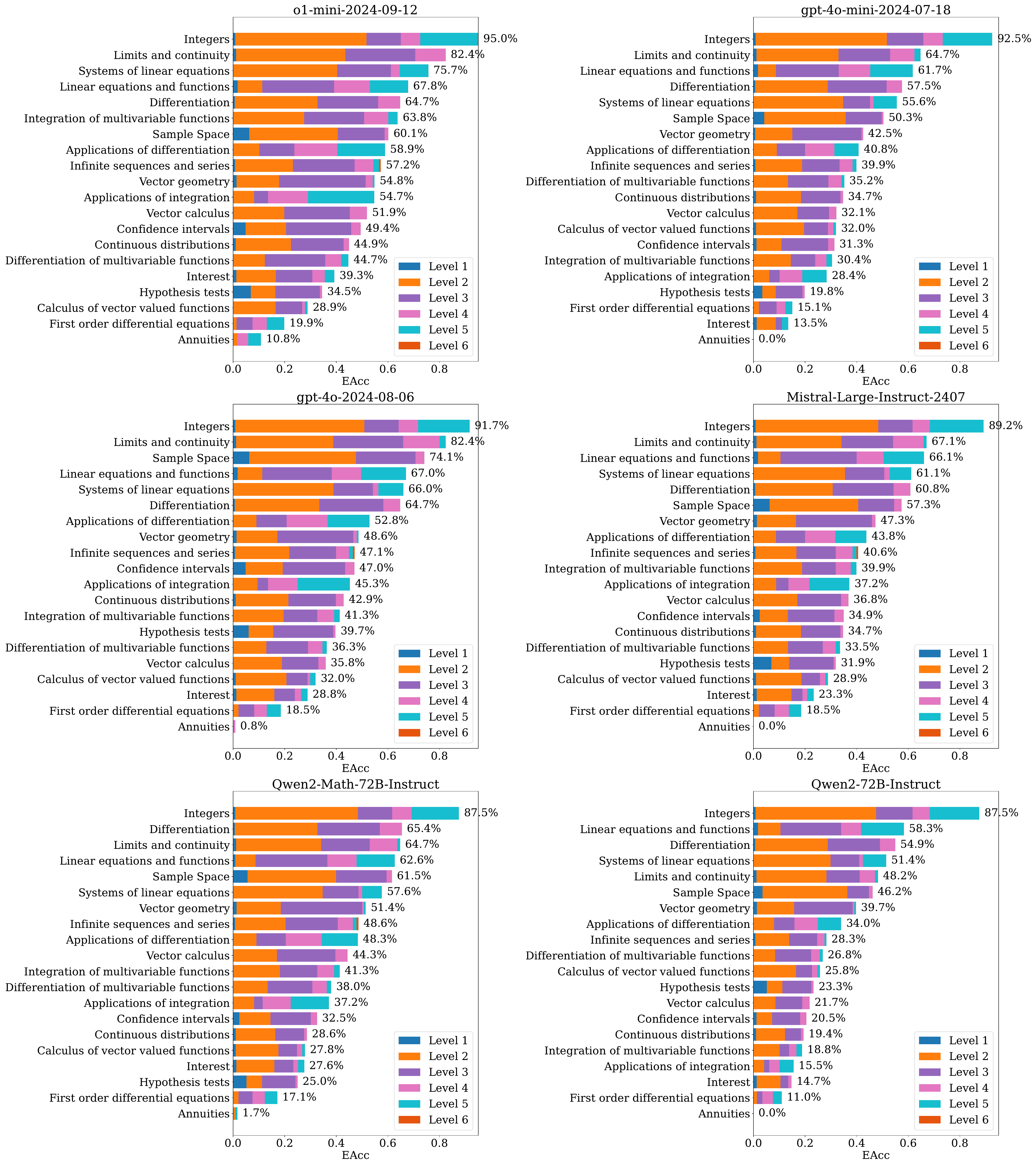}
    \caption{Model accuracy across topics}
    \label{fig:topic_bar_plots_1}
\end{figure}

\begin{figure}[ht!]
    \centering
    \includegraphics[width=0.8\paperwidth,keepaspectratio]{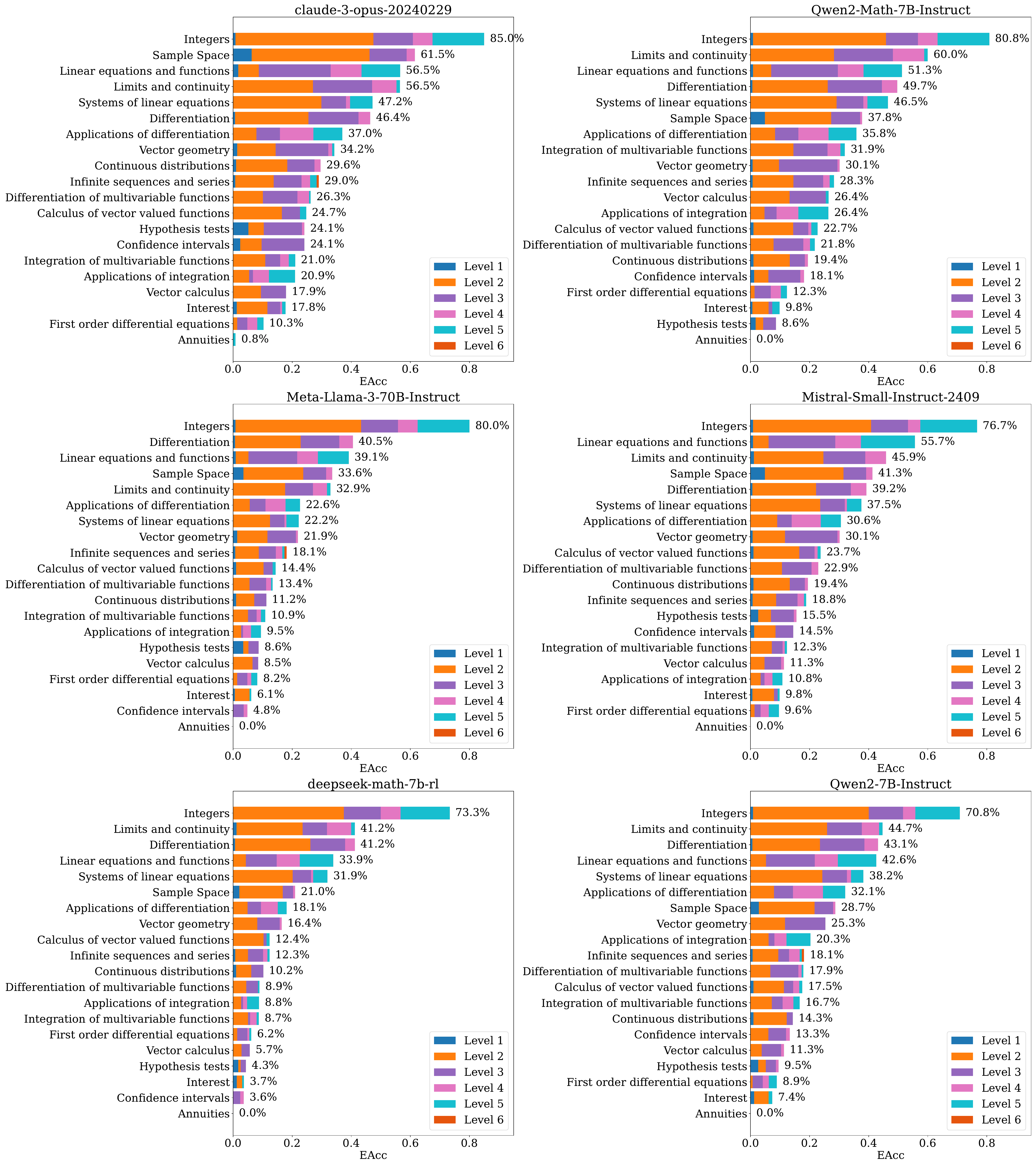}
    \caption{Model accuracy across topics}
    \label{fig:topic_bar_plots_2}
\end{figure}

\begin{figure}[ht!]
    \centering
    \includegraphics[width=0.8\paperwidth,keepaspectratio]{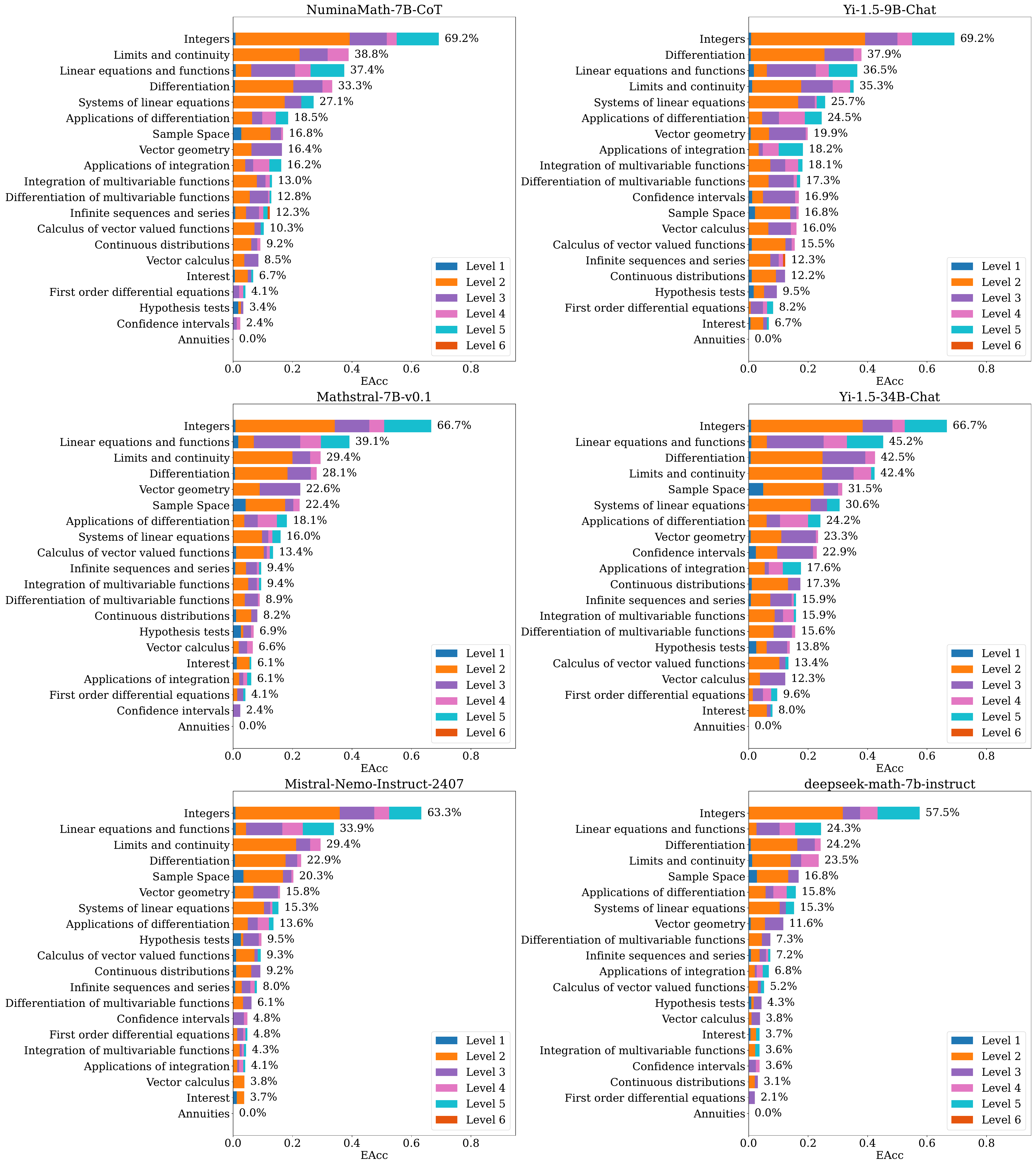}
    \caption{Model accuracy across topics}
    \label{fig:topic_bar_plots_3}
\end{figure}

\begin{figure}[ht!]
    \centering
    \includegraphics[width=0.8\paperwidth,keepaspectratio]{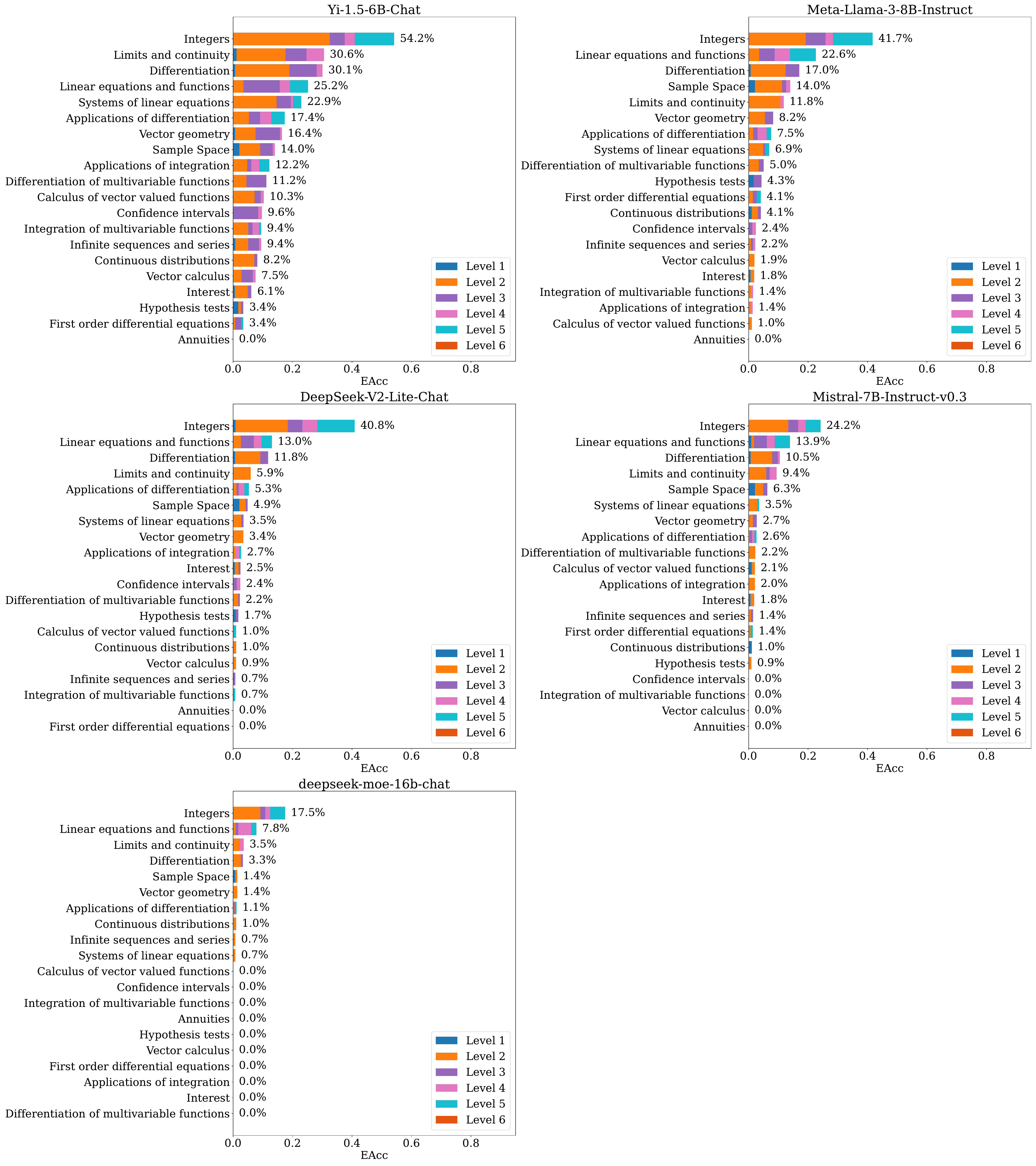}
    \caption{Model accuracy across topics}
    \label{fig:topic_bar_plots_4}
\end{figure}

\revise{The performances of differerent LLMs on 20 topics are shown in Figure \ref{fig:topic_bar_plots_1}, \ref{fig:topic_bar_plots_2}, \ref{fig:topic_bar_plots_3} and \ref{fig:topic_bar_plots_4}.
We observe that different LLMs exhibit varying performance patterns across these topics, and even models within the same family show differences in their rankings.}

\section{\revise{Refinement Results}}\label{app: php}

\revise{
Progressive-Hint Prompting (PHP) \citep{PHP2023Zheng} is a technique designed to enhance automatic, iterative interactions with LLMs. 
PHP uses previously generated answers as hints to progressively guide users toward the correct solutions. 
In our experiments, we employ the zero-shot manner to ensure a fair comparison with the primary experiments in Table~\ref{tab: main_results}, which helps to assess the impact of refinement on the performance of LLMs in solving undergraduate-level mathematical problems. 
We have set the maximum number of interaction rounds to five. 
To save cost, we only experiment with GPT-4o for closed-source LLMs.
The results are presented in Table~\ref{tabapp: php}. 
While PHP can improve AAcc and EAcc in most cases, the improvements are not substantial. 
There remains considerable potential for enhancing the mathematical reasoning abilities of LLMs in solving undergraduate-level mathematics. 
The results for PHP across different subjects for GPT-4o (see Table~\ref{tabapp:php_gpt4o}) indicate that the impact of PHP is subject-agnostic.
}

\begin{table*}[!thb]
\centering
\footnotesize
\caption{\textbf{Results of PHP on {\benchmark}} (all figures are in \%). 
"diff." refers to the improvement over results in Table~\ref{tab: main_results}.
The best results within column with "diff." are \textbf{bolded} and the worst results are \underline{underlined}.}
\scalebox{1.0}{
\begin{tabular}{lcccccc}
\hline
\textbf{Models} & \textbf{AAcc} & diff. \textbf{AAcc} & \textbf{EAcc}  &  diff. \textbf{EAcc} & \textbf{$\Delta$} & diff. \textbf{$\Delta$} \\
\hline
\multicolumn{6}{c}{{\textit{Closed-source LLMs}}} \\
\cdashline{1-7}
GPT-4o-2024-08-06 & 60.73 & +0.36 & 50.44 & +0.50 & 10.29 & -0.14  \\
\hline
\multicolumn{6}{c}{{\textit{Open-source Chat LLMs}}} \\
\cdashline{1-7}
Yi-1.5-6B-Chat & 26.11 & -0.14 & 15.57 & +0.24 & 10.54 & -0.38 \\
Mistral-7B-Instruct & 10.76 & +0.20 & 4.41 & -0.03 & 6.35 & +0.23\\
Qwen2-7B-Instruct & 36.38 & 0.0 & 25.29 & 0.14 & 11.09 & -0.14 \\
LLaMA3-8B-Instruct & 16.41 & -0.14 & 8.61 & -0.3 & 7.8 & +0.16 \\
Yi-1.5-9B-Chat & 34.50 & +0.21 & 21.75 & +0.63 & 12.75 & -0.42 \\
\hline
Mistral-Nemo-Instruct-2407 & 25.46 & +0.38 & 15.19 & -0.24 & 10.27 & \underline{+0.62}  \\ 
DeepSeek-MOE-16B-Chat & 6.34 & +0.54 & 2.07 & +0.11 & 4.27 & +0.44 \\
DeepSeek-V2-Lite-Chat & 13.98 & +0.9 & 6.44 & +0.75 & 7.54 & +0.25 \\
Mistral-Small-Instruct-2409 & 39.66 & -0.56 & 27.90 & -0.94 & 11.76 & +0.38 \\ 
Yi-1.5-34B-Chat & 37.57 & +0.04 & 23.83 & -0.51 & 13.74 & +0.47 \\
\hline
LLaMA3-70B-Instruct & 33.93 & +0.31 & 23.24 & -0.03 & 10.69 & +0.34 \\
Qwen2-72B-Instruct & 47.89 & +0.13 & 36.08 & +0.30 & 11.81 & -0.17 \\
Mistral-Large-Instruct-2407 & 55.97 & +0.04 & 45.17 & +0.13 & 10.8 & -0.09 \\
\hline
\multicolumn{6}{c}{{\textit{Specialized Mathematical LLMs}}} \\
\cdashline{1-7}
DeepSeek-Math-7B-Instruct & 24.57 & +0.68 & 14.90 & +1.29 & 9.67 & \textbf{-0.61} \\
DeepSeek-Math-7B-RL & 29.42 & -0.23 & 19.58 & +0.24 & 9.84 & -0.47 \\
NuminaMath-7B-CoT & 27.81 & \underline{-1.99} & 16.62 & \underline{-2.19} & 11.19 & +0.20\\
Mathstral-7B-v0.1 & 29.75 & \textbf{+1.24} & 19.26 & \textbf{+1.32} & 10.49 & -0.08 \\
Qwen2-Math-7B-Instruct & 42.65 & -1.08 & 31.18 & -1.08 & 11.47 & 0.0 \\
Qwen2-Math-72B-Instruct & 57.35 & + 0.32 & 46.04 & +0.19 & 11.31 & +0.13 \\
\hline
\end{tabular}
}

\label{tabapp: php}
\end{table*}

\begin{table}[!t]
\centering
\footnotesize
\caption{\textbf{Results across different subjects of PHP for GPT-4o}. (all figures are in \%). 
"diff." refers to the improvement over results in Table~\ref{tab: main_results}.
The best results within column with "diff." are \textbf{bolded} and the worst results are \underline{underlined}.}
\scalebox{1.0}{
\begin{tabular}{lcccccc}
\toprule
\textbf{Subject} & \textbf{AAcc} & diff. \textbf{AAcc} & \textbf{EAcc}  &  diff. \textbf{EAcc} & \textbf{$\Delta$} & diff. \textbf{$\Delta$} \\
\midrule
Arithmetic & 92.11 & +0.29 & 88.47 & +1.20 & 3.64 & -0.91 \\
Algebra & 72.33 & +0.97 & 65.35 & +0.68 & 6.98 & + 0.29 \\
Set theory and logic & 71.98 & -1.45 & 57.97 & -1.45 & 14.01 & 0.0 \\
Trigonometry & 67.79 & \underline{-1.69} & 53.37 & +1.68 & 14.42 & -3.37 \\
Combinatorics & 78.79 & +1.14 & 65.91 & +4.55 & 12.88 & \textbf{-3.41} \\
Geometry & 67.49 & +3.31 & 54.04 & +2.49 & 13.45 & \underline{+1.82} \\
Calculus single-variable & 65.11 & +0.24 & 53.26 & +0.21 & 11.85 & +0.03 \\
Calculus multivariable & 49.85 & +0.21 & 38.53 & +0.30 & 11.32 & -0.09 \\
Linear Algebra & 62.85 & +0.47 & 52.81 & +1.20 & 10.04 & -0.73 \\
Number Theory & 65.22 & +1.45 & 54.35 & +2.18 & 10.87 & -0.73 \\
Financial Mathematics & 29.87 & +0.49 & 20.81 & +2.60 & 9.06 & -2.11 \\
Probability & 65.38 & -0.59 & 52.68 & \underline{-2.08} & 12.7 & +1.49 \\
Statistics & 60.27 & -0.16 & 48.88 & -0.50 & 11.39 & +0.34 \\
Complex Analysis & 70.59 & +1.31 & 58.82 & 0.0 & 11.77 & +1.31 \\
Differential Equations & 25.03 & 0.33 & 18.03 & -1.31 & 7.00 & +1.64 \\
Abstract Algebra & 53.17 & \textbf{+4.76} & 35.71 & \textbf{+7.14} & 7.46 & -2.38 \\
\bottomrule
\end{tabular}
}

\label{tabapp:php_gpt4o}
\end{table}

\section{\revise{Reasoning Gap and Test Set Contamination}}\label{app: test_set_contamination}

\revise{Given the limited number of examples in the test set for one version (in terms of SFT), it is necessary to mix the test set with general mathematical SFT data. 
For our experiments, we adopt MetaMathQA \citep{metamath2023yu}, a high-quality SFT dataset for math word problems, which includes 395,000 training examples. 
We use Llama-3-8B as our base model and set the maximum output token length to 4096. 
Following \citet{dartmath2024tong}, we set the learning rate to 5e-5, use a warmup ratio of 0.03, adopt cosine decay, and train the model for one epoch. 
Training one model takes approximately three hours on four A100 GPUs. 
The results are presented in Table~\ref{tab:contamination}. 
We set the proportion of the test set (from one version) for SFT to 5\%, 10\%, 15\% and, 20\% and see how the reasoning gap varies.
As the proportion of the test set included in the training data increases, the reasoning gap ($\Delta$))also becomes more pronounced. This study serves as an initial investigation into test set contamination during the SFT stage. It is important to note that contamination at the pre-training stage is also a significant area of interest \citep{contamination22022razeghi, contamination12024jiang}.}

\end{document}